%% file: neurips_2026.tex
\documentclass{article}

 \usepackage[main, final]{neurips_2026}

\usepackage[utf8]{inputenc} 
\usepackage[T1]{fontenc}    
\usepackage{hyperref}       
\usepackage{url}            
\usepackage{booktabs}       
\usepackage{amsfonts}       
\usepackage{nicefrac}       
\usepackage{microtype}      
\usepackage{xcolor}         
\usepackage{amsmath}
\usepackage{booktabs}
\usepackage{multirow}
\usepackage{array}
\usepackage{graphicx} 
\usepackage{xcolor}
\usepackage[table]{xcolor}  
\usepackage{tikz}
\usepackage{caption}
\usepackage{amssymb}
\usepackage{pifont}
\usepackage{wrapfig}
\usepackage{titletoc}
\usepackage{amsmath,amssymb,amsthm}
\usepackage{algorithm}
\usepackage{algpseudocode}

\usepackage{siunitx}

\newcommand{\cmark}{\ding{51}}

\definecolor{red}{RGB}{240,128,128}
\definecolor{green}{RGB}{60,179,113}
\definecolor{bestcolor}{RGB}{245,222,179}
\definecolor{secondcolor}{RGB}{253,245,230}

\newcommand{\avgdelta}[2]{%
\begin{tikzpicture}[baseline=(n.base)]
\node[inner sep=0pt] (n) {#1};
\node[anchor=south east, xshift=+20pt, yshift=-2pt, font=\scriptsize] 
at (n.south east) {#2};
\end{tikzpicture}%
}

\newcommand{\avgdeltacolor}[3]{%
\begin{tikzpicture}[baseline=(n.base)]
\node[inner sep=0pt] (n) {#1};
\node[anchor=south east, xshift=+22pt, yshift=-2pt, 
font=\scriptsize, text=#3] 
at (n.south east) {#2};
\end{tikzpicture}%
}

\newcommand{\best}[1]{\cellcolor{bestcolor}\textbf{#1}}

\title{Experience Makes Skillful: \\ Enabling Generalizable Medical Agent Reasoning \\ via Self-Evolving Skill Memory}

\author{
Haoran Sun$^{1,2}$\thanks{These authors contributed equally.},
Wenjie Li$^{2}$\footnotemark[1],
Yujie Zhang$^{2}$\footnotemark[1],
Zekai Lin$^{1}$,
Fanrui Zhang$^{3}$,\\
\bfseries
Kaitao Chen$^{2}$,
Xingqi He$^{1}$,
Yichen Li$^{4}$,
Mianxin Liu$^{2}$,
Lei Liu$^{1}$\thanks{Corresponding authors.},
Yankai Jiang$^{2}$\footnotemark[2]\\[0.5em]
$^{1}$Fudan University\\
$^{2}$Shanghai Artificial Intelligence Laboratory\\
$^{3}$University of Science and Technology of China\\
$^{4}$Huazhong University of Science and Technology\\[0.5em]
\texttt{leiliu@fudan.edu.cn, jiangyankai@pjlab.org.cn}
}

\begin{document}

\maketitle

\begin{abstract}
Medical agent systems are increasingly expected to support interactive clinical decision making rather than only static question answering. In such settings, effective agents must reuse prior experience across evolving cases, yet existing memory mechanisms often retain raw historical traces that are redundant, noisy, and difficult to govern. More importantly, they rarely distinguish which memories are truly useful for future reasoning. This limits their ability to accumulate compact and reliable experience for long-horizon clinical reasoning.
To close this gap, we propose \textbf{SkeMex}, a post-deployment self-evolution framework that improves medical agents through a skill-based memory without updating model weights. SkeMex distills informative interaction trajectories into structured skills that encode reusable procedural knowledge, and organizes them into a multi-branch repository spanning general, task-specific, and action-level experience. To determine which memories should be reused and retained, SkeMex estimates context-dependent utility from environment feedback and uses it to guide value-aware retrieval and repository governance. A closed-loop ``Read--Write--Assess--Govern" lifecycle further supports continual evolution by writing new skills, updating utilities, promoting useful memories, and removing harmful entries.
Experiments across diverse clinical tasks show that SkeMex consistently outperforms representative memory-based agents in both offline and online settings. It also generalizes across model backbones and supports transferable skill memory. Code will be available at \href{https://github.com/manglu097/SkeMex}{https://github.com/manglu097/SkeMex}
\end{abstract}

\section{Introduction}

Recent advances in medical large language models (LLMs) \cite{dou2025baichuan, jiang2025hulu, sellergren2025medgemma, xu2025lingshu} and agent-based systems \cite{kim2024mdagents, li2024mmedagent, shi2024ehragent, tang2024medagents} have achieved strong performance on medical benchmarks, especially knowledge-intensive and exam-style datasets such as MedQA \cite{jin2021disease}, MedMCQA \cite{pal2022medmcqa}, and MedBullets \cite{chen2025benchmarking}. These benchmarks are widely used to evaluate medical reasoning and factual recall, and recent models now reach highly competitive results, sometimes approaching or exceeding human performance on selected subsets \cite{singhal2023large, singhal2025toward}. However, most evaluations remain static and single-turn, with fixed inputs and predefined criteria. Real clinical decision making is more dynamic. It unfolds over multiple steps and requires continuous interaction, evidence gathering, multimodal interpretation, hypothesis revision, and action adjustment under uncertainty \cite{esteva2019guide, topol2019high}.
From a cognitive perspective, clinicians do not rely only on isolated factual knowledge \cite{tulving1972episodic}. Their expertise develops through accumulated experience across cases. Prior encounters are recalled, compared, and gradually abstracted into reusable patterns that guide future decisions \cite{barrows1987clinical, kolodner1992introduction}. This process combines semantic knowledge, memory of past cases, and procedural knowledge about how to act in specific situations. It enables clinicians to adapt to new scenarios that are not fully covered by training data or textbooks. 
\begin{wrapfigure}{r}{0.5\linewidth}
    \centering
    \includegraphics[width=\linewidth]{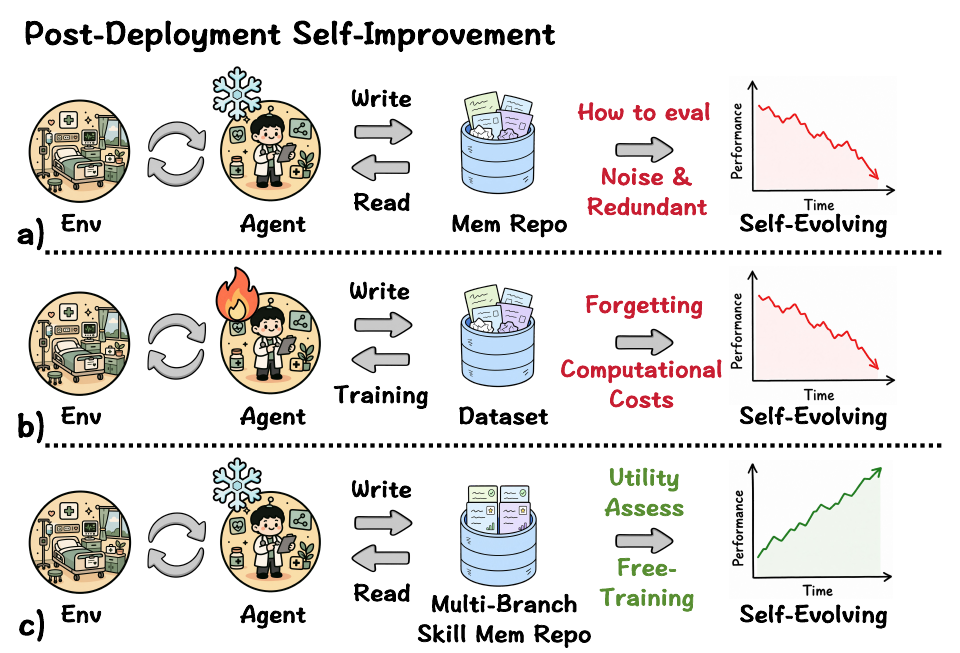}
    \caption{Comparison of (a) conventional memory, (b) training-based methods, and (c) our method.}
    \label{fig1}
\end{wrapfigure}
To address this gap, recent work explores memory-augmented medical agents that store and reuse past interactions \cite{lai2025patient, li2024agent, ren2025healthcare, wei2024medco}. Early methods mainly focus on intra-task memory, which records observations, reasoning steps, and tool interactions within a single clinical case to maintain long-horizon coherence \cite{hu2026landscape, xu2026comprehensive}. As LLMs gain longer contexts and stronger in-context reasoning \cite{qwen36_35b_a3b, team2024gemini}, memory is increasingly used not only to extend context, but also to organize and reuse experience across tasks.
This shift motivates inter-task memory, where agents use prior cases, decisions, or failures to solve new problems. Case-based reasoning recalls and adapts similar past cases \cite{guo2024ds, guo2025optimizing, kolodner1992introduction, zhou2025memento}, but storing raw cases often produces redundant, noisy, and instance-specific repositories. This limits generalization across diverse clinical scenarios \cite{shi2024ehragent, tang2024medagents, wei2024medco}. Recent work therefore distills trajectories into skills, which provide compact units of reusable procedural knowledge \cite{ni2026trace2skill, xia2026skillrl, zhang2026evoskills}. These skills capture recurring patterns of reasoning and action, making experience more transferable.
Despite this progress, current methods still face two limitations (Figure \ref{fig1}). First, some representative methods couple memory improvement with policy training or self-distillation, requiring parameter updates to incorporate experience-derived knowledge \cite{wang2026skill, xia2026skillrl}. This can be costly and may cause catastrophic forgetting or weak transfer across domains \cite{wu2025evolver, zhang2026memrl}. Second, case-based and skill-based memories often lack mechanisms for evaluating long-term usefulness. Redundant or low-quality entries can therefore accumulate over time. These limitations make it difficult for current memory systems to support improvement from experience.

In this work, we introduce \textbf{SkeMex}, a post-deployment self-evolution framework that enables medical agents to improve through external skill memory without modifying the backbone model. SkeMex treats interaction trajectories as sources of reusable experience and converts informative patterns into structured skills that encode actionable reasoning and decision-making procedures.
These skills are organized in a multi-branch repository covering general reasoning, task-specific knowledge, and action-level operations. During inference, SkeMex performs value-aware retrieval, selecting skills based on both semantic relevance and empirically estimated utility in related clinical contexts. To support continual improvement, it further distills successful or informative interactions into new or updated skills, estimates skill utility from observed outcomes, and maintains the repository through a closed-loop ``Read--Write--Assess--Govern" lifecycle. This lifecycle promotes high-utility skills, merges redundant ones, and removes low-quality or potentially harmful entries.
Accordingly, SkeMex transforms raw interaction histories into an evolving memory that supports reliable experience reuse. It enables experience-driven learning without parameter updates, making agents more scalable and adaptable across clinical environments. 
Our contributions are summarized as follows:
\begin{itemize}
    \item We propose \textbf{SkeMex}, a post-deployment self-evolution framework for medical agents that improves clinical reasoning through a skill-based memory without updating model weights.
    \item We formulate skill-memory evolution as a \textbf{non-parametric} reinforcement process, where clinical feedback provides reward signals to estimate context-dependent utility, a unified measure of memory effectiveness that guides both skill retrieval and repository governance.
    \item We introduce the \textbf{Read--Write--Assess--Govern} lifecycle, a closed-loop mechanism that converts trajectories into reusable skills and maintains a well-governed memory repository.
    \item We demonstrate that SkeMex is a \textbf{plug-and-play} framework that consistently improves performance across diverse clinical tasks, generalizes effectively across different model backbones, and supports transferable skill memory across heterogeneous task settings.
\end{itemize}

\section{Related Works}
\paragraph{LLM-based Medical Agents.}
Recent advances in medical foundation models, including Lingshu \cite{xu2025lingshu}, Hulu-Med \cite{jiang2025hulu}, and MedGemma \cite{sellergren2025medgemma}, have expanded medical reasoning across text, imaging, and multimodal data. Building on these models, medical agents have incorporated retrieval, tool use, and multi-agent collaboration to handle heterogeneous clinical evidence. For example, i-MedRAG \cite{xiong2024improving} and MedRAG \cite{zhao2025medrag} improve medical retrieval through iterative search and knowledge-guided reasoning. EHRAgent \cite{shi2024ehragent} uses code-based reasoning for structured EHR data and maintains long-term case memory, while MMedAgent \cite{li2024mmedagent} selects and composes specialized multimodal tools. At the system level, MedAgents \cite{tang2024medagents}, MDAgents \cite{kim2024mdagents}, MAM \cite{zhou2025mam}, and MedAgent-Pro \cite{wang2025medagent} decompose diagnosis into coordinated multi-agent workflows.
Despite strong task-specific performance, most of these systems still process cases independently and lack mechanisms for accumulating reusable experience. Recent work has begun to incorporate memory and self-improvement. Agent Hospital \cite{li2024agent} studies evolvable agents through simulated clinical practice, AMC \cite{lan2024depression} introduces structured memory for psychiatric tasks, and MACRO \cite{fan2026evolving} extracts reusable tools from execution trajectories. STELLA \cite{jin2025stella} and HealthFlow \cite{zhu2025healthflow} further explore evolving templates and policy refinement for biomedical research. However, these methods often tie memory to specific workflows or organize past cases with heuristic rules, which can limit cross-task generalization. Instead, SkeMex decouples memory evolution from fixed workflows. It distills reusable skills from trajectories, evaluates them with external feedback, and selectively retains high-value experience for continual cross-task learning.

\paragraph{Self-Evolving Memory.}
Memory mechanisms in LLM-based agents were initially introduced to address limited context windows, allowing systems to retain prior interactions or retrieved knowledge through static buffers or retrieval-augmented generation (RAG) \cite{lewis2020retrieval, packer2023memgpt, park2023generative, wang2023augmenting, zhong2024memorybank}. As LLMs support longer contexts \cite{qwen36_35b_a3b, team2024gemini}, memory is no longer only a remedy for context limits. It has become a structured store of experience for summarizing and reusing knowledge from environmental interactions \cite{gao2025survey}.
Early self-evolving memory systems stored raw trajectories or reflections to guide future actions \cite{shinn2023reflexion, wang2023voyager, wen2023dilu, zhao2024expel}. Recent work has moved toward structured and modular designs \cite{li2025memos, zhang2025memevolve}. Agent Workflow Memory \cite{wang2024agent} and Dynamic Cheatsheet \cite{suzgun2026dynamic} convert experiences into reusable procedural routines. In medicine, GSEM \cite{han2026gsem} represents clinical experience with a dual-layer memory graph, while HealthFlow \cite{zhu2025healthflow} organizes successful and failed procedures into structured knowledge for tool use and decision making. More recently, skills have emerged as a compact and generalizable memory form \cite{anthropic2026agentskills}. Trace2Skill \cite{ni2026trace2skill}, SkillClaw \cite{ma2026skillclaw}, and EvoSkills \cite{zhang2026evoskills} further show that trajectories can be distilled into hierarchical skill libraries.
However, managing skill memories remains challenging. Recent methods often frame skill evolution as reinforcement learning or optimization. SkillRL \cite{xia2026skillrl} and Skill-SD \cite{wang2026skill} integrate this process into policy training, but require parameter updates. In medical domains, where reliability are critical, this can be costly and may risk catastrophic forgetting or weaken previously learned clinical behaviors. Contrastly, SkeMex decouples skill-memory evolution from model training. It estimates skill utility from environment feedback to guide retrieval and governance, and updates the repository without modifying parameters.
\section{Method}

\subsection{Problem Formulation}
\label{sec:problem_formulation}

We formulate SkeMex as a Memory-based Markov Decision Process (M-MDP) \cite{zhou2025memento}, where an LLM-based agent interacts with an environment while consulting and updating a memory bank. 
\textbf{Memory-Based Markov Decision Process.}
Following prior work \cite{zhou2025memento}, we formalize this process as
\begin{equation}
\mathcal{T}_{\mathrm{M-MDP}} = \langle \mathcal{S}, \mathcal{A}, \mathcal{P}, \mathcal{E}, \gamma, \mathcal{M} \rangle,
\label{eq:mmdp}
\end{equation}
where $\mathcal{S}$ and $\mathcal{A}$ are the state and action spaces, $\mathcal{P}$ is the transition kernel, $\mathcal{E}$ is the reward function, $\gamma \in [0,1)$ is the discount factor, and $\mathcal{M}$ is the space of finite memory banks. At step $t$, the agent observes a state $s_t \in \mathcal{S}$, which includes the current problem, accumulated observations, and retained execution context. It then consults the memory bank $M_t = \{m_i\}_{i=1}^{N_t} \in \mathcal{M}$ and produces an action $a_t \in \mathcal{A}$, covering reasoning decisions, tool calls, and final responses.
Each memory unit is represented as $m_i = (k_i, c_i, u_i)$, where $k_i$ is a retrieval key, $c_i$ is reusable memory content, and $u_i \in \mathbb{R}$ is a utility statistic reflecting its historical contribution. In SkeMex, memory units are instantiated as skills. After executing $a_t$, the agent receives reward $r_t = \mathcal{E}(s_t, a_t)$ and transitions to $s_{t+1} \sim \mathcal{P}(\cdot \mid s_t, a_t)$. Meanwhile, the memory bank evolves through an update operator $\mathcal{U}$, written as $M_{t+1} = \mathcal{U}(M_t, s_t, m_t, a_t, r_t)$. 
Under this formulation, post-deployment improvement comes from better memory retrieval and memory evolution \cite{silver2025welcome}, rather than parameter updates \cite{ouyang2022training, rafailov2023direct, stiennon2020learning}.
\paragraph{Memory-Augmented LLM-based Agent.}
Based on the M-MDP, SkeMex is defined as a memory-augmented agent whose behavior depends on both the current state and the maintained memory bank. The memory bank $M_t$ is updated across episodes by $\mathcal{U}$, accumulating distilled experience from prior interactions. Given $(s_t, M_t)$, the agent retrieves memory units $m_t \subseteq M_t$ using $\mu(m_t \mid s_t, M_t)$, and then produces an action through the LLM $p_{\theta}(a_t \mid s_t, m_t)$. Formally, the overall policy is given by
\begin{equation}
\pi(a_t \mid s_t, M_t) = \sum_{m \in M_t} \mu\bigl(m \mid s_t, M_t\bigr)\, p_{\theta}\bigl(a_t \mid s_t, m\bigr).
\label{eq:overall_policy}
\end{equation}
Here, $\mu$ selects memory for the current state, while $p_{\theta}$ maps the state and retrieved memory into a reasoning decision, tool call, or final response.
A trajectory is denoted by $\tau = \{M_0, s_0, m_0, a_0, r_0, \ldots, M_{T-1}, s_{T-1}, m_{T-1}, a_{T-1}, r_{T-1}\}$, making explicit that memory is retrieved before each decision and updated after feedback. We factorize the probability as
\begin{equation}
\begin{aligned}
p(\tau) = p_0(M_0, s_0) \prod_{t=0}^{T-1} 
&\mu(m_t \mid s_t, M_t)\, p_{\theta}(a_t \mid s_t, m_t)\, \mathcal{E}(r_t \mid s_t, a_t) \\
&\mathcal{U}(M_{t+1} \mid M_t, s_t, m_t, a_t, r_t)\, \mathcal{P}(s_{t+1} \mid s_t, a_t),
\end{aligned}
\label{eq:trajectory_prob}
\end{equation}
where $p_0(M_0, s_0)$ denotes the initial distribution over the memory bank and task state. $\mathcal{E}(r_t \mid s_t, a_t)$ denotes the environment reward, and $\mathcal{U}(M_{t+1} \mid M_t, s_t, m_t, a_t, r_t)$ captures how feedback updates the memory bank. This factorization explicitly reflects the modular structure of SkeMex.

\begin{figure}[ht]
    \centering
    \includegraphics[width=\linewidth]{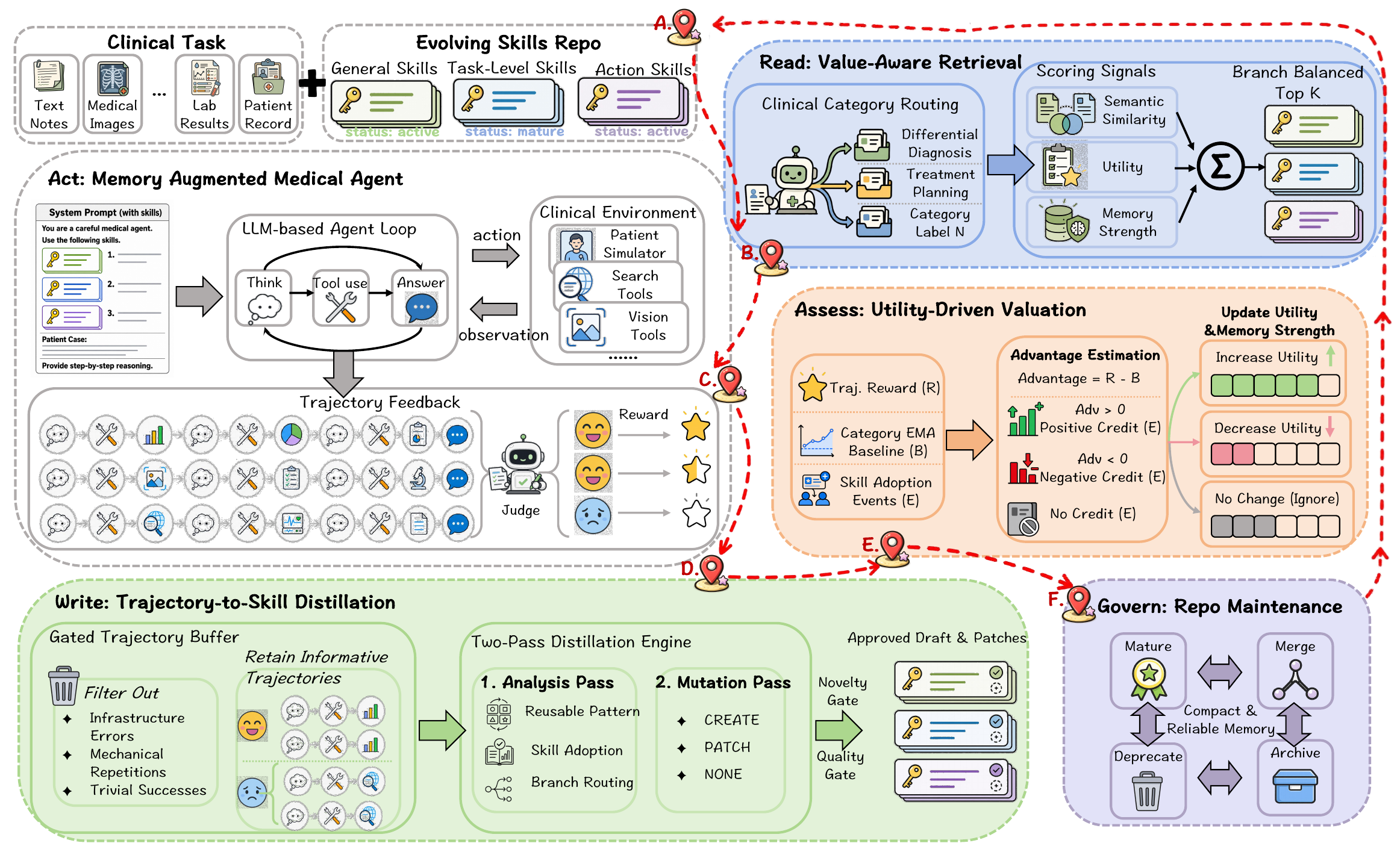}
    \caption{Overview of SkeMex. Components A--F constitute a closed-loop self-evolving cycle (A$\rightarrow$B$\rightarrow$C$\rightarrow$D$\rightarrow$E$\rightarrow$F$\rightarrow$A), enabling continual improvement through iterative memory operations.}
    \label{fig2}
\end{figure}
\subsection{Overview of SkeMex}
\label{sec:overview}
SkeMex is designed to help medical agents accumulate and reuse useful experience from interaction, rather than rely on static contextual augmentation. Its core component is a continuously evolving skills repo, which stores procedural knowledge distilled from prior trajectories.
As shown in Figure \ref{fig2}, SkeMex follows a closed-loop \textbf{Read--Write--Assess--Govern} lifecycle. The repo first supports reasoning through value-aware retrieval (Section \ref{sec:retrieval}). Interaction trajectories and feedback then drive skill distillation (Section \ref{sec:distillation}) and utility estimation (Section \ref{sec:valuation}). Finally, repository governance maintains compactness and reliability by managing skills (Section \ref{sec:lifecycle}). 
Through this cycle, SkeMex enables medical agents to improve from experience while keeping the skill repo compact and reusable.

In SkeMex, every memory unit is instantiated as a structured skill item. Therefore, $M_t$ can be naturally understood as a skills repo. Unlike raw trajectories, which are often lengthy, noisy, and instance-specific \cite{wei2025evo, zhang2025memevolve}, skills provide a compact and reusable representation of experience.
To organize skills at different levels of abstraction, the repo is divided into three branches:
\begin{equation}
M_t = M_t^{\mathrm{gen}} \cup M_t^{\mathrm{task}} \cup M_t^{\mathrm{act}}.
\label{eq:repo_partition}
\end{equation}
The \textbf{general branch} ($M_t^{\mathrm{gen}}$) stores transferable reasoning strategies and broad clinical principles. The \textbf{task-level branch} ($M_t^{\mathrm{task}}$) captures patterns tied to specific task families or clinical categories. The \textbf{action-level branch} ($M_t^{\mathrm{act}}$) records operational knowledge for tool use, such as parameter formatting.
This structure separates general reasoning, task-specific knowledge, and concrete actions during retrieval and valuation. It prevents skills at different abstraction levels from competing in the same pool and allows each branch to be managed within its own scope. See Appendix \ref{app:Case Study} for details.

\subsection{Value-aware Skill Retrieval}
\label{sec:retrieval}

Although the M-MDP allows retrieval at every step, dense retrieval in long-horizon medical tasks can add overhead and fragment the context \cite{lewis2020retrieval, liu2024lost}. Frequent context changes may also weaken consistency across steps. We therefore follow \cite{zhang2026memrl, zhou2025memento} and retrieve skills once at the episode onset ($t=0$). This provides a stable skill context for the full trajectory and reduces noise in utility estimation by using episodic return.
Specifically, retrieval starts with clinical category routing. Given the initial query $s_0$, the agent extracts a category label $\kappa_0$, such as \textit{Differential Diagnosis} or \textit{Treatment Planning}. This label constrains the task-level branch $M_0^{\mathrm{task}}$ and removes skills outside the relevant clinical context. We then apply multi-channel screening to obtain a candidate subset $\widetilde{M}_0 \subseteq M_0$. The screening uses similarity and historical reliability to keep re-ranking efficient and reduce irrelevant candidates.
Each candidate $m \in \widetilde{M}_0$ is then scored with semantic, utility, and temporal signals:
\begin{equation}
\mathrm{Score}_{ret}(m \mid s_0, \kappa_0) = \lambda_{\mathrm{sim}} \mathrm{Sim}(s_0, m) + \lambda_u U(m \mid \kappa_0) + \lambda_h h_0(m),
\label{eq:retrieval_score}
\end{equation}
where $\mathrm{Sim}(s_0, m)$ measures semantic match, $U(m \mid \kappa_0)$ denotes historical utility under clinical category $\kappa_0$, and $h_0(m)$ is memory strength with temporal decay, following the Ebbinghaus forgetting curve \cite{murre2015replication}. Category-conditioned utility is only used for general-branch skills, since broad clinical principles can vary in effectiveness across domains. Instead, task-level and action-level skills are already tied to their contexts. The decay term favors recently reinforced skills and limits the effect of outdated or rarely validated ones. Finally, branch-aware top-$K$ selection balances the three branches.

\subsection{Trajectory-to-Skill Distillation}
\label{sec:distillation}
To make memory writing depend on useful experience, we introduce a gated trajectory buffer $\mathcal{B}^{(w)}$ for each learning window $w$. The gate removes infrastructure errors, mechanical repetitions, and trivial successes. It keeps trajectories that contain meaningful multi-step reasoning or informative failures after skill injection. The buffer also uses a soft isolation policy to preserve a balanced mix of successful and failed trajectories. Each trajectory competes for retention with
$v(\tau) = (\alpha \log(1 + |\tau|) + \beta \cdot \mathbb{I}[\text{injected}]) / (1 + n_\kappa)$,
where $|\tau|$ is the number of reasoning steps, $\mathbb{I}[\text{injected}]$ indicates whether skills were retrieved, and $n_\kappa$ penalizes over-represented clinical categories. This design focuses memory updates on informative clinical patterns, such as causes of misdiagnosis or improved treatment planning, rather than low-value interactions.
Skill writing is performed through a two-pass process. Given a buffered trajectory $\tau \in \mathcal{B}^{(w)}$, the analysis pass extracts a reusable pattern $z_\tau$, identifies adoption signals of previously injected skills, and determines a writing intent $o_\tau \in \{\mathrm{CREATE}, \mathrm{PATCH}, \mathrm{NONE}\}$. It also assigns a target branch for the candidate skill. The mutation pass then turns $z_\tau$ into a new skill draft or applies a local update to an existing skill according to the predicted intent. We formalize this process with a window-level writing operator $\mathcal{W}$:
\begin{equation}
\mathcal{W}(M^{(w)}, \tau) = 
\begin{cases}
M^{(w)} \cup \{\hat{m}_{|\tau|}\}, & \text{if } o_\tau = \mathrm{CREATE}, \\
\mathrm{Patch}(M^{(w)}, \hat{m}_{|\tau|}), & \text{if } o_\tau = \mathrm{PATCH}, \\
M^{(w)}, & \text{if } o_\tau = \mathrm{NONE},
\end{cases}
\label{eq:memory_writing}
\end{equation}
where $\hat{m}_{|\tau|}$ is the candidate skill distilled from trajectory $\tau$, and $M^{(w)}$ is the skills repo snapshot at the end of window $w$. All samples in the same window share this snapshot, keeping $M_t$ consistent during writing.
Before a CREATE draft is committed to the repo, a review pass applies novelty and quality gates. The novelty gate rejects drafts that overlap with existing skills in the same branch and clinical category, and redirects them to PATCH. Additionally, the quality gate requires each skill to include a clear situational trigger and concrete clinical steps, rather than vague principles. 

\subsection{Utility-driven Skill Valuation}
\label{sec:valuation}

Adding a skill to the repo does not guarantee its long-term usefulness. Its value must be tested against clinical outcomes over time. Updating utility from per-sample outcomes can be noisy \cite{shao2024deepseekmath}, since medical tasks differ in difficulty. SkeMex therefore uses window-level valuation, aggregating feedback over multiple trajectories.
To ensure fair credit assignment, skills are evaluated by their relative advantage rather than absolute reward. For each clinical category $\kappa$, we maintain an exponential moving average of rewards $\bar{r}^{(w)}(\kappa)$. The advantage of a trajectory $\tau$ is defined as $A_\tau = r_\tau - \bar{r}^{(w)}(\kappa_\tau)$. Within window $w$, we assign credit to each adoption event using the following contribution function:
\begin{equation}
c(\tau, m_i) =
\begin{cases}
\lambda_{+} \cdot A_\tau, & \text{if } m_i \text{ is positively adopted}, \\
-\bigl(\lambda_{-} + \lambda_{\mathrm{harm}} \cdot \max(0, -A_\tau)\bigr) - \rho_i, & \text{if } m_i \text{ is negatively adopted}, \\
0, & \text{if } m_i \text{ is ignored},
\end{cases}+
\label{eq:contribution}
\end{equation}
where $\lambda_{+}$ scales positive credit, $\lambda_{-}$ is a base penalty for harmful adoption, and $\lambda_{\mathrm{harm}}$ increases the penalty when $A_\tau < 0$. The term $\rho_i = \epsilon \cdot u_i$ is a risk-sensitive regularizer proportional to the current utility of $m_i$, discouraging high-utility skills from accumulating unsafe behavior.
Furthermore, the utility of $m_i$ is updated by aggregating contributions from all adoption events $\mathcal{E}_i^{(w)}$ in window $w$:
\begin{equation}
u_i^{(w+1)} = \mathrm{clip}\!\left(u_i^{(w)} + \eta_i^{(w)} \cdot \frac{1}{|\mathcal{E}_i^{(w)}|} \sum_{\tau \in \mathcal{E}_i^{(w)}} c(\tau, m_i)\right),
\label{eq:utility_update}
\end{equation}
where $\eta_i^{(w)}$ follows a cosine warmup schedule based on cumulative adoption count. This enables faster adjustment for new skills and steadier updates for mature ones. The $\mathrm{clip}$ operation constrains utility to $[0, 1]$, preventing extreme values from dominating retrieval.
As described in Section~\ref{sec:retrieval}, retrieval uses category-conditioned utility $U(m \mid \kappa_0)$ for general-branch skills, with $u_i(\kappa)$ updated per category and global utility $u_i$ averaged across categories. Together, the category-aware baseline and branch-dependent valuation provide stable estimates for retrieval and memory maintenance.

\subsection{Closed-loop Self-evolution Memory}
\label{sec:lifecycle}

The framework described above, including retrieval, distillation, and valuation, forms a unified self-reinforcing cycle. We summarize this overall process with the following operator composition:
\begin{equation}
M^{(w+1)} = \mathcal{G}\Bigl(\mathcal{V}\bigl(\mathcal{W}(M^{(w)}, \mathcal{B}^{(w)})\bigr)\Bigr),
\label{eq:closed_loop}
\end{equation}
where $\mathcal{B}^{(w)}$ denotes gated trajectories in window $w$, $\mathcal{W}$ is the trajectory-to-skill writing operator, $\mathcal{V}$ is the utility valuation operator, and $\mathcal{G}$ is the repo-level governance operator.
At the start of each task, the agent retrieves high-utility skills from the current repo $M^{(w)}$. After task completion, the evaluated trajectory is added to $\mathcal{B}^{(w)}$. At the end of each window, $\mathcal{W}$ distills buffered trajectories into new or updated skills, and $\mathcal{V}$ updates utilities for skills with adoption signals. Governance $\mathcal{G}$ is applied every $N$ windows to keep the repo compact and usable. It merges redundant skills, deprecates low-utility ones, promotes consistently effective skills to mature status, and removes the lowest-utility skills when a branch exceeds its capacity $C^{\mathrm{gen}}, C^{\mathrm{task}}, C^{\mathrm{act}}$. The buffer is then cleared, and the next window starts with the updated repo.
SkeMex does not update the backbone model. Instead, interaction feedback updates the repository, which then guides future retrieval and reasoning.
\section{Experiments and Results}
\label{Experiments and Results}
\subsection{Experiment Settings}
\label{Experiment Settings}
\paragraph{Datasets}
We evaluate SkeMex on nine medical benchmarks covering clinical interaction and knowledge-intensive reasoning. The first group includes AgentClinic \cite{schmidgall2024agentclinic}, LiveClin \cite{wang2026liveclin}, MedJourney \cite{wu2024medjourney}, LiveMedBench \cite{yan2026livemedbench}, HealthBench \cite{arora2025healthbench}, and MediQ \cite{li2024mediq}. These benchmarks involve diagnosis, treatment planning, patient interaction, and rubric-based clinical evaluation. The second group includes MedXpertQA \cite{zuo2025medxpertqa}, MMMU \cite{yue2024mmmu}, and MMMU-Pro \cite{yue2025mmmu}, with the latter two restricted to the Health \& Medicine track. Several datasets include multimodal cases with medical images or clinical tables. 
To make skill accumulation informative, we prioritize diverse and relatively challenging data, since trivial examples provide limited reusable experience. 
Additionally, the accumulated skills are then evaluated on held-out in-domain data from the same benchmark families, while separate benchmark families are reserved for out-of-domain evaluation. Details are provided in Appendix~\ref{app:dataset_details}.

\paragraph{Implementation Details}
We use DeepSeek-V3.2 \cite{liu2025deepseek} as the main backbone and evaluate Qwen3.6-Plus \cite{qwen36plus} with the skill repo built by DeepSeek-V3.2 to test cross-model transfer. The same backbone is used for skill distillation, category classification, and governance. Semantic indexing uses text-embedding-3-large \cite{openai2024embedding}. Retrieval uses top-$K=6$ with $\lambda_{\mathrm{sim}}=0.4$, $\lambda_u=0.4$, and $\lambda_h=0.2$. The learning window contains 30 trajectories. Utility updates use $\lambda_{+}=1.0$, $\lambda_{-}=0.1$, $\lambda_{\mathrm{harm}}=0.5$, and a cosine warmup schedule for $\eta_i^{(w)}$ from 0.05 to 0.20.
The agent uses tools for medical search, knowledge lookup, clinical computation, multimodal analysis, and reasoning self-regulation. Interactive benchmarks also enable patient interaction and examination ordering. All agentic baselines share the same ReAct-style executor and benchmark-specific tool suite and permissions as SkeMex, providing a common execution setup for comparing memory mechanisms. We evaluate SkeMex in offline and online modes. Offline evaluation builds a skill repo from a static split and tests it on held-out in-domain and out-of-domain sets. Online evaluation treats each benchmark as a streaming task sequence, where the repo is updated during interaction. Details are provided in Appendix~\ref{app:tool-suite} \& \ref{app:implementation_details}.
\paragraph{Baselines \& Metrics}
We compare SkeMex with four groups of baselines: (i) medical specialist models \cite{jiang2025hulu, sellergren2025medgemma, xu2025lingshu}, which serve as domain-specific upper bounds, (ii) a memory-free ReAct agent with the same tools \cite{yao2022react}, (iii) retrieval-augmented reflection methods \cite{gou2023critic, shinn2023reflexion}, which reuse past reflections or critiques across tasks, and (iv) self-improving memory agents \cite{fang2025memp, han2026gsem, park2023generative, suzgun2026dynamic, tang2025agent, wang2023voyager, wang2024agent, wang2025mobile, wen2023dilu, wu2025evolver, zhang2025memevolve, zhao2024expel, zheng2025skillweaver}. All memory-based methods use the same training split for experience accumulation and retrieve memory at inference time. Close-ended benchmarks are scored by exact-match accuracy. HealthBench and LiveMedBench use rubric-based scoring, where Gemini-3-Flash \cite{google2025gemini3} judges responses against predefined clinical criteria. Details are provided in Appendix~\ref{Baselines & Metrics}.
\subsection{Main Results}
\label{Main Results}

\begin{table*}[ht]
\centering
\caption{Main results of performance comparison (\%) in the offline setting between SkeMex and baselines. 
The last column shows the improvement of memory-based methods over the memory-free ReAct baseline, highlighting the gains from memory. 
\textbf{Bold} numbers indicate the best performance.}
\label{offlinetab}
\setlength{\tabcolsep}{2pt}
\footnotesize
\renewcommand{\arraystretch}{0.2}
\setlength{\extrarowheight}{1pt}

\resizebox{\linewidth}{!}{
\begin{tabular}{m{2cm} m{1.2cm} cccc cccc}
\toprule
\multirow{2}{*}{\centering\textbf{Backbone}} 
& \multirow{2}{*}{\centering\textbf{Method}} 
& \multicolumn{4}{c}{\textbf{Text}} 
& \multicolumn{3}{c}{\textbf{Multimodal}} 
& \multirow{2}{*}{\textbf{Avg.}} \\
\cmidrule(lr){3-6} \cmidrule(lr){7-9}
& & \textbf{LiveClin} & \textbf{MedXpertQA} & \textbf{HealthBench} & \textbf{LiveMedBench}
& \textbf{LiveClin} & \textbf{MedXpertQA} & \textbf{MMMU} & \\
\midrule

HuluMed-32B & CoT & 76.24 & 32.97 & 9.11 & 35.58 & 58.92 & 38.51 & 58.27 & 44.23 \\[2.5pt]
Lingshu-32B & CoT & 72.28 & 23.78 & 8.88 & 30.45 & 58.92 & 32.43 & 49.64 & 39.48 \\[2.5pt]
MedGemma-27B & CoT & 84.16 & 29.73 & 14.83 & 40.64 & 59.46 & 41.89 & 44.60 & 45.04 \\[2.5pt]
\midrule

\multirow{18}{*}[-60pt]{%
\shortstack[c]{%
\includegraphics[height=3em]{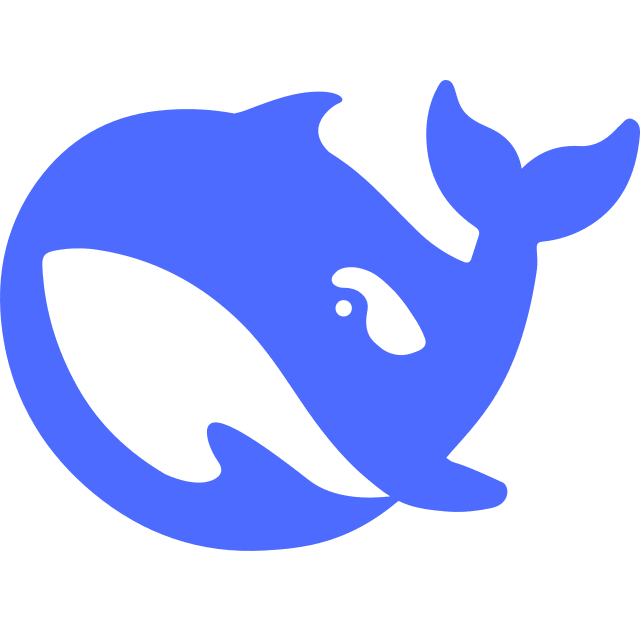}\\[+12pt]
\textbf{DeepSeek-V3.2}%
}%
}
& CoT & 83.17 & 32.43 & 22.42 & 44.93 & 57.30 & 42.57 & 42.45 & 46.47 \\[2.5pt]
& ReAct & 85.15 & 33.51 & 19.06 & 48.64 & 58.38 & 46.62 & 46.04 & 48.20 \\
& Reflexion & 81.19 & 28.11 & 24.54 & 38.75 & 58.92 & 45.95 & 58.99 & \avgdelta{48.06}{-0.14} \\
& CRITIC & 85.15 & 31.89 & 21.53 & 39.26 & 59.46 & 41.22 & 63.31 & \avgdelta{48.83}{+0.63} \\

\cmidrule(lr){2-10}
& Voyager & 83.17 & 29.73 & 23.74 & 52.74 & \best{61.62} & 45.95 & 59.71 & \avgdelta{50.95}{+2.75} \\
& DILU & 82.18 & 32.43 & 23.37 & 53.94 & 59.46 & 46.62 & 62.59 & \avgdelta{51.51}{+3.31} \\
& ExPeL & 82.18 & 33.51 & 23.63 & 37.72 & 59.46 & 45.95 & 57.55 & \avgdelta{48.57}{+0.37} \\
& GM & 80.20 & 30.27 & 23.49 & 54.10 & 57.84 & 46.62 & 61.87 & \avgdelta{50.63}{+2.43} \\
& Memp & 75.25 & 31.89 & 21.44 & 54.79 & 56.22 & 44.59 & 58.99 & \avgdelta{49.02}{+0.82} \\

\cmidrule(lr){2-10}
& SkillWeaver & 91.09 & 32.43 & 23.42 & 52.15 & 57.30 & 44.59 & 62.59 & \avgdelta{51.94}{+3.74} \\
& AWM & 79.21 & 33.51 & 22.84 & 53.39 & 58.92 & 45.95 & 56.83 & \avgdelta{50.09}{+1.89} \\
& Agent KB & 86.14 & 33.51 & 23.08 & 54.59 & 58.92 & 46.62 & 58.99 & \avgdelta{51.69}{+3.49} \\
& Evolver & \best{92.08} & 32.43 & 21.59 & 51.86 & 57.84 & 46.62 & 60.43 & \avgdelta{51.84}{+3.64} \\

\cmidrule(lr){2-10}
& DC & 82.18 & 32.43 & 23.76 & 53.76 & 56.76 & 46.62 & 58.27 & \avgdelta{50.54}{+2.34} \\
& MobileE & 82.18 & 33.51 & 21.97 & 48.96 & 58.92 & 42.57 & 58.27 & \avgdelta{49.48}{+1.28} \\
& CFM & 81.19 & 33.51 & 23.26 & 54.29 & 59.46 & 44.59 & 61.15 & \avgdelta{51.07}{+2.87} \\
& GSEM & 81.19 & 34.59 & 26.00 & 53.20 & \best{61.62} & 48.65 & 60.43 & \avgdelta{52.24}{+4.04} \\

\cmidrule(lr){2-10}
& SkeMex & \best{92.08} & \best{35.68} & \best{27.65} & \best{57.95} & \best{61.62} & \best{50.68} & \best{66.91} & \avgdelta{\best{56.08}}{\best{+7.88}} \\

\midrule

\multirow{18}{*}[-60pt]{%
\shortstack[c]{%
\includegraphics[height=3em]{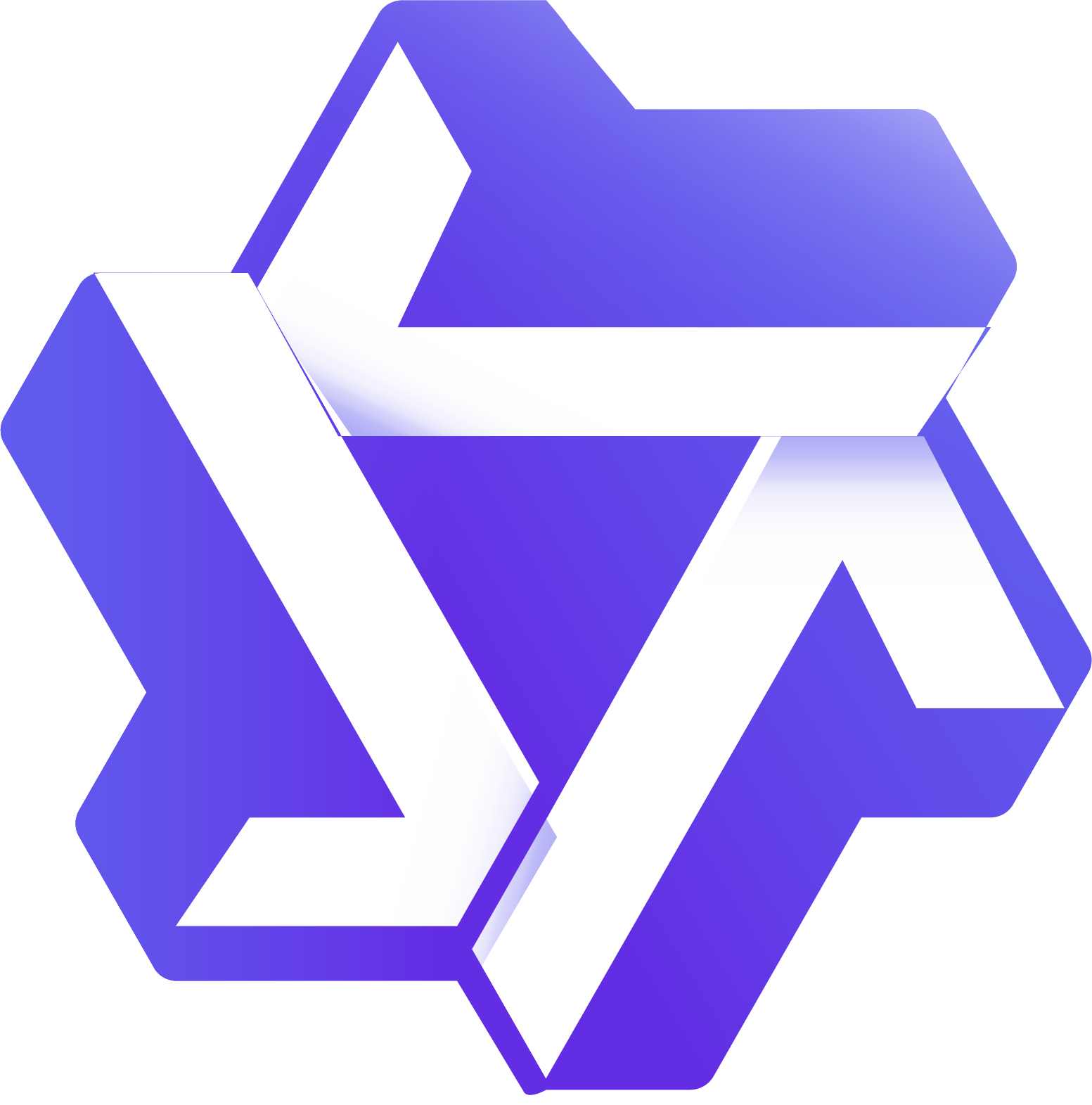}\\[+12pt]
\textbf{Qwen3.6-Plus}%
}%
}
& ReAct & 81.19 & 35.14 & 23.98 & 46.93 & 60.54 & 47.30 & 45.32 & 48.63 \\[2.5pt]
& Reflexion & 85.15 & 42.16 & 26.89 & 48.72 & 70.27 & 48.65 & 58.99 & \avgdeltacolor{54.40}{+5.77}{black} \\
& CRITIC & 86.14 & 44.32 & 26.29 & 47.75 & 68.65 & 48.65 & 58.27 & \avgdeltacolor{54.30}{+5.67}{black} \\

\cmidrule(lr){2-10}
& Voyager & 82.18 & 41.08 & 25.44 & 49.17 & 70.81 & 49.32 & 52.52 & \avgdeltacolor{52.93}{+4.30}{black} \\
& DILU & 86.14 & 40.54 & 26.89 & 49.97 & 68.65 & 50.68 & 52.52 & \avgdeltacolor{53.62}{+4.99}{black} \\
& ExPeL & 84.16 & 44.32 & 27.56 & 47.42 & 70.27 & 50.68 & 57.55 & \avgdeltacolor{54.57}{+5.94}{black} \\
& GM & 86.14 & 42.16 & 28.66 & 48.69 & 70.81 & 54.73 & 57.55 & \avgdeltacolor{55.54}{+6.91}{black} \\
& Memp & 82.18 & 43.78 & 26.74 & 49.06 & 70.81 & 52.03 & 55.40 & \avgdeltacolor{54.29}{+5.66}{black} \\

\cmidrule(lr){2-10}
& SkillWeaver & 86.14 & 43.24 & 27.36 & 50.20 & 71.35 & 51.35 & 58.99 & \avgdeltacolor{55.52}{+6.89}{black} \\
& AWM & 84.16 & 44.32 & 26.60 & 50.39 & 70.81 & 52.70 & 58.27 & \avgdeltacolor{55.32}{+6.69}{black} \\
& Agent KB & 83.17 & 42.70 & 27.23 & 49.01 & 70.27 & 52.03 & 54.68 & \avgdeltacolor{54.15}{+5.52}{black} \\
& Evolver & 84.16 & 44.32 & 25.85 & 47.26 & 69.73 & 52.03 & 54.68 & \avgdeltacolor{54.00}{+5.37}{black} \\

\cmidrule(lr){2-10}
& DC & 85.15 & 42.16 & 25.06 & 48.90 & 70.81 & 50.00 & 53.24 & \avgdeltacolor{53.62}{+4.99}{black} \\
& MobileE & 85.15 & 43.24 & 27.50 & 48.44 & 70.81 & 49.32 & 55.40 & \avgdeltacolor{54.27}{+5.64}{black} \\
& CFM & \best{91.09} & 43.78 & 28.28 & 50.07 & 69.73 & 51.35 & 58.27 & \avgdeltacolor{56.08}{+7.45}{black} \\
& GSEM & 83.17 & 44.32 & 27.57 & 47.61 & 70.81 & 52.03 & 58.27 & \avgdeltacolor{54.83}{+6.20}{black} \\

\cmidrule(lr){2-10}
& SkeMex & \best{91.09} & \best{46.49} & \best{31.79} & \best{53.97} & \best{74.59} & \best{54.73} & \best{61.87} & \avgdeltacolor{\best{59.22}}{\best{+10.59}}{black} \\

\bottomrule
\end{tabular}
}
\vspace{-15pt}
\end{table*}
\vspace{-10pt}
\begin{figure}[ht]
    \centering
    \includegraphics[width=\linewidth]{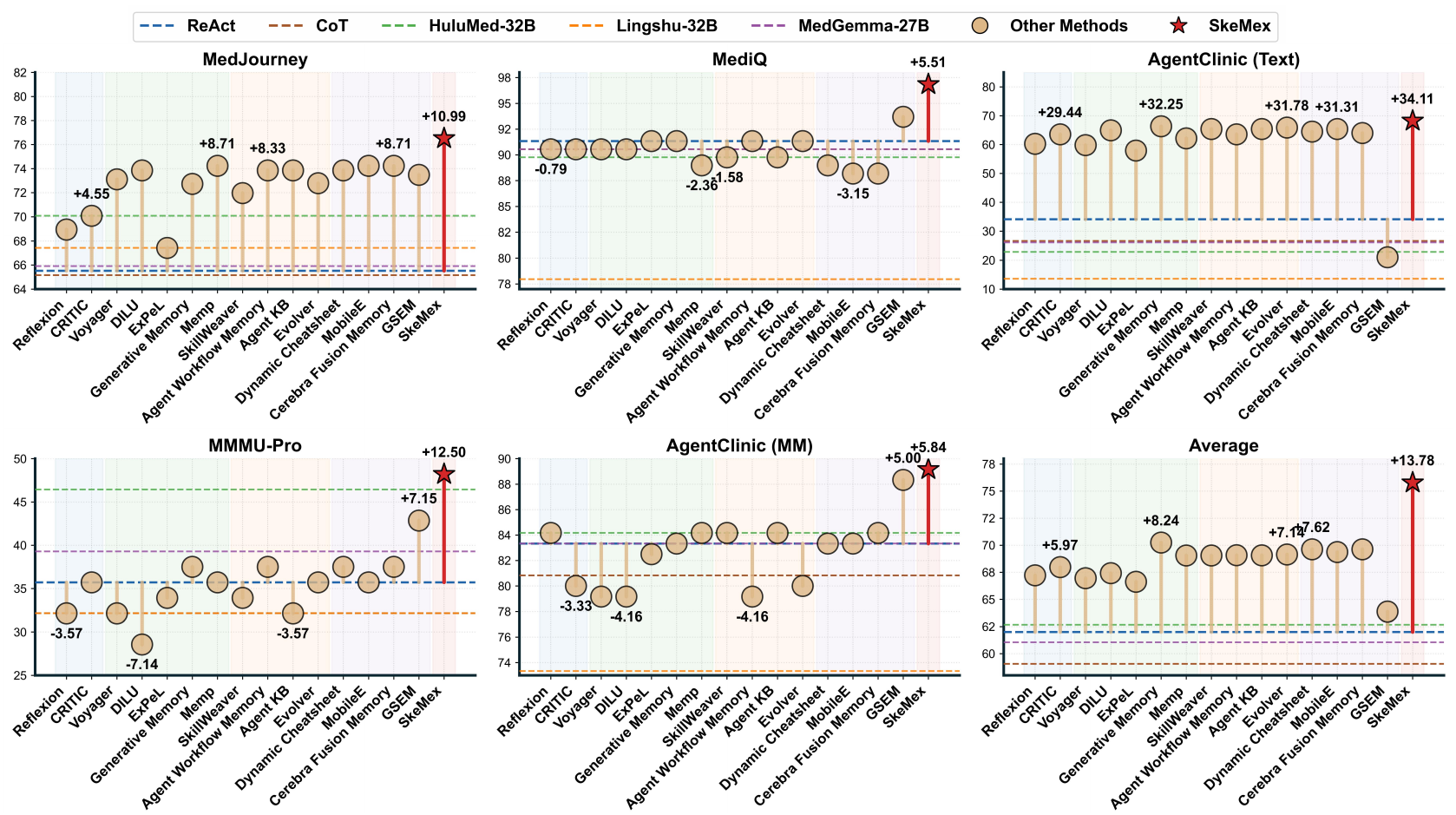}
    \caption{Main results on out-of-domain benchmarks (offline). Background colors denote different types of self-evolving memory methods, with blue for reflection-based methods and red for ours.}
    \label{fig3}
\end{figure}

\paragraph{Offline Mode}
Table~\ref{offlinetab} reports the offline in-domain evaluation, where the skill repo is built from the training split and fixed during testing. This setting tests whether prior medical trajectories can be converted into reusable skills for held-out cases. SkeMex achieves the best average performance on both backbones. With DeepSeek-V3.2, it improves ReAct from 48.20\% to 56.08\%, yielding \textbf{+7.88} points and outperforming the strongest non-SkeMex memory baseline by 3.84 points. With Qwen3.6-Plus, it raises ReAct from 48.63\% to 59.22\%, yielding \textbf{+10.59} points and a 3.14-point lead over the strongest competing memory method. These gains are consistent across models and benchmarks, especially on tasks requiring evidence organization, multi-step reasoning, and verification.
Figure~\ref{fig3} further evaluates the frozen repo on unseen benchmark families. SkeMex remains the strongest method, with a \textbf{+13.78} point gain over ReAct, compared with about +8.24 points for the strongest competing memory baseline. The gain is especially large on AgentClinic-Text, where SkeMex improves by +34.11 points. Several memory baselines fall below ReAct on MediQ or show limited gains on AgentClinic-MM, suggesting negative transfer from less structured memory. In contrast, SkeMex stays above ReAct across all out-of-domain benchmarks, showing more stable transfer.

\paragraph{Online Mode}
Table~\ref{tab:ablation_lifecycle_stages} isolates the major lifecycle stages on five benchmarks across two backbones, with tools, prompts, and inference budgets held fixed within each backbone. We disable skill writing by freezing the repository at initialization; disable utility valuation by assigning equal utilities and removing feedback-based updates; disable governance by removing maturation, merging, deprecation, and capacity control; and disable value-aware retrieval by using similarity-only ranking. Full SkeMex achieves the highest average score across both backbones. Removing skill writing reduces the average by 7.88--9.06 points, while removing utility valuation reduces it by 6.06--6.87 points. Value-aware retrieval and governance contribute 5.58--6.26 and 3.20--3.69 points, respectively. These consistent gains support the complementary roles of experience distillation, feedback-based valuation, utility-guided reuse, and repository maintenance in the closed-loop lifecycle. The following analyses examine the designs within these stages.
\subsection{Ablation Studies and Analyses}
\label{Ablation Studies and Analyses}
\textbf{Skill-memory lifecycle.}
Table~\ref{tab:ablation_lifecycle_stages} evaluates lifecycle stages on five benchmarks across two backbones under matched tools, prompts, and inference budgets. The ablations freeze the initial repository (w/o writing), fix equal utilities without feedback updates (w/o valuation), remove maturation, merging, deprecation, and capacity control (w/o governance), or use similarity-only ranking (w/o value-aware retrieval). Full SkeMex performs best on average with both backbones. Removing writing and valuation reduces average scores by 7.88--9.06 and 6.06--6.87 points, respectively; removing value-aware retrieval and governance reduces them by 5.58--6.26 and 3.20--3.69 points. These results support the complementary roles of skill creation, feedback-based valuation, retrieval, and maintenance. We further examine their internal designs below.

\begin{wraptable}{r}{0.65\linewidth}
\centering
\captionsetup{justification=raggedright, singlelinecheck=false}
\caption{Ablation study on skill-memory lifecycle stages across two backbones. HB, AC, LMB, LC, and MXQA denote HealthBench, AgentClinic, LiveMedBench, LiveClin, and MedXpertQA; \_T and \_M indicate text-only and multimodal settings. $\Delta$ is the average score change relative to full SkeMex.}
\label{tab:ablation_lifecycle_stages}
\setlength{\tabcolsep}{2.5pt}
\renewcommand{\arraystretch}{1.05}
\scriptsize
\resizebox{\linewidth}{!}{
\begin{tabular}{l c c c c c c c}
\toprule
\textbf{Setting} & \textbf{HB} & \textbf{AC\_T} & \textbf{LMB} & \textbf{LC\_M} & \textbf{MXQA\_M} & \textbf{Avg.} & $\boldsymbol{\Delta}$ \\
\midrule
\rowcolor{gray!10}
\multicolumn{8}{l}{\textit{DeepSeek-V3.2}} \\
w/o Skill Writing         & 21.71 & 60.28 & 50.00 & 53.51 & 41.22 & 45.34 & $-7.88$ \\
w/o Utility Valuation     & 22.86 & 63.08 & 51.61 & 55.68 & 42.57 & 47.16 & $-6.06$ \\
w/o Governance            & 25.14 & 65.42 & 54.52 & 58.38 & 46.62 & 50.02 & $-3.20$ \\
w/o Value-aware Retrieval & 19.26 & 62.62 & 53.90 & 55.14 & 47.30 & 47.64 & $-5.58$ \\
\rowcolor{gray!15}
\textbf{Full} & \textbf{27.65} & \textbf{68.22} & \textbf{57.95} & \textbf{61.62} & \textbf{50.68} & \textbf{53.22} & -- \\
\midrule
\rowcolor{gray!10}
\multicolumn{8}{l}{\textit{Qwen3.6-Plus}} \\
w/o Skill Writing         & 24.57 & 56.54 & 43.87 & 65.41 & 45.27 & 47.13 & $-9.06$ \\
w/o Utility Valuation     & 26.29 & 59.81 & 43.87 & 68.65 & 47.97 & 49.32 & $-6.87$ \\
w/o Governance            & 28.57 & 62.62 & 48.06 & 71.89 & 51.35 & 52.50 & $-3.69$ \\
w/o Value-aware Retrieval & 26.29 & 60.75 & 43.55 & 69.73 & 49.32 & 49.93 & $-6.26$ \\
\rowcolor{gray!15}
\textbf{Full} & \textbf{31.79} & \textbf{65.89} & \textbf{53.97} & \textbf{74.59} & \textbf{54.73} & \textbf{56.19} & -- \\
\bottomrule
\end{tabular}
}
\vspace{-10pt}
\end{wraptable}

\textbf{Buffer management and skill encoding }
Table~\ref{tab:ablation_buffer} shows that repository quality depends strongly on which trajectories are written and how they are encoded. Full SkeMex achieves the best average score of 53.22\%. Removing buffer gating gives the largest drop to 47.56\%, indicating that noisy or irrelevant trajectories can corrupt skill extraction. Encoding quality is also important. Single-Prompt Encoding lowers the average to 50.97\%, and removing draft review further reduces it to 48.82\%. Other buffer variants also underperform the full model, suggesting that stable skill evolution benefits from selective trajectory filtering, informative case retention, and review before repository insertion.

\begin{wraptable}{r}{0.65\linewidth}
\captionsetup{justification=raggedright, singlelinecheck=false}
\caption{Main results in the online setting for SkeMex and representative methods. 
AgentClinic and LiveClin are abbreviated as ``AC'' and ``LC'', with ``\_T'' and ``\_M'' denoting text-only and multimodal settings. 
Values show changes from the previous epoch, where red and green indicate decreases and gains.}
\label{onlinetab}
\setlength{\tabcolsep}{3pt}
\renewcommand{\arraystretch}{0.95}
\resizebox{\linewidth}{!}{
\begin{tabular}{l c c c c c c c}
\toprule
\textbf{Method} & \textbf{Epoch} 
& \textbf{AC\_T} 
& \textbf{AC\_M} 
& \textbf{LC\_T} 
& \textbf{LC\_M} 
& \textbf{LiveMedBench} 
& \textbf{Avg.} \\
\midrule
\multirow{3}{*}{Agent KB} 
& epoch@1 & 72.90 & 86.67 & 89.11 & 64.86 & 57.68 & 74.24 \\
& epoch@2 & \avgdeltacolor{73.83}{+0.93}{green} & \avgdeltacolor{89.17}{+2.50}{green} & \avgdeltacolor{90.10}{+0.99}{green} & \avgdeltacolor{66.49}{+1.63}{green} & \avgdeltacolor{57.16}{-0.52}{red} & \avgdeltacolor{75.35}{+1.11}{green} \\
& epoch@3 & \avgdeltacolor{75.23}{+1.40}{green} & \avgdeltacolor{87.50}{-1.67}{red} & \avgdeltacolor{91.09}{+0.99}{green} & \avgdeltacolor{67.57}{+1.08}{green} & \avgdeltacolor{57.37}{+0.21}{green} & \avgdeltacolor{75.75}{+0.40}{green} \\

\midrule
\multirow{3}{*}{AWM} 
& epoch@1 & 70.09 & 89.17 & 83.17 & 61.08 & 56.45 & 71.99 \\
& epoch@2 & \avgdeltacolor{73.36}{+3.27}{green} & \avgdeltacolor{89.17}{+0.00}{gray} & \avgdeltacolor{79.21}{-3.96}{red} & \avgdeltacolor{64.32}{+3.24}{green} & \avgdeltacolor{56.08}{-0.37}{red} & \avgdeltacolor{72.43}{+0.44}{green} \\
& epoch@3 & \avgdeltacolor{74.30}{+0.94}{green} & \avgdeltacolor{91.67}{+2.50}{green} & \avgdeltacolor{82.18}{+2.97}{green} & \avgdeltacolor{63.24}{-1.08}{red} & \avgdeltacolor{56.29}{+0.21}{green} & \avgdeltacolor{73.54}{+1.11}{green} \\

\midrule
\multirow{3}{*}{Evolver} 
& epoch@1 & 72.43 & 85.83 & 93.07 & 65.41 & 56.91 & 74.73 \\
& epoch@2 & \avgdeltacolor{74.30}{+1.87}{green} & \avgdeltacolor{88.33}{+2.50}{green} & \avgdeltacolor{96.04}{+2.97}{green} & \avgdeltacolor{65.95}{+0.54}{green} & \avgdeltacolor{55.43}{-1.48}{red} & \avgdeltacolor{76.01}{+1.28}{green} \\
& epoch@3 & \avgdeltacolor{75.70}{+1.40}{green} & \avgdeltacolor{89.17}{+0.84}{green} & \avgdeltacolor{99.01}{+2.97}{green} & \avgdeltacolor{64.86}{-1.09}{red} & \avgdeltacolor{56.09}{+0.66}{green} & \avgdeltacolor{76.97}{+0.96}{green} \\

\midrule
\multirow{3}{*}{Memp} 
& epoch@1 & 72.43 & 87.50 & 84.16 & 64.32 & 55.18 & 72.72 \\
& epoch@2 & \avgdeltacolor{73.83}{+1.40}{green} & \avgdeltacolor{87.50}{+0.00}{gray} & \avgdeltacolor{88.12}{+3.96}{green} & \avgdeltacolor{64.32}{+0.00}{gray} & \avgdeltacolor{55.25}{+0.07}{green} & \avgdeltacolor{73.81}{+1.09}{green} \\
& epoch@3 & \avgdeltacolor{73.36}{-0.47}{red} & \avgdeltacolor{85.83}{-1.67}{red} & \avgdeltacolor{89.11}{+0.99}{green} & \avgdeltacolor{65.95}{+1.63}{green} & \avgdeltacolor{55.77}{+0.52}{green} & \avgdeltacolor{74.00}{+0.19}{green} \\

\midrule
\multirow{3}{*}{SkillWeaver} 
& epoch@1 & 70.56 & 89.17 & 94.06 & 60.00 & 56.25 & 74.01 \\
& epoch@2 & \avgdeltacolor{73.83}{+3.27}{green} & \avgdeltacolor{85.83}{-3.34}{red} & \avgdeltacolor{94.06}{+0.00}{gray} & \avgdeltacolor{61.08}{+1.08}{green} & \avgdeltacolor{56.10}{-0.15}{red} & \avgdeltacolor{74.18}{+0.17}{green} \\
& epoch@3 & \avgdeltacolor{74.30}{+0.47}{green} & \avgdeltacolor{88.33}{+2.50}{green} & \avgdeltacolor{93.07}{-0.99}{red} & \avgdeltacolor{60.00}{-1.08}{red} & \avgdeltacolor{56.18}{+0.08}{green} & \avgdeltacolor{74.38}{+0.20}{green} \\

\midrule
\multirow{3}{*}{SkeMex} 
& epoch@1 & 72.90 & 90.00 & 95.05 & 65.41 & 58.59 & 76.39 \\
& epoch@2 & \avgdeltacolor{74.77}{+1.87}{green} & \avgdeltacolor{90.83}{+0.83}{green} & \avgdeltacolor{96.04}{+0.99}{green} & \avgdeltacolor{66.49}{+1.08}{green} & \avgdeltacolor{58.71}{+0.12}{green} & \avgdeltacolor{77.37}{+0.98}{green} \\
& epoch@3 & \avgdeltacolor{76.17}{+1.40}{green} & \avgdeltacolor{93.33}{+2.50}{green} & \avgdeltacolor{96.04}{+0.00}{gray} & \avgdeltacolor{67.57}{+1.08}{green} & \avgdeltacolor{59.68}{+0.97}{green} & \avgdeltacolor{78.56}{+1.19}{green} \\
\bottomrule
\end{tabular}
}
\vspace{-10pt}  

\end{wraptable}

\begin{wrapfigure}{r}{0.43\linewidth}
    \centering
    \includegraphics[width=\linewidth]{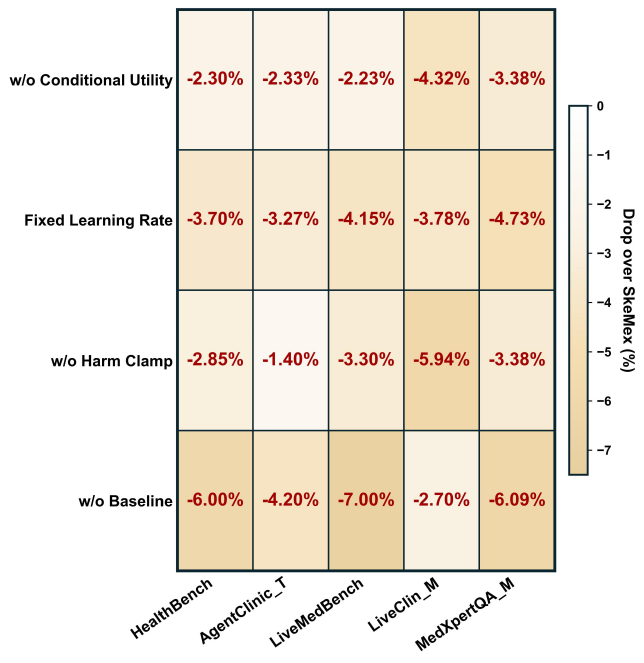}
    \caption{Ablation study on valuation module. Cell values show gaps from SkeMex.}
    \label{fig4}
    \vspace{-10pt}
\end{wrapfigure}

\begin{table*}[ht]
\centering

\begin{minipage}[t]{0.52\textwidth}
\centering
\caption{Ablation study on buffer and encoding. Dataset names are abbreviated for compactness.}
\label{tab:ablation_buffer}

\setlength{\tabcolsep}{2.5pt}
\renewcommand{\arraystretch}{1.05}
\scriptsize

\resizebox{\linewidth}{!}{
\begin{tabular}{l c c c c c c}
\toprule
\textbf{Setting} & \textbf{HB} & \textbf{AC\_T} & \textbf{LMB} & \textbf{LC\_M} & \textbf{MXQA\_M} & \textbf{Avg.} \\
\midrule
\rowcolor{gray!10}
\multicolumn{7}{l}{\textit{Buffer Management}} \\
w/o Buffer Gating    & 20.47 & 65.42 & 54.47 & 56.22 & 41.22 & 47.56 \\
w/o Buffer Splitting & 25.58 & 64.02 & 55.47 & 60.00 & 47.97 & 50.61 \\
w/o Capacity Control & 26.47 & 66.36 & 56.87 & 59.46 & 50.00 & 51.83 \\
FIFO Buffer          & 25.63 & 66.36 & 56.85 & 58.92 & 48.65 & 51.28 \\
\midrule
\rowcolor{gray!10}
\multicolumn{7}{l}{\textit{Encoding}} \\
Single-Prompt        & 25.46 & 65.42 & 55.31 & 60.00 & 48.65 & 50.97 \\
w/o Draft Review     & 23.57 & 65.42 & 55.66 & 56.22 & 43.24 & 48.82 \\
\midrule
\rowcolor{gray!15}
\textbf{Full}        & \textbf{27.65} & \textbf{68.22} & \textbf{57.95} & \textbf{61.62} & \textbf{50.68} & \textbf{53.22} \\
\bottomrule
\end{tabular}
}
\end{minipage}
\hfill
\begin{minipage}[t]{0.46\textwidth}
\centering
\caption{Ablation study on branch combinations. G, T, and A denote general, task-level, and action-level branches, respectively.}
\label{tab:ablation_branch}

\setlength{\tabcolsep}{2.5pt}
\renewcommand{\arraystretch}{1.05}
\scriptsize

\resizebox{\linewidth}{!}{
\begin{tabular}{c c c c c c c c c c}
\toprule
\multicolumn{4}{c}{\textbf{Setting}} 
& \multirow{2}{*}[-2pt]{\textbf{HB}} 
& \multirow{2}{*}[-2pt]{\textbf{AC\_T}} 
& \multirow{2}{*}[-2pt]{\textbf{LMB}} 
& \multirow{2}{*}[-2pt]{\textbf{LC\_M}} 
& \multirow{2}{*}[-2pt]{\textbf{MXQA\_M}} 
& \multirow{2}{*}[-2pt]{\textbf{Avg.}} \\
\cmidrule(lr){1-4}
\# & \textbf{G} & \textbf{T} & \textbf{A} & & & & & & \\
\midrule
1 & \cmark &  &  & 22.22 & 60.28 & 50.61 & 50.27 & 37.84 & 44.24 \\
1 &  & \cmark &  & 19.13 & 60.75 & 49.47 & 52.43 & 45.27 & 45.41 \\
2 & \cmark & \cmark &  & 23.97 & 64.95 & 50.77 & 54.05 & 42.57 & 47.26 \\
2 & \cmark &  & \cmark & 23.04 & 64.49 & 52.49 & 60.00 & 43.24 & 48.65 \\
2 &  & \cmark & \cmark & 19.12 & 61.21 & 46.07 & 52.97 & 45.95 & 45.06 \\
\midrule
\rowcolor{gray!15}
3 & \cmark & \cmark & \cmark 
& \textbf{27.65} & \textbf{68.22} & \textbf{57.95} 
& \textbf{61.62} & \textbf{50.68} & \textbf{53.22} \\
\bottomrule
\end{tabular}
}
\end{minipage}
\vspace{-7pt}
\end{table*}
\textbf{Utility-driven skill valuation}
Figure~\ref{fig4} reports performance drops relative to SkeMex, showing that reliable skill scoring needs more than a simple update rule. Removing baseline correction causes the largest degradation, with drops up to 7.00\% on LiveMedBench and 6.09\% on MedXpertQA-MM. This supports the role of category-aware reward normalization. A fixed learning rate also leads to broad declines, suggesting that adaptive updates are needed for both new and mature skills. Removing harm clamping causes a notable drop on LiveClin-MM, showing the value of harmful feedback control. The consistent decline without conditional utility indicates that skill value should depend on clinical context.

\textbf{Multi-branch memory structure.}
Table~\ref{tab:ablation_branch} shows that the three memory branches provide complementary signals. The full design achieves the highest average score of 53.22\%, outperforming the best partial variant, General + Action, by 4.57 points. This indicates that task-specific medical knowledge is still needed alongside general reasoning and action-level guidance. 
Two-branch and single-branch variants lag behind the full model, showing that no single abstraction level is sufficient. Combining general, task-level, and action-level skills is important for robust medical problem solving.
Additional ablations and extended analyses are provided in Appendix~\ref{app:ablation} and~\ref{app:Further Analyses}.

\section{Conclusion}
In this paper, we introduced \textbf{SkeMex}, a post-deployment self-evolution framework that enables medical agents to improve through skill-based memory without updating model weights. By combining reusable skill distillation, utility-driven valuation, and closed-loop memory governance, SkeMex supports reliable experience accumulation and reuse across diverse clinical tasks. Extensive experiments demonstrate consistent improvements over ReAct and representative memory-based agents, as well as strong generalization across model backbones and task settings. We believe SkeMex offers a scalable step toward medical agents that can mature through continued clinical experience.

\par
\makeatletter
\WF@finale
\makeatother
\section{Acknowledgements and Disclosures}
This research was funded by the Shanghai Science and Technology Innovation Action Plan in Computational Biology (24JS2840200), and the National Key Research and Development Program of China (2024YFA1307702).
The authors declare no competing interests and no financial relationships outside the submitted work that could be perceived as influencing it.


\clearpage
\bibliography{ref}
\bibliographystyle{plain}

\newpage
\appendix
\startcontents[appendix]
\section*{Appendix Contents}
\printcontents[appendix]{l}{1}{\setcounter{tocdepth}{2}}
\clearpage

\section{Agent Architecture}
\label{app:agent_architecture}

In this section, we describe the execution backbone of our agent system and explains how the skill memory introduced in the main paper is connected to task execution. The agent is implemented as a bounded ReAct loop \cite{yao2022react}, where each task is solved through an iterative sequence of model decisions, tool executions, observations, and final response generation. The evolution module is attached to this backbone through a runtime skill injection interface, while the core agent loop remains responsible for maintaining the task state, enforcing the output protocol, invoking tools, and deciding when to terminate. This separation allows the learned skill memory to guide the agent without changing the basic semantics of execution. In other words, the system can use accumulated experience to influence the agent's decisions while keeping the underlying task solving process stable.

\subsection{Runtime Components}
\label{app:agent_components}

For each input sample, the system constructs or reuses an \texttt{AgentLoop} instance. The loop is organized around several runtime components that jointly support structured reasoning, tool use, memory rendering, and context control. The language model interface receives the system prompt and the per step task prompt, then produces structured outputs under the required protocol. The tool registry exposes dataset specific tools that can be called by name, renders the available tool list into the system prompt, and executes one selected tool call at a time. The conversation memory stores the local interaction history of the current task, including the original user request, previous assistant outputs, and observations returned by tools. In addition, the evolution runtime context injects the task category and retrieved skills into the prompt at each step. This injected block is rendered dynamically and is not written into the conversation memory, which avoids repeated duplication across turns. When enabled, the context guard monitors the accumulated trajectory and adds control signals when the task becomes long, repetitive, or close to the context budget.

\begin{table}[ht]
\centering
\caption{Runtime components used by the agent execution backbone.}
\begin{tabular}{p{0.22\linewidth}p{0.70\linewidth}}
\toprule
\textbf{Component} & \textbf{Role in execution} \\
\midrule
Language model & Produces structured outputs under the system prompt and the current task prompt. Each output contains a reasoning block followed by either a tool call or a final response. \\
Tool registry & Renders the available tool list into the system prompt and executes one selected tool call at a time. It also receives per sample runtime context, such as case information or dataset metadata. \\
Conversation memory & Stores the task local interaction history and renders it into a plain text context block before each model call. The history contains the user request, previous assistant outputs, and observations returned by tools. \\
Evolution runtime context & Injects retrieved experience skills and task categories into the prompt at every step. This block is rendered freshly and is not written into the conversation memory, which avoids repeated duplication across turns. \\
Context guard & Provides intra-task context control through loop breaking, confirmed finding pinning, and budget aware history trimming. \\
\bottomrule
\end{tabular}
\label{tab:agent_runtime_components}
\end{table}

At the beginning of each sample, the batch runner clears both the conversation memory and the recorded step history. This reset makes each benchmark instance an independent task and prevents information leakage from earlier samples. If the input sample contains images, their paths are appended to the user request as a numbered list. For vision capable models, the image contents are passed only to the first model call. Later steps rely on the textual conversation history, which avoids repeatedly sending the same visual input and keeps subsequent prompts more compact. Before execution starts, the tool registry is also updated with the sample specific runtime context, so that tools can access the information required by the current benchmark instance.

\subsection{Stepwise Execution Protocol}
\label{app:agent_protocol}

The agent follows a strict structured output protocol that turns open ended model generation into a bounded and auditable action process. For multi step problems, the model may begin with an optional \texttt{\textless planning\textgreater} block, but every turn must contain a \texttt{\textless reasoning\textgreater} block. This reasoning block must be immediately followed by exactly one action. The model either emits a \texttt{\textless tool\textgreater} block containing a tool name and a JSON input, or emits a \texttt{\textless response\textgreater} block that terminates the task with the final answer. By requiring every step to end in one of these two actions, the protocol converts free form generation into a small finite action interface and keeps the execution flow deterministic. At each step, the prompt is assembled from the system instruction, the rendered conversation history, the original user request, and the evolution runtime context when available. The system instruction defines the output grammar, the available tools, and the rule that only one tool may be called in a single turn. The conversation history provides the accumulated task context, including previous assistant outputs and tool observations. The original request is repeated to help the model retain the task objective after several interaction rounds. The evolution block adds retrieved skills and task categories when the evolution module has provided them. These skills are presented as prioritized experience references with explicit applicability conditions, so the model can consult them before selecting the next action and refer to the corresponding skill identifiers when a skill influences its reasoning.

After the model produces an output, the system records the complete content in memory and parses it according to the protocol. If the output contains a valid tool block, the parser extracts the tool name and JSON input, the tool registry executes the selected tool, and the returned observation is appended to the conversation memory before the loop proceeds to the next step. If the output contains a response block, the task terminates and the full trajectory is saved for later analysis. When the output violates the required format, the agent performs controlled recovery by appending a corrective observation rather than terminating immediately. For example, malformed JSON produces a parse error observation, an unknown tool produces an explicit tool name error, and a reasoning block without a following action prompts the model to complete the missing tool or response block. The loop is also constrained by a final step convergence rule. Once the maximum number of allowed steps is reached, the next prompt explicitly forbids further tool use and requires the model to provide a final response. If the model still fails to generate a valid response tag, the system removes structural tags and uses the remaining content as a fallback answer. This rule prevents long running tasks from ending without an output and makes the bounded step budget compatible with benchmark evaluation.

\subsection{Integration of Skill Memory into the Agent Loop}
\label{app:skill_memory_integration}

The skill memory does not replace the agent policy, but enters the agent through a runtime context that is set immediately before solving each sample. The evolution runner first assigns a task category to the sample and retrieves a small set of relevant skills from the skill repository. It then passes the task category, the rendered skill block, and the identifiers of the injected skills to the agent through the context setter. During every step of the same task, this context is rendered into the prompt. Since the block is not persisted in conversation memory, the same skill content does not accumulate across turns. This design creates a clear boundary between retrieval and action. Retrieval determines which experiences are visible to the model, while the agent still decides whether the applicability conditions are satisfied in the current step. A retrieved skill therefore functions as a conditional reference rather than a hard command, which allows the agent to adopt it, ignore it, or override it according to the current observations and task state.

The recorded trajectory also provides evidence for later skill evolution. By inspecting the reasoning traces, tool calls, observations, and final answers, the trajectory encoder can identify which injected skills were adopted, ignored, or harmful. In this way, the execution backbone supplies the behavioral evidence needed by the evolution loop, while the evolution loop supplies compact task specific guidance to the execution backbone.

\subsection{Intra-Task Context Control}
\label{app:intra_task_context_control}

Medical tasks often require multiple tool calls, intermediate observations, and careful synthesis. If the agent simply accumulates the full interaction history, several practical problems may arise. The agent may repeat the same ineffective action, important early evidence may become less visible after many later observations, and long tool outputs may exceed the intended prompt budget. To address these issues, the context guard introduces three mechanisms within a single task: loop breaking, confirmed finding pinning, and budget aware history trimming. These mechanisms do not modify the stored trajectory. Instead, they only adjust how the history is rendered into the next prompt, so the complete execution record remains available for downstream analysis. The loop breaking mechanism examines recent actions before each step. If the same tool action with the same input has been repeated for a configured number of consecutive steps, the context guard appends a notice to the prompt, telling the model that the repeated action has not produced progress and that it should change strategy, use a different tool, or provide a final answer when sufficient evidence is already available. This mechanism is conservative because it only activates on exact repeated action signatures, which allows the agent to reuse the same tool when the inputs are meaningfully different.

The confirmed finding pinning mechanism is designed to preserve important early evidence. Once the task reaches a configured step index, the context guard extracts observations from the early steps and uses a lightweight model call to summarize only confirmed findings. The resulting summary is stored as pinned findings and appended to future prompt histories under a dedicated marker. This summary is generated once per task to limit overhead, and its purpose is not to introduce new information but to keep verified facts visible as the conversation becomes longer. This is especially useful in medical reasoning, where early observations may contain patient attributes, test results, or constraints that should remain available during final synthesis. 

The budget aware history trimming mechanism further controls prompt length. Before each model call, the context guard estimates the token usage of the step history. If the estimate exceeds a configured fraction of the context budget, older observations are shortened while the most recent steps are preserved in fuller form. The rendered history also receives a context trim notice, which tells the model that earlier content has been compressed and that it should rely on pinned findings and recent steps. This differs from ordinary truncation because the complete trajectory remains stored in the agent record. Only the prompt view is compressed, which means downstream trajectory analysis and skill evolution can still access the untrimmed execution trace.

\begin{table}[ht]
\centering
\caption{The three intra-task context control mechanisms used by the agent.}
\begin{tabular}{p{0.24\linewidth}p{0.30\linewidth}p{0.36\linewidth}}
\toprule
\textbf{Mechanism} & \textbf{Trigger} & \textbf{Effect on the next prompt} \\
\midrule
Loop breaking & The same action and input are repeated for the configured number of recent steps. & Adds a warning that asks the model to avoid repeating the action and to change strategy or answer. \\
Confirmed finding pinning & The task reaches the configured pinning step and early observations are available. & Adds a concise confirmed findings block distilled from early observations. \\
Budget aware trimming & The estimated prompt history length exceeds the configured budget ratio. & Shortens older observations, preserves recent steps, and adds a notice explaining that earlier observations were compressed. \\
\bottomrule
\end{tabular}
\label{tab:context_guard_mechanisms}
\end{table}

Together, these mechanisms make the agent loop more stable without changing the external task interface. Loop breaking reduces redundant exploration, confirmed finding pinning protects salient evidence, and budget aware trimming keeps the prompt within a controlled length. Since all three mechanisms are injected as runtime context rather than stored as ordinary user messages, they guide the next decision while preserving the faithfulness of the underlying trajectory. This design is also compatible with the self evolution pipeline because the saved step history still records the original model outputs, tool calls, observations, errors, and final answer for later encoding and utility assessment.

\section{Tool Suite}
\label{app:tool-suite}

Our agent is equipped with a modular tool suite that decomposes clinical problem solving into reusable capabilities. These tools cover six major categories: \textbf{external evidence retrieval, structured medical knowledge access, quantitative clinical computation, multimodal perception, reasoning control, and benchmark-specific interaction}. External evidence retrieval includes general web search and medical evidence retrieval. Structured medical knowledge access provides drug information lookup, drug interaction checking, and biomedical concept lookup. Quantitative clinical computation includes a safe numerical calculator, a medical unit converter, and a clinical score calculator. Multimodal perception separates diagnostic image analysis from OCR and chart reading. Reasoning control supports intermediate reflection and final answer verification. Benchmark-specific tools provide patient dialogue, examination requests, and image access for interactive clinical benchmarks. All tools share a common execution interface: each call receives a structured payload, returns a compact observation, and exposes a parameter schema that can be injected into the agent prompt or requested when needed. This design keeps the action space flexible enough for heterogeneous medical tasks while preserving a uniform execution protocol across datasets. The detailed prompts for tools that involve LLM calls are provided in Appendix~\ref{app:prompts}.

\begin{itemize}
    \item \textbf{General web search tool.}
    The general web search tool is used when the agent needs current or open-domain information, such as recent clinical updates, public medical resources, or general background facts that may not be reliably captured by the model itself. In our implementation, this tool is supported by the Tavily Search API \cite{tavilyai_github}. Given a natural-language query, it retrieves relevant web results and returns a compact synthesized observation for the agent to use in subsequent reasoning.

    \item \textbf{Medical retrieval tool.}
    The medical retrieval tool is used when the agent needs domain-specific medical evidence for diagnosis, treatment, pharmacology, or biomedical reasoning. It provides access to curated medical sources such as PubMed, textbooks, StatPearls, Wikipedia, and combined biomedical corpora. In implementation, the tool uses a MedRAG-style \cite{xiong2024improving} retrieval backend when local corpora are available, and can fall back to public scholarly search or PubMed-based retrieval when needed. Its purpose is to ground the agent's reasoning in medical references rather than relying only on internal model knowledge.

    \item \textbf{Drug information lookup tool.}
    The drug information lookup tool is used when the agent needs medication-specific information, including indications, mechanisms, contraindications, adverse reactions, warnings, and drug classes. It accepts a drug name and returns a concise summary of clinically relevant information. The implementation prioritizes local DrugBank-style resources \cite{wishart2018drugbank} when available, with public sources such as RxNorm and OpenFDA labels used as fallbacks. This tool helps the agent reason about medication choice, adverse effects, and contraindications.

    \item \textbf{Drug interaction checking tool.}
    The drug interaction checking tool is used when a case involves multiple medications or when a proposed treatment must be checked for safety. It takes two or more drug names and reports known pairwise interactions when available. The implementation uses local structured interaction data when present and can fall back to public drug interaction resources \cite{wishart2018drugbank}. This tool is intended to support medication reconciliation, treatment safety assessment, and adverse-effect reasoning.

    \item \textbf{Biomedical concept lookup tool.}
    The biomedical concept lookup tool is used when the agent needs to clarify a disease, symptom, procedure, or clinical term using standardized biomedical knowledge. It can provide definitions, semantic categories, and related concepts. When configured, the tool queries UMLS resources \cite{bodenreider2004unified}, and otherwise falls back to public NLM or MedlinePlus-style sources. This tool is useful for disambiguating clinical terminology and connecting patient-facing descriptions to standardized medical concepts.

    \item \textbf{Safe numerical calculator.}
    The safe numerical calculator is used for arithmetic and formula-based reasoning, especially when exact computation is needed for dosage, rate conversion, numerical comparison, or risk-score components. The implementation evaluates mathematical expressions through a restricted calculator rather than allowing arbitrary code execution. This helps reduce simple numerical errors while keeping the returned observation concise.

    \item \textbf{Medical unit converter.}
    The medical unit converter is used when clinical values or laboratory results are reported in different units. It supports common clinical conversions involving mass, volume, concentration, pressure, temperature, selected analyte-specific conversions, electrolyte equivalents, and HbA1c. The tool takes a value, source unit, target unit, and optionally a substance name for analyte-specific conversion. Its purpose is to normalize values before applying clinical thresholds or scoring rules.

    \item \textbf{Clinical score calculator.}
    The clinical score calculator is used when a task requires a standardized clinical risk score, severity score, or physiological index. It supports commonly used scores such as CHA$_2$DS$_2$-VASc, HAS-BLED, HEART, Wells scores, PERC, CURB-65, SOFA, qSOFA, NEWS2, Child-Pugh, MELD, Glasgow Coma Scale, ABCD$^2$, Glasgow-Blatchford, RCRI, BMI, eGFR, Cockcroft-Gault, and MDRD. The tool computes the score from the provided clinical variables and returns both the result and a brief interpretation. This helps reduce errors in threshold-based clinical decision making.

    \item \textbf{Medical image analysis tool.}
    The medical image analysis tool is used when the task contains clinically meaningful visual content, such as X-ray, CT, MRI, ECG images, pathology slides, fundus photographs, or other medical images. The tool receives a single image and asks a vision-language medical model \cite{jiang2025hulu} to produce structured clinical findings, an impression, and possible differential diagnoses. Its purpose is to support diagnostic reasoning when the image itself contains clinically relevant evidence.

    \item \textbf{OCR and chart reading tool.}
    The OCR and chart reading tool is used for document-like images where the main information is textual, numerical, or tabular. Examples include laboratory reports, medication charts, scanned notes, clinical forms, and tables embedded in images. Unlike the medical image analysis tool, this tool focuses on extracting displayed information \cite{bai2025qwen3} rather than interpreting visual pathology. This separation helps the agent distinguish between reading clinical data and diagnosing from medical images.

    \item \textbf{Reflection tool.}
    The reflection tool is used during intermediate reasoning when the agent needs to check whether its current reasoning is complete, whether additional evidence is needed, or whether important differentials or assumptions have been missed. It provides a compact critique and a suggested next step. In implementation, this tool uses a critic-style model when available, with simpler fallback checks when necessary. Its purpose is to improve reasoning control rather than to generate a separate final answer.

    \item \textbf{Verifier tool.}
    The verifier tool is used near the end of a trajectory to assess a proposed final answer before delivery. It compares the original question and candidate answer from an independent evaluation perspective, focusing on factual correctness, clinical safety, and completeness. This tool separates answer generation from answer checking, allowing the agent to audit its final response before producing it.

\begin{table*}[t]
\caption{Dataset-specific tool availability. A check mark indicates that the corresponding tool is enabled.}
\label{tab:dataset-tool-matrix}
\centering

\small
\setlength{\tabcolsep}{4pt}          
\renewcommand{\arraystretch}{1.25}   

\resizebox{\textwidth}{!}{%
\begin{tabular}{lcccccccccccccccc}
\toprule

\textbf{Dataset} &
\rotatebox{60}{\textbf{Web}} &
\rotatebox{60}{\textbf{Med. Ret.}} &
\rotatebox{60}{\textbf{Drug Info}} &
\rotatebox{60}{\textbf{Drug Int.}} &
\rotatebox{60}{\textbf{Concept}} &
\rotatebox{60}{\textbf{Calc.}} &
\rotatebox{60}{\textbf{Unit}} &
\rotatebox{60}{\textbf{Score}} &
\rotatebox{60}{\textbf{Image}} &
\rotatebox{60}{\textbf{OCR}} &
\rotatebox{60}{\textbf{Reflect}} &
\rotatebox{60}{\textbf{Verify}} &
\rotatebox{60}{\textbf{AC Reply}} &
\rotatebox{60}{\textbf{AC Exam}} &
\rotatebox{60}{\textbf{AC Image}} \\

\midrule

AgentClinic-MM      & \cmark & \cmark & \cmark & \cmark & \cmark & \cmark & \cmark & \cmark & \cmark & \cmark & \cmark & \cmark & \cmark & \cmark & \cmark &  \\
AgentClinic-Text    & \cmark & \cmark & \cmark & \cmark & \cmark & \cmark & \cmark & \cmark &        &        & \cmark & \cmark & \cmark & \cmark &        &  \\

\addlinespace[3pt]   

LiveClin-MM         & \cmark & \cmark & \cmark & \cmark & \cmark & \cmark & \cmark & \cmark & \cmark & \cmark & \cmark & \cmark &        &        &        &  \\
LiveClin-Text       & \cmark & \cmark & \cmark & \cmark & \cmark & \cmark & \cmark & \cmark &        &        & \cmark & \cmark &        &        &        &  \\

\addlinespace[3pt]

LiveMedBench        & \cmark & \cmark & \cmark & \cmark & \cmark & \cmark & \cmark & \cmark &        &        & \cmark & \cmark &        &        &        &  \\

\addlinespace[3pt]

MMMU                & \cmark & \cmark & \cmark & \cmark & \cmark & \cmark & \cmark & \cmark & \cmark & \cmark & \cmark & \cmark &        &        &        &  \\
MMMU-Pro            & \cmark & \cmark & \cmark & \cmark & \cmark & \cmark & \cmark & \cmark & \cmark & \cmark & \cmark & \cmark &        &        &        &  \\

\addlinespace[3pt]

MedJourney          & \cmark & \cmark & \cmark & \cmark & \cmark & \cmark & \cmark & \cmark &        &        & \cmark & \cmark &        &        &        &  \\
MedXpertQA-MM       & \cmark & \cmark & \cmark & \cmark & \cmark & \cmark & \cmark & \cmark & \cmark & \cmark & \cmark & \cmark &        &        &        &  \\
MedXpertQA-Text     & \cmark & \cmark & \cmark & \cmark & \cmark & \cmark & \cmark & \cmark &        &        & \cmark & \cmark &        &        &        &  \\

\addlinespace[3pt]

MediQ               & \cmark & \cmark & \cmark & \cmark & \cmark & \cmark & \cmark & \cmark &        &        & \cmark & \cmark &        &        &        &  \\
HealthBench         & \cmark & \cmark & \cmark & \cmark & \cmark & \cmark & \cmark & \cmark &        &        & \cmark & \cmark &        &        &        &  \\

\bottomrule
\end{tabular}
}
\end{table*}

\item \textbf{AgentClinic patient reply tool.}
    The AgentClinic patient reply tool is used in interactive clinical cases when the doctor-agent needs additional subjective history from the patient \cite{schmidgall2024agentclinic}. The agent asks a question, and the tool returns a first-person patient response based on patient-visible information in the benchmark case. The implementation prevents the simulated patient from revealing hidden diagnoses, objective test results, or information that a real patient would not know. This tool supports natural history taking while preserving benchmark constraints.

    \item \textbf{AgentClinic examination request tool.}
    The AgentClinic examination request tool is used when the doctor-agent needs objective clinical findings, such as vital signs, physical examination findings, laboratory results, imaging reports, or neurological examination results \cite{schmidgall2024agentclinic}. The agent submits a concise examination request, and the tool returns findings that are available in the bound case. If the requested information is not available, the tool reports that limitation instead of inventing new findings. This separates subjective history gathering from objective clinical investigation.

    \item \textbf{AgentClinic image request tool.}
    The AgentClinic image request tool is used in multimodal AgentClinic cases when an associated image is available \cite{schmidgall2024agentclinic}. It retrieves the image path or URL from the benchmark case but does not interpret the image. The agent can then pass the retrieved image to the medical image analysis tool when visual interpretation is needed. This keeps benchmark data access separate from image-based clinical reasoning.



\end{itemize}

\section{Additional Details of SkeMex}
\label{app:method-details}

This section provides an operational summary of SkeMex as introduced in the main text. We use the same M-MDP notation \(T_{M\text{-}MDP}=\langle S,A,P,E,\gamma,M\rangle\), where \(s_t\in S\) is the agent state, \(a_t\in A\) is a reasoning action, tool action, or final answer action, and \(M\) is the external skill memory. Each memory unit is instantiated as a skill \(m_i=(k_i,c_i,u_i)\), where \(k_i\) is the retrieval key, \(c_i\) is reusable procedural content, and \(u_i\) is an estimated utility statistic. At learning window \(w\), the current skill repository is denoted by \(M^{(w)}\), which contains general, task-level, and action-level branches. 
For a task instance \(x_n\), the agent constructs an initial state \(s_0(x_n)\), routes the task to a clinical category \(\kappa_n\), retrieves a compact skill set \(m_n\subseteq M^{(w)}\) using the retrieval distribution \(\mu(m_n\mid s_0(x_n),M^{(w)})\), and then acts with the backbone policy \(p_\theta(a_t\mid s_t,m_n)\). Since SkeMex retrieves skills once at the episode onset, \(m_n\) is kept fixed throughout the trajectory. SkeMex does not update \(\theta\). Instead, post-deployment improvement comes from evolving \(M\). 
After the agent completes the task, the resulting trajectory is written as \(\tau_n=(x_n,\kappa_n,m_n,H_n,\hat{y}_n,y_n,r_n)\), where \(H_n=\{(s_t,a_t,o_t)\}_{t=0}^{T_n-1}\) stores the step history, \(o_t\) is the observation returned by the environment or tools, \(\hat{y}_n\) is the final answer, \(y_n\) is the reference signal when available, and \(r_n\) is the task reward. A trajectory gate keeps only informative traces in the window buffer \(\mathcal{B}^{(w)}\), such as nontrivial successes, useful failures, or trajectories that reveal whether retrieved skills helped or harmed the agent. 
At the end of each learning window, the writing operator \(\mathcal{W}\) distills buffered trajectories into new skills or local patches to existing skills, the valuation operator \(\mathcal{V}\) updates skill utilities from reward and adoption evidence, and the governance operator \(\mathcal{G}\) maintains the repository by merging redundant skills, promoting stable skills, deprecating harmful or low-utility skills, and enforcing branch capacity constraints. Here, \(\mathcal{W}(M^{(w)},\mathcal{B}^{(w)})\) denotes applying the trajectory-level writing process over all gated trajectories in the window, followed by draft review. The overall memory update follows the closed-loop form
\begin{equation}
M^{(w+1)} = \mathcal{G}\Bigl(\mathcal{V}\bigl(\mathcal{W}(M^{(w)}, \mathcal{B}^{(w)})\bigr)\Bigr),
\label{eq:closed_loop2}
\end{equation}
which corresponds to the Read, Write, Assess, and Govern lifecycle described in the main paper.

\begin{algorithm}[ht]
\caption{Closed-loop skill memory evolution in SkeMex}
\label{alg:skemex-closed-loop}
\begin{algorithmic}[1]
\Require Initial skill repository \(M^{(0)}\), task stream \(\mathcal{D}=\{x_n\}_{n=1}^{N}\), retrieval budget \(K\), window size \(L\), governance period \(q\)
\Require Backbone policy \(p_\theta\), retrieval distribution \(\mu\), writing operator \(\mathcal{W}\), valuation operator \(\mathcal{V}\), governance operator \(\mathcal{G}\)
\State Initialize window index \(w\gets 0\) and buffer \(\mathcal{B}^{(w)}\gets\emptyset\)
\For{\(n=1\) to \(N\)}
    \State Construct initial state \(s_0\gets s_0(x_n)\)
    \State Route the task to clinical category \(\kappa_n\)
    \State \textbf{Read:} retrieve skills \(m_n\gets \mathrm{TopK}_{K}\big(\mu(\cdot\mid s_0,M^{(w)})\big)\), using \(\kappa_n\) for category-aware scoring
    \State Inject \(m_n\) into the agent context and keep it fixed for the episode
    \State Initialize step history \(H_n\gets\emptyset\) and time step \(t\gets 0\)
    \While{the episode has not terminated}
        \State Decode action \(a_t\sim p_\theta(\cdot\mid s_t,m_n)\)
        \State Execute \(a_t\) in the environment or tool interface and observe \(o_t\)
        \State Append \((s_t,a_t,o_t)\) to \(H_n\)
        \If{\(a_t\) is a final-answer action}
            \State Terminate the episode
        \Else
            \State Update state \(s_{t+1}\gets \mathrm{Update}(s_t,a_t,o_t)\)
            \State \(t\gets t+1\)
        \EndIf
    \EndWhile
    \State Obtain final answer \(\hat{y}_n\), optional reference \(y_n\), and reward \(r_n\)
    \State Build trajectory \(\tau_n=(x_n,\kappa_n,m_n,H_n,\hat{y}_n,y_n,r_n)\)
    \State \textbf{Write buffer:} if \(\mathrm{Gate}(\tau_n)=1\), add \(\tau_n\) to \(\mathcal{B}^{(w)}\)
    \If{\(|\mathcal{B}^{(w)}|=L\)}
        \State \textbf{Write:} construct candidate memory \(\widehat{M}^{(w)}\gets \mathcal{W}(M^{(w)},\mathcal{B}^{(w)})\)
        \State \textbf{Assess:} update utilities \(\widetilde{M}^{(w)}\gets \mathcal{V}(\widehat{M}^{(w)},\mathcal{B}^{(w)})\)
        \If{\((w+1) \bmod q = 0\)}
            \State \textbf{Govern:} set \(M^{(w+1)}\gets \mathcal{G}(\widetilde{M}^{(w)})\)
        \Else
            \State Set \(M^{(w+1)}\gets \widetilde{M}^{(w)}\)
        \EndIf
        \State \(w\gets w+1\)
        \State Reset buffer \(\mathcal{B}^{(w)}\gets\emptyset\)
    \EndIf
\EndFor
\State \Return evolved skill repository \(M^{(w)}\)
\end{algorithmic}
\end{algorithm}

\section{Dataset Details}
\label{app:dataset_details}

We evaluate SkeMex on nine medical benchmarks that jointly cover interactive clinical decision making, patient-centered clinical reasoning, knowledge-intensive medical question answering, and multimodal medical understanding. The selected benchmarks differ in task format, modality, evaluation protocol, and clinical scope, which allows us to test whether the skill repository can support both text-only and multimodal reasoning across heterogeneous medical settings.

\begin{itemize}
    \item \textbf{AgentClinic.}
    AgentClinic \cite{schmidgall2024agentclinic} is a simulated clinical environment designed for evaluating medical agents in interactive scenarios. The doctor-agent can ask patients questions, collect additional history, request examinations, and use multimodal information before making diagnostic or treatment decisions. We use both the text-only and multimodal settings, which allows us to evaluate whether SkeMex helps in sequential information gathering and image-supported clinical reasoning.

    \item \textbf{LiveClin.}
    LiveClin \cite{wang2026liveclin} is a live clinical benchmark built from contemporary peer-reviewed case reports. It is designed to reduce data leakage and knowledge obsolescence by updating the benchmark over time. Its cases cover realistic clinical pathways and include both text-only and multimodal questions. We use LiveClin to evaluate whether the agent can handle recent, clinically grounded, and often complex patient cases.

    \item \textbf{MedJourney.}
    MedJourney \cite{wu2024medjourney} evaluates model performance across a broad patient journey rather than isolated exam-style questions. It covers multiple stages of healthcare delivery, including planning, access, clinical service, and ongoing care. We include MedJourney because it tests whether the agent can reason over patient-centered clinical workflows and produce contextually appropriate decisions.

    \item \textbf{LiveMedBench.}
    LiveMedBench \cite{yan2026livemedbench} is a contamination-resistant benchmark built from real-world medical cases and evaluated with case-specific rubric criteria. It emphasizes open-ended clinical reasoning, but its rubric structure makes automatic evaluation more reliable than unconstrained free-form judging. We use LiveMedBench to assess whether SkeMex improves clinically complete and patient-specific answers under rubric-based evaluation.

    \item \textbf{HealthBench.}
    HealthBench \cite{arora2025healthbench} evaluates health-related conversations with physician-authored rubrics. It covers diverse healthcare scenarios, including clinical advice, safety, communication, guideline adherence, and user-facing health support. We include HealthBench because it provides a structured way to evaluate open-ended medical responses beyond simple exact-match accuracy.

    \item \textbf{MediQ.}
    MediQ \cite{li2024mediq} focuses on interactive question asking under incomplete clinical information. Instead of forcing the model to answer immediately, the benchmark emphasizes whether the agent can identify missing information and ask useful follow-up questions. We use MediQ to evaluate information-seeking behavior and reliable reasoning under uncertainty.

    \item \textbf{MedXpertQA.}
    MedXpertQA \cite{zuo2025medxpertqa} is an expert-level medical reasoning benchmark spanning multiple specialties and body systems. It contains both text and multimodal subsets, with questions designed to require advanced medical knowledge and multi-step reasoning. We use both subsets to evaluate whether SkeMex can improve difficult specialty-level reasoning and multimodal interpretation.

    \item \textbf{MMMU.}
    MMMU \cite{yue2024mmmu} is a multidisciplinary multimodal benchmark collected from college-level exams, quizzes, and textbooks. We use only the Health \& Medicine track. This subset includes medical images, diagrams, charts, tables, and knowledge-intensive questions, making it useful for testing multimodal medical understanding in an academic setting.

    \item \textbf{MMMU-Pro.}
    MMMU-Pro \cite{yue2025mmmu} strengthens MMMU by filtering out questions that can be answered from text alone, augmenting answer choices, and introducing settings that require stronger visual-textual integration. We use the Health \& Medicine portion to evaluate whether the agent can handle more robust multimodal medical questions where superficial textual shortcuts are less effective.
\end{itemize}

To make skill accumulation informative and evaluation reliable, we apply a unified preprocessing pipeline before offline evolution and testing. This pipeline also helps reduce unnecessary experimental cost by removing samples that are unlikely to provide reusable experience or stable evaluation signals. First, when official difficulty labels, benchmark metadata, or subset annotations are available, we prioritize difficult or reasoning-intensive samples. For datasets without explicit difficulty annotations, we retain cases that require multi-step reasoning, integration of multiple clinical clues, tool use, or interpretation of multimodal evidence. This choice is important because trivial cases often provide limited reusable experience for skill construction.
Second, we remove examples whose query description is inconsistent with the attached visual input. This includes cases where the recorded number of images does not match the prompt, image paths are missing or invalid, or the question refers to a figure although the sample is configured as text-only. Third, we exclude fully open-ended examples that do not provide answer options, reference answers, rubric criteria, or other reliable evaluation signals. Open-ended datasets are retained only when they include structured scoring rubrics or benchmark-provided evaluation criteria. Fourth, we normalize answer formats, option labels, image path fields, and modality tags across datasets so that the agent receives a consistent input format. Finally, we remove near-duplicate samples across skill accumulation and test splits to reduce leakage.
This preprocessing strategy preserves challenging and clinically diverse cases while filtering out samples that are ambiguous, unevaluable, modality-inconsistent, or unlikely to contribute useful reusable skills. It also keeps the experimental cost manageable without weakening the main purpose of the benchmark pool: evaluating whether SkeMex can accumulate and transfer meaningful clinical experience.

After preprocessing, the evaluation pool contains 3,278 samples across 12 dataset configurations derived from the nine benchmarks. AgentClinic, LiveClin, and MedXpertQA are separated into text-only and multimodal configurations because they use different input modalities and tool availability. MMMU and MMMU-Pro are restricted to their Health \& Medicine subsets. Table~\ref{tab:dataset_statistics} reports the resulting data size and category distribution. For compactness, we use the following abbreviations: DD for differential diagnosis, TP for treatment planning, IMG for medical imaging interpretation, GFA for guideline update and formula adjustment, LNC for lifestyle and nutrition counseling, DOC for clinical documentation and protocol adherence, EPI for epidemiologic study design and control selection, DDI for drug interaction and medication safety review, ACAD for non clinical academic problem solving, PVH for preventive vaccination and travel health counseling, PERI for perioperative patient education and counseling, SURG for intraoperative surgical safety and procedure guidance, LIT for literature retrieval, OB for obstetric delivery assistance, GEN for genetic linkage and recombination analysis, BILL for clinical billing and administrative support, SOC for non clinical social interaction, SPAT for spatial reasoning and image navigation, and FLY for fitness to fly and physiologic risk assessment.

\begin{table*}[ht]
\centering
\small
\setlength{\tabcolsep}{4pt}
\renewcommand{\arraystretch}{1.25}

\caption{Dataset size and category distribution after preprocessing.}
\label{tab:dataset_statistics}

\begin{tabular}{p{0.18\textwidth} p{0.08\textwidth} p{0.66\textwidth}}
\toprule
\textbf{Dataset} & \textbf{Samples} & \textbf{Category distribution} \\
\midrule

AgentClinic Text & 214 & DD 214 \\
AgentClinic MM   & 120 & DD 115, TP 2, IMG 3 \\

\addlinespace[3pt]

HealthBench & 350 & DD 93, GFA 57, LNC 67, DOC 49, TP 32, LIT 14, BILL 9, PVH 8, SOC 7, OB 6, DDI 6, IMG 1, PERI 1 \\

\addlinespace[3pt]

LiveClin Text & 205 & DD 78, TP 77, GFA 11, IMG 22, DDI 6, PERI 4, LNC 2, OB 2, PVH 1, LIT 1, DOC 1 \\
LiveClin MM   & 371 & DD 159, IMG 104, TP 69, SURG 13, GFA 8, OB 8, LNC 3, DOC 2, PERI 2, DDI 2, LIT 1 \\

\addlinespace[3pt]

LiveMedBench & 623 & DD 254, TP 247, LNC 41, PVH 22, DDI 21, PERI 16, IMG 7, GFA 5, DOC 3, LIT 2, FLY 2, OB 1, SOC 1, SURG 1 \\

\addlinespace[3pt]

MediQ      & 127 & DD 74, TP 34, GFA 7, PVH 4, IMG 2, DDI 2, DOC 2, FLY 1, BILL 1 \\
MedJourney & 264 & DD 115, TP 102, DDI 16, IMG 13, GFA 8, DOC 5, BILL 2, PVH 2, LNC 1 \\

\addlinespace[3pt]

MedXpertQA Text & 372 & DD 182, TP 145, GFA 8, IMG 7, DOC 7, SURG 5, DDI 5, PVH 3, LNC 2, FLY 2, OB 2, PERI 2, LIT 1, EPI 1 \\
MedXpertQA MM   & 299 & DD 152, TP 99, IMG 36, DDI 4, EPI 4, SURG 3, GFA 1 \\

\addlinespace[3pt]

MMMU     & 277 & IMG 65, DD 65, EPI 61, ACAD 39, GEN 13, GFA 12, SPAT 10, DOC 5, SOC 4, TP 1, DDI 1, LIT 1 \\
MMMU-Pro & 56  & IMG 21, DD 16, ACAD 12, EPI 6, GEN 1 \\

\midrule
\textbf{Total} & \textbf{3,278} & \textbf{19 task categories across text-only, multimodal, interactive, and rubric-based clinical settings} \\
\bottomrule
\end{tabular}
\end{table*}

\clearpage

\section{Baselines \& Metrics}
\label{Baselines & Metrics}

\subsection{Baselines}
\label{app:baselines}

We compare SkeMex with four groups of baselines. The first group consists of medical specialist models, which provide domain-specific reference points for medical reasoning and multimodal medical understanding. The second group contains a memory free tool-using agent, which isolates the effect of the proposed skill memory under the same tool access setting. The third group includes reflection based methods, which reuse prior critiques or self-reflections to support later reasoning. The fourth group contains self improving memory agents, which store, distill, or organize experience across tasks. Unless otherwise specified, all agentic baselines are equipped with the same tool suite as SkeMex when the corresponding benchmark permits tool use. For memory based baselines, we use the same training split for experience accumulation and retrieve stored experience at inference time according to each method's original design or the closest faithful adaptation.

\begin{itemize}
    \item \textbf{Lingshu.}
    Lingshu \cite{xu2025lingshu} is a medical specialist foundation model designed for unified medical understanding and reasoning. It supports both text and multimodal medical tasks, which makes it a useful reference point for evaluating clinical and visual medical reasoning.

    \item \textbf{Hulu-Med.}
    Hulu-Med \cite{jiang2025hulu} is a medical-domain model developed for broad clinical and biomedical reasoning. It serves as another specialist baseline that reflects the performance of adapting the model itself to medical tasks. 

    \item \textbf{MedGemma.}
    MedGemma \cite{sellergren2025medgemma} is a family of medical models for text and image comprehension. We include it as a specialist reference for benchmarks that involve medical knowledge and multimodal understanding.

    \item \textbf{Vanilla ReAct.}
    Vanilla ReAct \cite{yao2022react} is a tool-using agent that interleaves reasoning and actions. We use the same task prompts and the same available tools as SkeMex, but remove external skill memory, skill retrieval, utility estimation, and repository governance. This baseline directly measures how much improvement comes from the evolving skill memory rather than from tool access alone.

    \item \textbf{Reflexion.}
    Reflexion \cite{shinn2023reflexion} improves agents by writing natural-language reflections after receiving feedback. These reflections are reused in later trials to guide behavior. To make it suitable for our cross-task setting, we allow stored reflections to be retrieved for related medical tasks. This baseline tests whether reflective feedback alone is sufficient for reusable clinical improvement.

    \item \textbf{CRITIC.}
    CRITIC \cite{gou2023critic} uses external feedback to critique and revise model outputs. We adapt it by storing previous critiques and retrieving relevant critique records during later tasks. This baseline evaluates whether reusable critique memory can support medical reasoning without explicitly distilling experience into structured skills.

    \item \textbf{Voyager.}
    Voyager \cite{wang2023voyager} is a lifelong agent that accumulates reusable skills from experience. Although it was originally designed for open-ended embodied exploration, it provides a useful reference for direct skill accumulation. In our adaptation, successful medical trajectories are converted into reusable procedures that can be retrieved for similar cases.

    \item \textbf{DILU.}
    DILU \cite{wen2023dilu} stores and reuses prior decision-making experience to guide later actions. We instantiate it as an experience memory baseline in which previous medical reasoning traces are summarized and retrieved for the current task. This comparison tests whether experience reuse without explicit utility guided skill governance can match SkeMex.

    \item \textbf{ExPeL.}
    ExPeL \cite{zhao2024expel} extracts natural-language lessons from prior task experiences and uses them to improve future behavior. It represents an experience distillation baseline that stores compact insights rather than raw trajectories. We include it to compare SkeMex with a method that learns reusable lessons but does not perform adoption-aware utility estimation or repository-level governance.

    \item \textbf{GenerativeMemory.}
    GenerativeMemory \cite{park2023generative} stores observations and experiences, then retrieves and synthesizes them to support later behavior. We adapt it as a global experience memory baseline for medical tasks. This baseline tests whether general episodic memory and reflection can provide the same benefit as the procedural skill abstraction used in SkeMex.

    \item \textbf{Memp.}
    Memp \cite{fang2025memp} improves task solving by retrieving useful past experience and injecting it into the prompt. We include it as a lightweight memory baseline because it directly tests the value of prompt-level memory augmentation. Compared with Memp, SkeMex treats memory as an evolving repository of reusable skills whose utilities are updated from later outcomes.

    \item \textbf{SkillWeaver.}
    SkillWeaver \cite{zheng2025skillweaver} discovers and refines reusable skills from agent experience. It is one of the closer baselines to SkeMex because both methods focus on skill-level experience reuse. The comparison examines whether skill discovery alone is sufficient, or whether value-aware retrieval and memory governance provide additional benefits in heterogeneous medical benchmarks.

    \item \textbf{AgentWorkflowMemory.}
    AgentWorkflowMemory \cite{wang2024agent} stores successful workflows and reuses them when similar tasks appear. We adapt it by mining reusable medical reasoning and tool-use workflows from the training split. This baseline evaluates the benefit of workflow-level memory, while SkeMex further allows skills to be created, patched, valued, merged, or deprecated over time.

    \item \textbf{AgentKB.}
    AgentKB \cite{tang2025agent} builds a knowledge base for agents from previous interactions and task-solving experience. In our experiments, it represents a knowledge-base-oriented memory baseline where stored entries are retrieved to guide later reasoning. We include it to test whether a general agent knowledge base can provide the same benefit as a structured skill repository.

    \item \textbf{Evolver.}
    Evolver \cite{wu2025evolver} is a self improving agent framework that refines behavior from accumulated experience. We include it as a representative evolution-oriented baseline. Its comparison with SkeMex highlights the difference between general experience refinement and our explicit lifecycle for reading, writing, assessing, and governing skill memory.

    \item \textbf{DynamicCheatsheet.}
    DynamicCheatsheet \cite{suzgun2026dynamic} maintains an adaptive cheatsheet that is updated across tasks and reused as persistent test-time memory. It represents a compact global memory baseline. SkeMex differs by retrieving task-relevant skills rather than relying on a single evolving cheatsheet.

    \item \textbf{MobileE.}
    MobileE \cite{wang2025mobile} is a memory based agent method developed for interactive agent settings. We include it to examine whether experience reuse mechanisms designed for interactive environments transfer to medical reasoning benchmarks. In our adaptation, stored experiences are retrieved when they match the current clinical task context.

    \item \textbf{CerebraFusionMemory.}
    CerebraFusionMemory \cite{zhang2025memevolve} integrates multiple memory signals to improve agent behavior. We include it as a stronger global memory baseline because it can combine heterogeneous past information. This comparison evaluates whether memory fusion alone can match SkeMex's explicit utility estimation and repository maintenance.

    \item \textbf{GSEM.}
    GSEM \cite{han2026gsem} is a graph based self evolving memory method for clinical reasoning. It is the closest baseline in spirit because it also structures medical experience memory. We include it to compare SkeMex with a clinical memory evolution method that organizes experience through graph structure, while SkeMex represents reusable knowledge as procedural skills and updates them through outcome based utility assessment.
\end{itemize}

These baselines cover several levels of adaptation. The no-memory setting is represented by Vanilla ReAct. Reflective memory methods include Reflexion and CRITIC. Experience distillation methods include Voyager, DILU, ExPeL, GenerativeMemory, and Memp. Skill and workflow memory methods include SkillWeaver, AgentWorkflowMemory, AgentKB, and Evolver. Global or persistent memory methods include DynamicCheatsheet, MobileE, and CerebraFusionMemory. Graph structured clinical memory is represented by GSEM. In addition, Lingshu, Hulu-Med, and MedGemma serve as medical specialist references. SkeMex is distinguished from these baselines by combining a structured skill repository, value-aware retrieval, category-normalized utility estimation, and periodic memory governance. This design allows the agent not only to reuse previous experience, but also to estimate whether each stored skill remains useful for future clinical tasks.

\subsection{Metrics}
\label{app:metrics}

We use dataset-specific metrics that match the answer format and evaluation protocol of each benchmark. For standard close-ended benchmarks, including most multiple-choice datasets, the primary metric is accuracy. Let \(\mathcal{I}\) denote the set of samples that are successfully evaluated, excluding only failures such as missing inputs or invalid benchmark records. The accuracy is computed as
\begin{equation}
    \mathrm{Acc}
    =
    \frac{1}{|\mathcal{I}|}
    \sum_{i\in\mathcal{I}} c_i ,
\end{equation}
where \(c_i\in\{0,1\}\) is the per-sample correctness indicator. For multiple-choice tasks, we normalize both model predictions and reference answers into uppercase option letters. Let \(Y_i\subseteq\mathcal{O}_i\) be the set of correct options for sample \(i\), where \(\mathcal{O}_i\) is the option set, and let \(\Phi(\hat{y}_i)\subseteq\mathcal{O}_i\) be the set of options parsed from the model response \(\hat{y}_i\). The correctness indicator is
\begin{equation}
    c_i
    =
    \mathbb{I}\{\Phi(\hat{y}_i)=Y_i\}.
\end{equation}
For single-answer questions, this reduces to exact option matching. Common answer formats such as \(A\), \((A)\), \([A]\), \(A.\), and \(A:\) are normalized before comparison. If a response cannot be parsed into a valid option, it is counted as incorrect rather than removed from the denominator. This makes the evaluation deterministic and avoids giving credit to ambiguous answer formats.

For open-ended datasets whose official answers are not simple option letters, such as AgentClinic Text and MedJourney, we use a semantic-equivalence judge. The judge receives the reference answer and the model prediction, then returns a binary verdict indicating whether the prediction is correct or semantically equivalent to the reference. The resulting verdict is used as \(c_i\) in the same accuracy computation above. Thus, these datasets are still reported as accuracy, but correctness is determined by semantic equivalence rather than exact string or option matching.

For HealthBench and LiveMedBench, we use rubric-based scoring because their outputs are free-form clinical responses. Each sample \(i\) contains a list of rubric criteria
\begin{equation}
    \mathcal{C}_i=\{(q_{ij},w_{ij})\}_{j=1}^{J_i},
\end{equation}
where \(J_i\) denotes the number of rubric criteria for sample \(i\), \(q_{ij}\) is the \(j\)-th clinical criterion, and \(w_{ij}\) is its associated weight. We use Gemini-3-Flash \cite{google2025gemini3} to evaluate all criteria for a sample in a single call. The judge returns a binary verdict \(b_{ij}\in\{0,1\}\) for each criterion in the original rubric order. Positive criteria indicate required clinical information or behavior, while negative criteria indicate undesirable content such as unsafe advice, unsupported claims, or clinically inappropriate recommendations. A positive criterion contributes its weight when satisfied. A negative criterion has a negative weight and contributes only when the response commits the specified error.

The normalized rubric score for sample \(i\) is
\begin{equation}
    s_i
    =
    \operatorname{clip}_{[0,1]}
    \left(
    \frac{
    \sum_{j=1}^{J_i} b_{ij} w_{ij}
    }{
    \sum_{j=1}^{J_i} \max(w_{ij},0)
    }
    \right),
\end{equation}
where \(\operatorname{clip}_{[0,1]}(\cdot)\) clamps the value to the interval \([0,1]\). If a sample has no positive-weight criterion, its score is set to \(0\). The benchmark-level rubric score is then computed as
\begin{equation}
    \mathrm{Score}
    =
    \frac{1}{|\mathcal{I}|}
    \sum_{i\in\mathcal{I}} s_i .
\end{equation}
For HealthBench, this is reported as the mean sample score. For LiveMedBench, the same computation is reported as the mean case score.

The same scoring functions are also reused by the memory-evolution pipeline to convert task outcomes into scalar rewards. Accuracy-based datasets provide binary rewards through \(c_i\), while rubric-based datasets provide continuous rewards through \(s_i\in[0,1]\). This keeps the reported benchmark metrics and the Assess-stage utility updates aligned under the same per-sample evaluation rule.

\section{Implementation Details}
\label{app:implementation_details}

\paragraph{Data split.}
We evaluate SkeMex under both in-domain and out-of-domain settings. For benchmarks used in the in-domain evaluation, we split the available samples into approximately equal training and testing partitions. The training split is used only for experience accumulation, skill writing, utility valuation, and repository governance. The corresponding test split is held out for evaluation. For out-of-domain benchmarks, no samples are used during skill accumulation, and the full processed subset is reserved for testing. This protocol allows us to evaluate both whether the skill repository improves performance on related held-out cases and whether the evolved skills can transfer to benchmark families that were not used during repository construction. Table~\ref{tab:implementation-data-split} summarizes the data split used in our experiments.

\begin{table}[t]
\centering
\small
\setlength{\tabcolsep}{6pt}
\renewcommand{\arraystretch}{1.2}

\caption{Dataset statistics and train/test split used in our experiments.}
\label{tab:implementation-data-split}

\begin{tabular}{l S S S c}
\toprule
\textbf{Dataset} & \textbf{Total} & \textbf{Train} & \textbf{Test} & \textbf{Type} \\
\midrule

\multicolumn{5}{l}{\textit{Out-of-domain (OOD)}} \\

AgentClinic Text & 214 & 0 & 214 & OOD \\
AgentClinic MM   & 120 & 0 & 120 & OOD \\
MediQ            & 127 & 0 & 127 & OOD \\
MMMU-Pro         & 56  & 0 & 56  & OOD \\
MedJourney       & 264 & 0 & 264 & OOD \\

\addlinespace[4pt]

\multicolumn{5}{l}{\textit{In-domain (ID)}} \\

MMMU             & 277 & 138 & 139 & ID \\
HealthBench      & 350 & 175 & 175 & ID \\
LiveClin Text    & 205 & 104 & 101 & ID \\
LiveClin MM      & 371 & 186 & 185 & ID \\
LiveMedBench     & 623 & 313 & 310 & ID \\
MedXpertQA Text  & 372 & 187 & 185 & ID \\
MedXpertQA MM    & 299 & 151 & 148 & ID \\

\bottomrule
\end{tabular}
\end{table}

\paragraph{Models and execution setup.}
We use DeepSeek-V3.2 \cite{liu2025deepseek} as the main backbone in the primary experiments. The same backbone is used for agent reasoning, trajectory-to-skill distillation, clinical category classification, utility-related judgment, and repository governance. To test cross-model transfer, we additionally evaluate Qwen3.6-Plus \cite{qwen36plus} with the skill repository accumulated by DeepSeek-V3.2, without re-distilling or rewriting the repository. This setting examines whether the learned skills encode transferable procedural knowledge rather than model-specific response patterns. Semantic indexing of skill items uses text-embedding-3-large \cite{openai2024embedding}. Unless otherwise stated, model calls are executed through API endpoints, while the skill repository, retrieval indices, utility records, and trajectory traces are maintained locally.

\paragraph{Agent execution.}
Each task is executed as a bounded ReAct-style trajectory. At the beginning of an episode, the agent receives the task input, the available tool descriptions, and the retrieved skill snippets. At each step, the agent either continues reasoning, calls a tool with structured arguments, or emits a final answer. Tool outputs are appended to the trajectory as observations and can be used by later steps as well as by the trajectory-to-skill distillation module. We set the maximum number of agent steps to \(7\). If the agent does not produce a final answer within this budget, the trajectory is marked as a maximum-step failure and is excluded from skill writing by the buffer filter. For reproducibility and later analysis, each interaction is saved in a structured JSONL record containing the task input, retrieved skills, intermediate actions, tool observations, final answer, and evaluation result.

\paragraph{Skill retrieval.}
Skill retrieval is performed once at the beginning of each episode. The current task is first assigned to a clinical category, which is used to guide category-aware retrieval. The default retrieval budget is \(K=6\). Candidate skills are scored by combining semantic similarity, utility, and memory strength:
\begin{equation}
    \lambda_{\mathrm{sim}}=0.4,\qquad
    \lambda_{u}=0.4,\qquad
    \lambda_{h}=0.2 .
\end{equation}
We use a minimum similarity threshold of \(0.2\) and apply a \(10\%\) bonus to mature skills. Retrieval is branch-aware, so that general skills, task-level skills, and action-level skills can all contribute to the final retrieved set when available. When the candidate pool is sufficiently large, a lightweight pre-screening step is applied before final ranking. In our configuration, pre-screening is enabled when the number of candidates exceeds \(5\), considers up to \(5\) candidates per branch, and uses a maximum generation budget of \(1024\) tokens.

\paragraph{Skill writing and utility valuation.}
The learning buffer operates over fixed windows. In our implementation, the buffer window size is \(30\) trajectories and the maximum retained buffer capacity is \(20\) trajectories. The buffer removes trajectories that are unlikely to contain reusable information, including simple successes completed within two steps, repetitive loops, and trajectories terminated by the maximum-step limit. At the end of each window, retained trajectories are processed by the encoder, which may propose one of three outcomes: creating a new skill, patching an existing skill, or making no memory update. Newly created skills are initialized with utility \(0.5\) and must pass novelty and quality checks before being inserted into the repository.

Utility valuation uses category-normalized rewards and adoption-aware credit assignment. Positive adoption is scaled by \(\lambda_{+}=1.0\). Negative adoption uses a base penalty \(\lambda_{-}=0.10\) and an additional harm scale \(\lambda_{\mathrm{harm}}=0.5\). Category reward baselines are updated with exponential moving average coefficient \(\alpha=0.2\), require at least \(3\) samples before being treated as reliable, and default to \(0.5\) when no stable estimate is available. Skill utilities are clipped to \([0,1]\). The adoption-count-dependent learning rate follows a warmup and decay schedule, with base learning rate \(0.05\), maximum learning rate \(0.20\), \(5\) warmup steps, and \(20\) decay steps.

\paragraph{Repository governance.}
Repository governance is triggered every \(2\) learning windows. It merges highly similar skills, deprecates consistently low-utility skills, promotes stable high-utility skills to mature status, and enforces branch-wise capacity constraints. The merge similarity threshold is \(0.8\). Skills with utility below \(0.3\) are eligible for deprecation, while skills with utility at least \(0.75\) and usage count at least \(15\) are eligible for mature status. The default capacities are \(12\) general skills, \(8\) task-level skills per category, and \(5\) action-level skills per tool. These limits keep the repository compact and prevent unbounded growth during continual evolution.

\paragraph{Context management.}
We use the context guard described in Appendix~\ref{app:intra_task_context_control} to prevent long trajectories and verbose tool observations from exceeding the model context budget. The default token budget is \(16{,}384\). When the rendered context becomes too long, the system compresses earlier observations while preserving the latest reasoning state, the retrieved skills, and pinned key findings. The trim ratio is \(0.8\). The system keeps the last \(3\) steps in full and compresses older tool observations to at most \(200\) characters. The loop detector uses a repeat threshold of \(3\), and key findings are pinned after \(5\) steps when applicable.

\paragraph{Tool execution and external services.}
The tool suite combines API-based tools and locally served components. General web search is implemented through the Tavily Search API \cite{tavilyai_github}. Medical concept lookup, medication lookup, PubMed-oriented retrieval, reflection, verification, and patient simulation are executed through external APIs or API-compatible endpoints when available. The medical retrieval tool uses a MedRAG-style backend with MedCPT as the retriever, PubMed as the default corpus, top-\(k=3\) evidence retrieval, HNSW indexing, and corpus caching enabled \cite{xiong2024improving}. The multimodal image-analysis tool is served by a locally deployed Hulu-Med-32B endpoint \cite{jiang2025hulu}, while OCR and chart reading use the Qwen-VL-OCR API \cite{bai2025qwen3}. The agent controller communicates with all of these services only through tool calls, so large tool models do not need to run on the same machine as the controller.

\paragraph{Computational resources.}
When LLMs and tool models are accessed through APIs, SkeMex is lightweight on the controller side. A CPU-only machine with \(8\) to \(16\) CPU cores and \(32\)GB memory is sufficient for running the agent loop, maintaining the skill repository, storing trajectory traces, and performing embedding-based retrieval over the skill index. The main memory bottleneck comes from local medical retrieval when corpora, dense retrieval indices, HNSW structures, and corpus caches are loaded. In that setting, we recommend \(64\) to \(128\)GB system memory depending on the number and size of cached corpora. The medical image-analysis tool is deployed as a separate vLLM service \cite{kwon2023efficient} on a single A100 GPU. This deployment keeps the vision-language inference cost isolated from the CPU-side SkeMex controller. The controller only sends image-analysis requests through the tool interface, while repository maintenance, trajectory logging, and embedding-based retrieval remain lightweight CPU-side operations.

\begin{table}[ht]
\centering
\small
\setlength{\tabcolsep}{5pt}
\renewcommand{\arraystretch}{1.2}

\caption{Core hyperparameters of SkeMex used in all main experiments.}
\label{tab:implementation-hyperparameters}

\begin{tabular}{l l c}
\toprule
\textbf{Module} & \textbf{Hyperparameter} & \textbf{Value} \\
\midrule

\multicolumn{3}{l}{\textit{Retrieval}} \\
Number of retrieved skills $K$            &              & 6 \\
Similarity weight $\lambda_{\mathrm{sim}}$ &              & 0.4 \\
Utility weight $\lambda_u$                 &              & 0.4 \\
Memory-strength weight $\lambda_h$         &              & 0.2 \\

\addlinespace[3pt]

\multicolumn{3}{l}{\textit{Memory update}} \\
Learning window size $L$                   &              & 30 \\

\addlinespace[3pt]

\multicolumn{3}{l}{\textit{Utility estimation}} \\
Positive advantage scale $\lambda_{+}$      &              & 1.0 \\
Negative base penalty $\lambda_{-}$         &              & 0.10 \\
Negative harm scale $\lambda_{\mathrm{harm}}$ &            & 0.5 \\
Category EMA coefficient $\alpha$           &              & 0.2 \\

\addlinespace[3pt]

\multicolumn{3}{l}{\textit{Governance}} \\
Governance period                          &              & Every 2 windows \\
Merge similarity threshold                 &              & 0.8 \\
Mature utility threshold                   &              & 0.75 \\

\bottomrule
\end{tabular}
\end{table}

\section{Sensitivity Analysis}
\label{app:sensitivity_analysis}

\paragraph{Overall setup.}
We conduct sensitivity analysis on two rubric based benchmarks, HealthBench and LiveMedBench, to examine whether SkeMex depends strongly on a narrow set of hyperparameter choices. Unless otherwise specified, we use the default configuration with retrieval budget \(K=6\), retrieval weights \((\lambda_{\mathrm{sim}}, \lambda_u, \lambda_h)=(0.4,0.4,0.2)\), learning window size \(L=30\), category baseline update coefficient \(\alpha=0.20\), and maximum utility update step \(\eta_{\max}=0.20\). This default setting obtains 27.65 on HealthBench and 57.95 on LiveMedBench, with an average score of 42.80. In each analysis, we vary one hyperparameter while keeping all others fixed, so that the effect of each design choice can be examined in isolation.

\begin{figure}[ht]
    \centering
    \includegraphics[width=\linewidth]{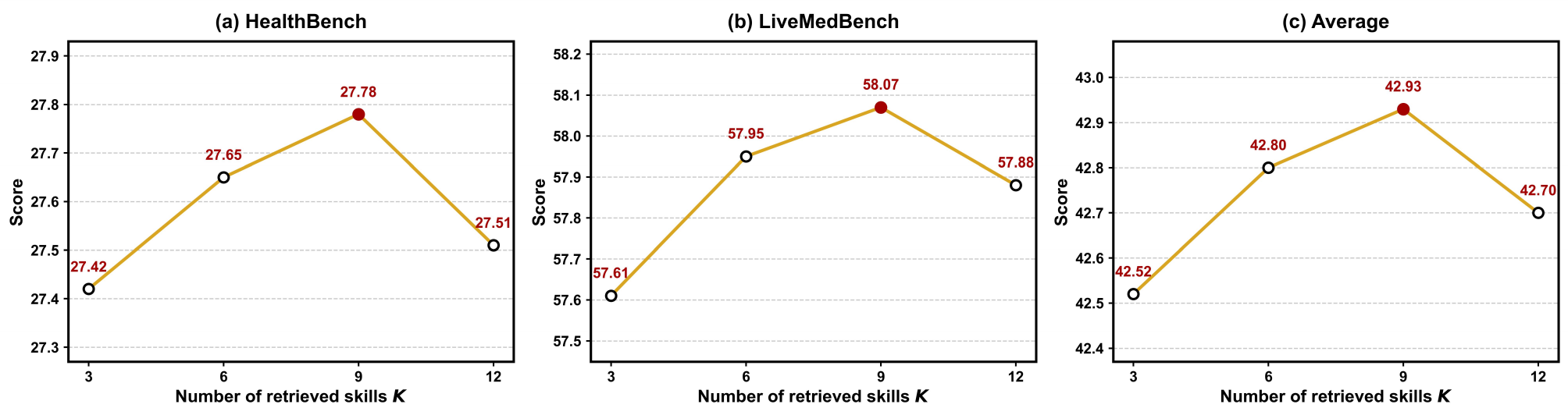}
    \caption{Sensitivity analysis with respect to the number of retrieved skills $K$.}
    \label{fig:sensitivity_k}
\end{figure}

\paragraph{Retrieval budget.}
We first vary the number of retrieved skills \(K\), as shown in Figure~\ref{fig:sensitivity_k}. The average score increases from 42.52 at \(K=3\) to 42.80 at the default value \(K=6\), and reaches 42.93 at \(K=9\). When \(K\) is further increased to 12, the average score slightly decreases to 42.70. This trend suggests that retrieving too few skills may provide insufficient procedural guidance, while retrieving too many skills may introduce redundant or weakly relevant information into the prompt. The overall variation is small across all tested values, which indicates that SkeMex is not highly sensitive to the exact retrieval budget as long as a moderate number of skills is available.

\begin{figure}[ht]
    \centering
    \includegraphics[width=\linewidth]{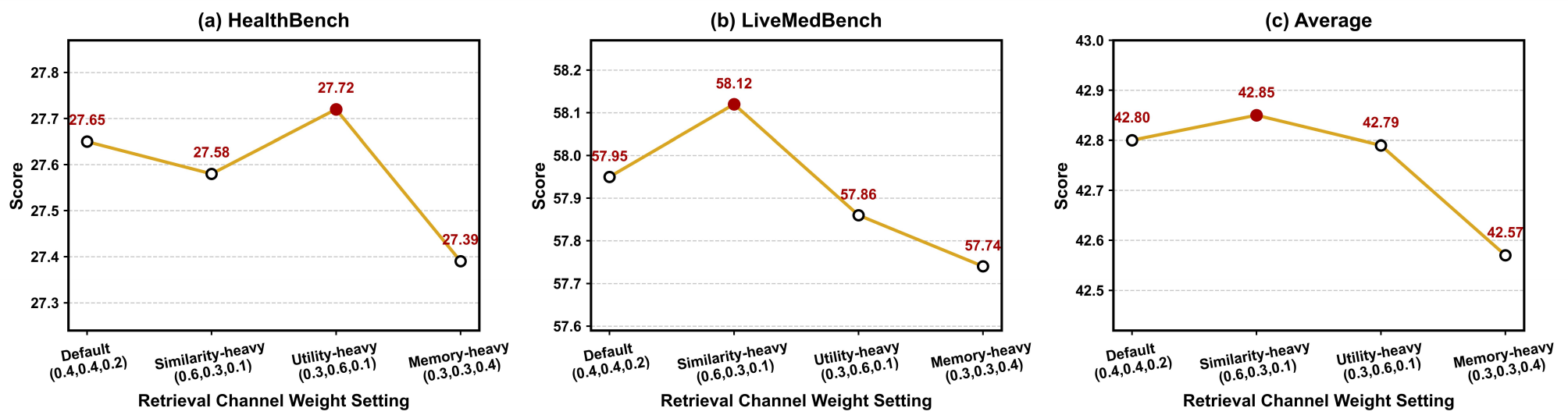}
    \caption{Sensitivity analysis with respect to retrieval channel weights $(\lambda_{\mathrm{sim}}, \lambda_{\mathrm{u}}, \lambda_{\mathrm{h}})$.}
    \label{fig:sensitivity_retrieval_weights}
\end{figure}

\paragraph{Retrieval weights.}
Figure~\ref{fig:sensitivity_retrieval_weights} evaluates how the relative weights of semantic similarity, empirical utility, and memory strength affect retrieval quality. The default setting achieves an average score of 42.80. Increasing the semantic similarity weight gives a slightly higher average score of 42.85, mainly due to the improvement on LiveMedBench. Increasing the utility weight produces a comparable average score of 42.79 and yields the best HealthBench score among the tested settings. By contrast, the memory heavy setting decreases the average score to 42.57. These results suggest that semantic relevance and estimated utility are both important for selecting useful skills, while placing too much emphasis on memory strength alone can reduce retrieval precision by favoring recently reinforced skills that may not be the best match for the current case.

\begin{figure}[ht]
    \centering
    \includegraphics[width=\linewidth]{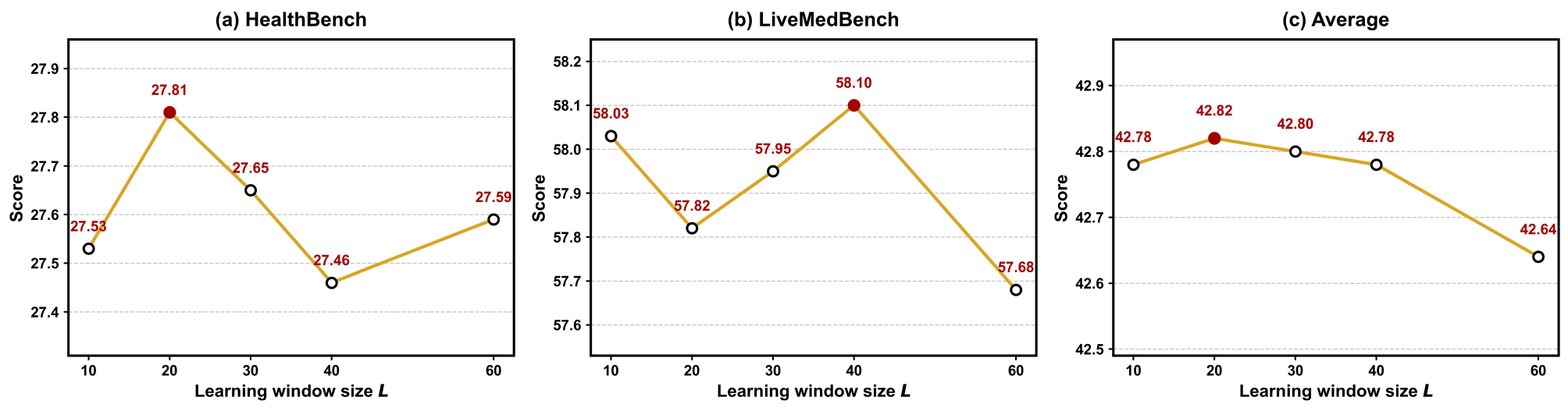}
    \caption{Sensitivity analysis with respect to the learning window size $L$ on rubric-based benchmarks.}
    \label{fig:sensitivity_window}
\end{figure}

\paragraph{Learning window size.}
The effect of the learning window size \(L\) is reported in Figure~\ref{fig:sensitivity_window}. The average scores are 42.78, 42.82, 42.80, 42.78, and 42.64 for \(L \in \{10,20,30,40,60\}\), respectively. The results do not show a monotonic pattern, suggesting that the optimal window size is closely tied to both the amount and the distribution of training data. Smaller windows allow the repository to react more quickly to recent feedback, which can be useful when new trajectories are diverse and informative. However, when the window contains only a small number of samples from each clinical category, utility estimates may become noisy and sensitive to local fluctuations. Larger windows aggregate more trajectories before updating the repository, which can improve stability when the training stream is sufficiently large and balanced, but may slow adaptation when the data distribution shifts or when rare categories are underrepresented. The default value \(L=30\) lies in a stable region and performs nearly the same as the best tested value, indicating a reasonable balance between update stability and responsiveness under our data scale and category distribution.

\begin{figure}[ht]
    \centering
    \includegraphics[width=\linewidth]{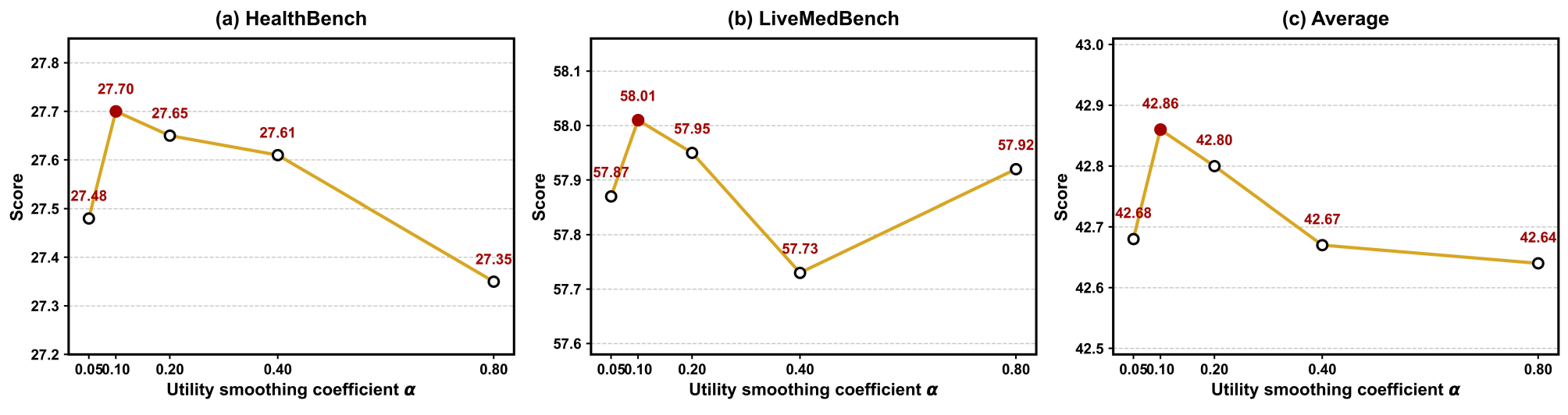}
    \caption{Sensitivity analysis with respect to the utility smoothing coefficient $\alpha$ on rubric-based benchmarks.}
    \label{fig:sensitivity_alpha}
\end{figure}

\paragraph{Category baseline update coefficient.}
Figure~\ref{fig:sensitivity_alpha} studies the coefficient \(\alpha\) used to update category reward baselines. The best average score is obtained at \(\alpha=0.10\), with an average of 42.86, while the default value \(\alpha=0.20\) achieves 42.80. Larger values such as \(\alpha=0.40\) and \(\alpha=0.80\) lead to slightly lower average scores, although the degradation remains limited. Since a larger \(\alpha\) makes the category baseline more responsive to the current window, overly large values may cause the baseline to track short term reward fluctuations too closely. A moderate value therefore provides a better balance between stability and responsiveness.

\begin{figure}[ht]
    \centering
    \includegraphics[width=\linewidth]{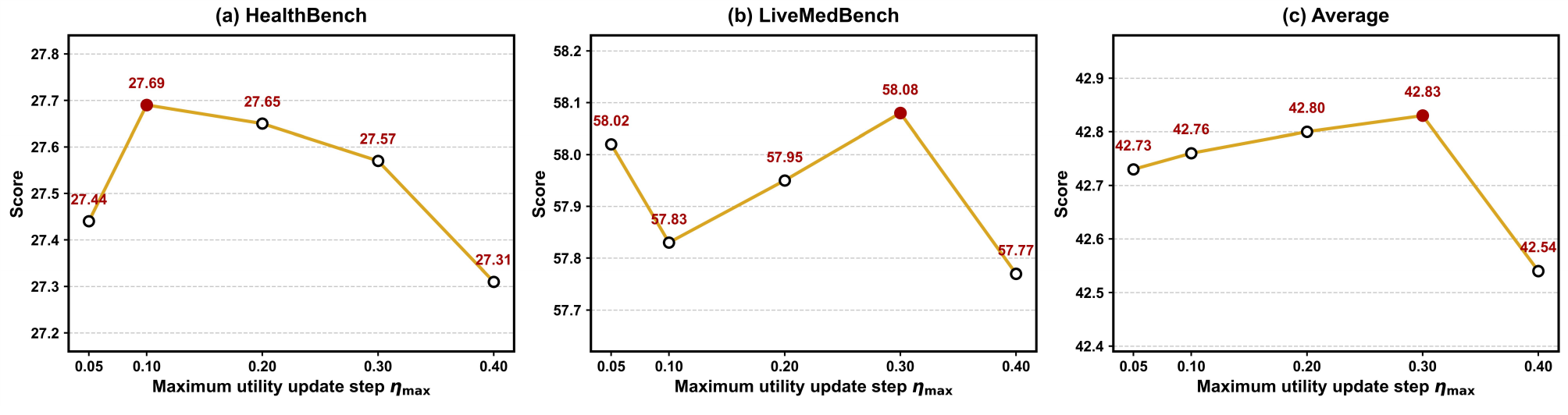}
    \caption{Sensitivity analysis with respect to the maximum utility update step $\eta_{\max}$ on rubric-based benchmarks.}
    \label{fig:sensitivity_eta_max}
\end{figure}

\paragraph{Utility update step.}
Figure~\ref{fig:sensitivity_eta_max} examines the maximum utility update step \(\eta_{\max}\). The average score changes only mildly across different values and reaches 42.83 at \(\eta_{\max}=0.30\), which is close to the default score of 42.80. When the update step is increased to 0.40, the average score drops to 42.54. This suggests that excessively large utility updates can amplify short term feedback noise and make skill valuation less stable. The default setting remains within a robust operating range, avoiding both overly conservative updates and overly reactive utility shifts.

Finally, the sensitivity results show that SkeMex remains stable across a broad range of hyperparameter choices. The largest variations appear when the retrieved skill set is too small, when memory strength is overemphasized in retrieval, or when utility updates become too aggressive. Even in these cases, the performance differences remain modest. These findings suggest that the default configuration provides a balanced trade off among retrieval coverage, retrieval precision, adaptation speed, and utility estimation stability.

\section{Ablation Study}
\label{app:ablation}

\begin{table}[ht]
\raggedleft
\caption{Ablation study on value-aware skill retrieval across different configurations.}
\label{tab:ablation_retrieval}
\setlength{\tabcolsep}{3pt}
\renewcommand{\arraystretch}{1.2}

\resizebox{\linewidth}{!}{
\begin{tabular}{l c c c c c c}
\toprule
\textbf{Setting} 
& \textbf{HealthBench} 
& \textbf{AgentClinic\_T} 
& \textbf{LiveMedBench} 
& \textbf{LiveClin\_M} 
& \textbf{MedXpertQA\_M} 
& \textbf{Avg.} \\
\midrule

w/o Prescreen        & 24.62 & 67.76 & 56.23 & 58.92 & 45.95 & 50.69 \\
w/o Memory Channel   & 26.20 & 67.29 & 54.61 & 58.38 & 47.30 & 50.76 \\
Similarity Only      & 19.26 & 62.62 & 53.90 & 55.14 & 47.30 & 47.64 \\
Utility Only         & 24.65 & 64.02 & 51.74 & 55.14 & 45.95 & 48.30 \\
LLM Ranking Only     & 25.46 & 65.89 & 54.78 & 60.54 & 47.97 & 50.93 \\

\midrule
\textbf{Full (SkeMex)} & \textbf{27.65} & \textbf{68.22} & \textbf{57.95} & \textbf{61.62} & \textbf{50.68} & \textbf{53.22} \\

\bottomrule
\end{tabular}
}
\end{table}

\paragraph{Value-aware skill retrieval.}
Table~\ref{tab:ablation_retrieval} studies how different retrieval signals contribute to selecting useful skills from the repository. The full model achieves the best average score of 53.22\%, showing that effective skill retrieval requires more than surface-level semantic matching. Removing the memory-strength channel lowers the average to 50.76\%, while using LLM ranking alone reaches 50.93\%. These drops indicate that neither a recency-aware memory signal nor structured retrieval scores can be fully replaced by direct LLM reranking. Although LLM ranking can identify seemingly relevant skills, it does not explicitly account for whether a skill has been recently reinforced or has shown stable utility across prior trajectories.

The decline becomes larger when retrieval relies on a single signal. Similarity Only obtains 47.64\%, which is 5.58 points below the full model, and Utility Only obtains 48.30\%, which is 4.92 points lower. This pattern suggests that the two signals capture different aspects of skill usefulness. Semantic similarity can retrieve skills that are topically close to the current case, but such skills may still be ineffective if they have not led to successful outcomes in similar settings. Utility alone can favor historically successful skills, but may select overly general procedures that do not match the current clinical context. The full retrieval score avoids these failure modes by jointly considering semantic relevance, empirical effectiveness, and memory strength.

Prescreening also contributes to retrieval quality. Removing it reduces the average score to 50.69\%, suggesting that early candidate filtering helps remove irrelevant or weakly matched skills before final ranking. This is particularly important in a growing skill repository, where noisy candidates can crowd out more useful skills if all entries are passed directly to the final selection stage. Overall, the results show that value-aware retrieval benefits from a balanced combination of relevance, utility, and memory-strength signals. This combination allows SkeMex to retrieve skills that are not only related to the case, but also empirically reliable and recently validated.

\begin{table}[ht]
\raggedleft
\caption{Ablation study on closed-loop self-evolution memory lifecycle.}
\label{tab:ablation_lifecycle}
\setlength{\tabcolsep}{3pt}
\renewcommand{\arraystretch}{1.2}

\resizebox{\linewidth}{!}{
\begin{tabular}{l c c c c c c}
\toprule
\textbf{Setting} 
& \textbf{HealthBench} 
& \textbf{AgentClinic\_T} 
& \textbf{LiveMedBench} 
& \textbf{LiveClin\_M} 
& \textbf{MedXpertQA\_M} 
& \textbf{Avg.} \\
\midrule

w/o Deprecation        & 23.43 & 67.76 & 56.08 & 55.14 & 43.92 & 49.27 \\
w/o Memory Merging     & 23.78 & 64.02 & 54.08 & 57.84 & 48.65 & 49.67 \\
w/o Capacity Control   & 25.77 & 64.95 & 56.56 & 60.00 & 47.97 & 51.05 \\
w/o Maturation         & 23.17 & 66.36 & 51.18 & 56.76 & 40.54 & 47.60 \\

\midrule
\textbf{Full (SkeMex)}   & \textbf{27.65} & \textbf{68.22} & \textbf{57.95} & \textbf{61.62} & \textbf{50.68} & \textbf{53.22} \\

\bottomrule
\end{tabular}
}
\end{table}

\paragraph{Closed-loop memory lifecycle.}
Table~\ref{tab:ablation_lifecycle} evaluates the contribution of repository governance mechanisms in the closed-loop memory lifecycle. The full SkeMex obtains the highest average score of 53.22\%, which shows that memory evolution requires not only writing new skills but also maintaining the repository after skills have been created. Among the ablations, removing maturation causes the largest drop, reducing the average score to 47.60\%. This 5.62-point decline suggests that the system benefits from distinguishing repeatedly validated skills from newly created or less stable drafts. Without maturation, retrieval may overuse skills that have not yet accumulated enough positive evidence, which weakens long-term reliability.

A substantial degradation is also observed when deprecation is removed. The average score falls to 49.27\%, indicating that obsolete, harmful, or consistently low-utility skills can introduce noise into later retrieval. This result highlights the importance of allowing the repository to forget or down-weight entries that no longer provide useful guidance. Disabling memory merging produces a similar decline, with the average score decreasing to 49.67\%. This shows that redundancy is not only a storage issue, but also a retrieval issue. When highly similar skills remain separate, they can crowd the candidate pool and make it harder for the agent to retrieve diverse and complementary guidance.

Capacity control has a smaller but still meaningful effect. Without capacity limits, the average score drops to 51.05\%, suggesting that unconstrained repository growth gradually weakens retrieval quality even when other lifecycle operations remain active. As more skills accumulate, the repository can become harder to search and more vulnerable to low-value candidates unless branch-wise capacity is regulated. Taken together, these results show that a reliable skill memory depends on the full lifecycle of creation, validation, consolidation, pruning, and capacity control. In SkeMex, governance is therefore not a peripheral cleanup step, but a core component that keeps the evolving repository compact, selective, and clinically dependable.

\clearpage
\section{Further Analyses}
\label{app:Further Analyses}

\subsection{Offline OOD Generalization}
\label{app:offline_ood}

\begin{table*}[ht]
\centering
\caption{Main results on out-of-domain benchmarks in the offline setting. 
All datasets are excluded from training and used solely for evaluation. 
The last column reports the improvement of memory-based methods over the memory-free ReAct baseline, highlighting the benefits of memory. 
\textbf{Bold} numbers indicate the best performance.}
\label{oodtab2}
\setlength{\tabcolsep}{2pt}
\footnotesize
\renewcommand{\arraystretch}{0.5}
\setlength{\extrarowheight}{1pt}

\resizebox{\linewidth}{!}{
\begin{tabular}{m{2cm} m{1.2cm} ccc cc c}
\toprule
\multirow{2}{*}{\centering\textbf{Backbone}} 
& \multirow{2}{*}{\centering\textbf{Method}} 
& \multicolumn{3}{c}{\textbf{Text}} 
& \multicolumn{2}{c}{\textbf{Multimodal}} 
& \multirow{2}{*}{\textbf{Avg.}} \\
\cmidrule(lr){3-5} \cmidrule(lr){6-7}
& & \textbf{MedJourney} & \textbf{MediQ} & \textbf{AgentClinic\_T}
& \textbf{MMMU-Pro} & \textbf{AgentClinic\_M} & \\
\midrule

Hulu-Med-32B & CoT & 70.08 & 89.76 & 22.90 & 46.43 & 84.17 & 62.67 \\[2.5pt]
Lingshu-32B & CoT & 67.42 & 77.95 & 13.55 & 32.14 & 73.33 & 52.88 \\[2.5pt]
MedGemma-27B & CoT & 65.91 & 90.55 & 26.17 & 39.29 & 83.33 & 61.05 \\[2.5pt]
\midrule

\multirow{18}{*}[-60pt]{%
\shortstack[c]{%
\includegraphics[height=3em]{logo1.png}\\[+12pt]
\textbf{DeepSeek-V3.2}%
}%
}
& CoT & 65.15 & 90.55 & 26.64 & 32.14 & 80.83 & 59.06 \\[2.5pt]
& ReAct & 65.53 & 91.34 & 34.11 & 35.71 & 83.33 & 62.01 \\
& Reflexion & 68.94 & 90.55 & 60.28 & 32.14 & 84.17 & \avgdelta{67.22}{+5.21} \\
& CRITIC & 70.08 & 90.55 & 63.55 & 35.71 & 80.00 & \avgdelta{67.98}{+5.97} \\

\cmidrule(lr){2-8}
& Voyager & 73.11 & 90.55 & 59.81 & 32.14 & 79.17 & \avgdelta{66.96}{+4.95} \\
& DILU & 73.86 & 90.55 & 64.95 & 28.57 & 79.17 & \avgdelta{67.42}{+5.41} \\
& ExPeL & 67.42 & 91.34 & 57.94 & 33.93 & 82.50 & \avgdelta{66.63}{+4.62} \\
& GM & 72.73 & 91.34 & 66.36 & 37.50 & 83.33 & \avgdelta{70.25}{+8.24} \\
& Memp & 74.24 & 88.98 & 62.15 & 35.71 & 84.17 & \avgdelta{69.05}{+7.04} \\

\cmidrule(lr){2-8}
& SkillWeaver & 71.97 & 89.76 & 65.42 & 33.93 & 84.17 & \avgdelta{69.05}{+7.04} \\
& AWM & 73.86 & 91.34 & 63.55 & 37.50 & 79.17 & \avgdelta{69.08}{+7.07} \\
& Agent KB & 73.86 & 89.76 & 65.42 & 32.14 & 84.17 & \avgdelta{69.07}{+7.06} \\
& Evolver & \best{78.79} & 91.34 & 65.89 & 35.71 & 80.00 & \avgdelta{70.35}{+8.34} \\

\cmidrule(lr){2-8}
& DC & 73.86 & 88.98 & 64.49 & 37.50 & 83.33 & \avgdelta{69.63}{+7.62} \\
& MobileE & 74.24 & 88.19 & 65.42 & 35.71 & 83.33 & \avgdelta{69.38}{+7.37} \\
& CFM & 74.24 & 88.19 & 64.02 & 37.50 & 84.17 & \avgdelta{69.62}{+7.61} \\
& GSEM & 73.48 & 93.70 & 21.03 & 42.86 & 88.33 & \avgdelta{63.88}{+1.87} \\

\cmidrule(lr){2-8}
& SkeMex & 76.52 & \best{96.85} & \best{68.22} & \best{48.21} & \best{89.17} & \avgdelta{\best{75.79}}{\best{+13.78}} \\

\midrule

\multirow{18}{*}[-60pt]{%
\shortstack[c]{%
\includegraphics[height=3em]{logo2.png}\\[+12pt]
\textbf{Qwen3.6-Plus}%
}%
}
& ReAct & 68.56 & 92.13 & 34.58 & 41.07 & 85.00 & 64.27 \\[1.5pt]
& Reflexion & 79.17 & 96.06 & 62.15 & 44.64 & 90.00 & \avgdeltacolor{74.40}{+10.13}{black} \\
& CRITIC & 79.92 & 96.06 & 61.21 & 42.86 & 86.67 & \avgdeltacolor{73.35}{+9.08}{black} \\

\cmidrule(lr){2-8}
& Voyager & 79.92 & 95.28 & 64.02 & 44.64 & 90.00 & \avgdeltacolor{74.77}{+10.50}{black} \\
& DILU & 78.79 & 95.28 & 64.02 & 46.43 & 89.17 & \avgdeltacolor{74.74}{+10.47}{black} \\
& ExPeL & 79.17 & 95.28 & 64.49 & 44.64 & 89.17 & \avgdeltacolor{74.55}{+10.28}{black} \\
& GM & 78.79 & 93.70 & 63.08 & 46.43 & 89.17 & \avgdeltacolor{74.23}{+9.96}{black} \\
& Memp & 78.41 & 96.85 & 63.08 & 41.07 & 90.83 & \avgdeltacolor{74.05}{+9.78}{black} \\

\cmidrule(lr){2-8}
& SkillWeaver & 79.92 & 96.85 & 63.08 & 41.07 & 88.33 & \avgdeltacolor{73.85}{+9.58}{black} \\
& AWM & 79.92 & 96.85 & 64.02 & 44.64 & 88.33 & \avgdeltacolor{74.75}{+10.48}{black} \\
& Agent KB & 79.55 & 96.85 & 64.02 & 44.64 & 89.17 & \avgdeltacolor{74.84}{+10.57}{black} \\
& Evolver & 79.17 & 96.06 & 63.55 & 46.43 & 90.83 & \avgdeltacolor{75.21}{+10.94}{black} \\

\cmidrule(lr){2-8}
& DC & 78.79 & 96.06 & 55.14 & 44.64 & 90.83 & \avgdeltacolor{73.09}{+8.82}{black} \\
& MobileE & 78.41 & 96.85 & 64.02 & 46.43 & 89.17 & \avgdeltacolor{74.97}{+10.70}{black} \\
& CFM & 79.92 & 96.85 & 60.75 & 41.07 & 90.00 & \avgdeltacolor{73.72}{+9.45}{black} \\
& GSEM & 80.30 & 96.85 & 29.91 & 39.29 & 89.17 & \avgdeltacolor{67.10}{+2.83}{black} \\

\cmidrule(lr){2-8}
& SkeMex & \best{81.44} & \best{97.64} & \best{65.89} & \best{51.79} & \best{94.17} & \avgdeltacolor{\best{78.18}}{\best{+13.91}}{black} \\

\bottomrule
\end{tabular}
}
\end{table*}

Table~\ref{oodtab2} reports the complete offline out-of-domain results for both backbones, complementing Figure~\ref{fig3} where only the DeepSeek-V3.2 results are visualized due to space constraints. In this setting, all evaluation datasets are excluded from skill-repo construction, and the frozen repo is directly transferred to unseen benchmark families. SkeMex achieves the best average performance on both backbones. With DeepSeek-V3.2, SkeMex improves ReAct from 62.01\% to 75.79\%, yielding a \textbf{+13.78} point gain and outperforming the strongest competing memory baseline by 5.44 points. With Qwen3.6-Plus, SkeMex improves ReAct from 64.27\% to 78.18\%, yielding a \textbf{+13.91} point gain and a 2.97-point lead over the strongest competing memory method. These results show that the offline skill repo transfers consistently across model families, rather than benefiting only a single backbone.

The dataset-level results further support this conclusion. Under DeepSeek-V3.2, SkeMex obtains the best score on MediQ, AgentClinic-Text, MMMU-Pro, and AgentClinic-MM, with especially large gains over ReAct on AgentClinic-Text and MMMU-Pro by +34.11 and +12.50 points, respectively. Under Qwen3.6-Plus, SkeMex achieves the best score on all five out-of-domain benchmarks, including MedJourney, MediQ, AgentClinic-Text, MMMU-Pro, and AgentClinic-MM. The largest improvements over ReAct again appear on AgentClinic-Text, MMMU-Pro, and AgentClinic-MM, with gains of +31.31, +10.72, and +9.17 points, respectively. In contrast, several memory baselines exhibit less stable transfer. For example, GSEM obtains only +1.87 and +2.83 average gains under DeepSeek-V3.2 and Qwen3.6-Plus, and falls below ReAct on AgentClinic-Text in both cases. Overall, the cross-backbone results indicate that structured skill abstraction and utility-guided retrieval help reduce negative transfer and preserve robust generalization to unseen medical benchmark families.

\subsection{Cross Backbone Generalization}
\label{app:cross_backbone_generalization}

We further examine whether SkeMex remains effective when the underlying backbone model changes. This analysis tests whether the learned skill repository captures reusable medical experience across model families, rather than exploiting response patterns specific to a single backbone. We evaluate three additional backbones, Qwen3.6-Max-Preview \cite{qwen36_max_preview}, Kimi-2.6 \cite{moonshot2026kimiK26}, and GLM-5.1 \cite{zeng2026glm}, on four representative benchmarks: HealthBench, MMMU-Pro, AgentClinic, and MedXpertQA. For each backbone, we compare SkeMex with ReAct and three memory based baselines under the same evaluation protocol. The three memory based baselines are selected as the strongest self-evolving memory methods from their respective categories in our main experiments.

\begin{figure}[ht]
\centering
\includegraphics[width=\linewidth]{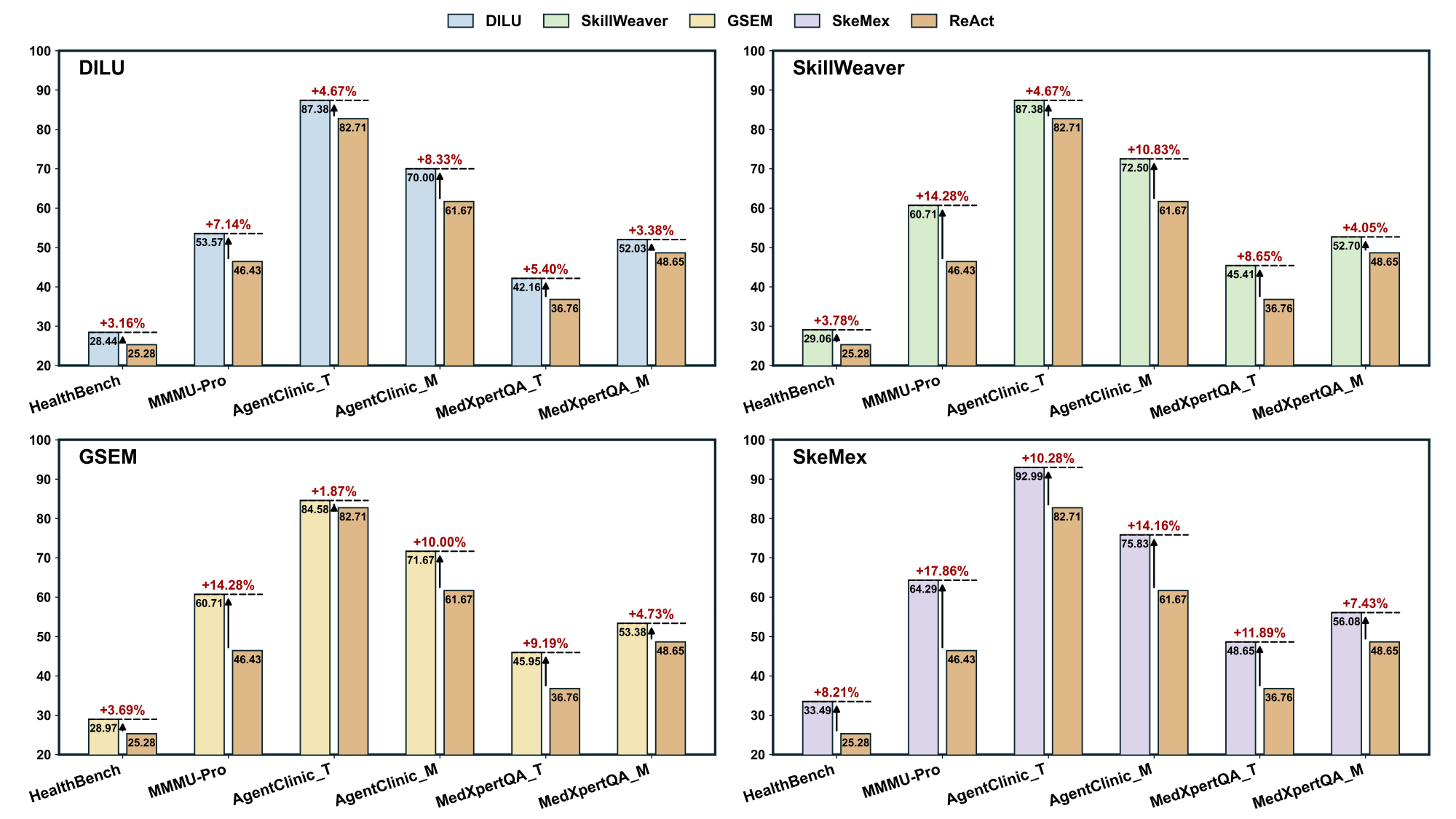}
\caption{Cross backbone generalization results on Qwen3.6-Max-Preview.}
\label{fig:backbone_qwen36_max_preview}
\end{figure}

\begin{figure}[ht]
\centering
\includegraphics[width=\linewidth]{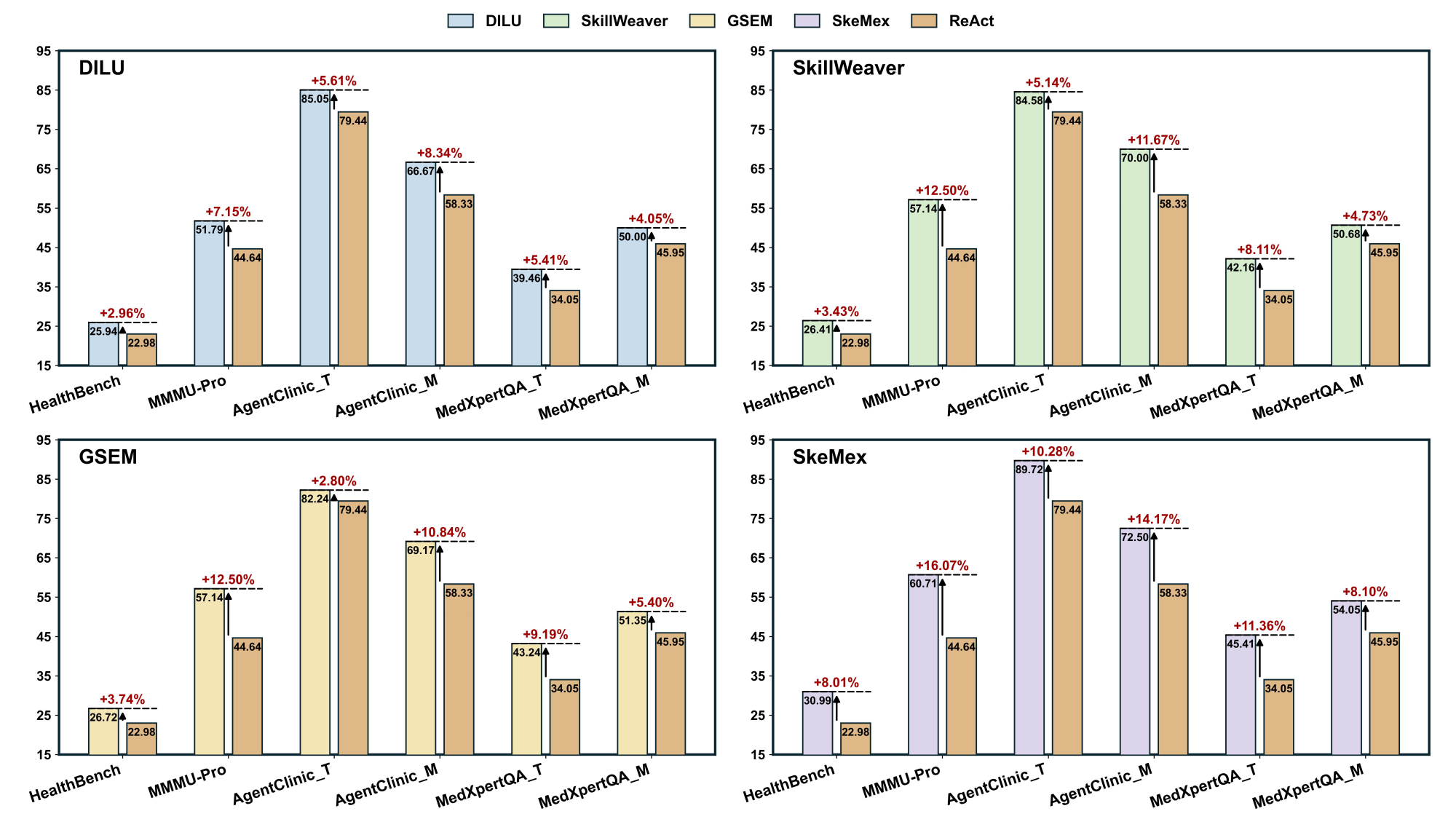}
\caption{Cross backbone generalization results on Kimi-2.6.}
\label{fig:backbone_kimi26}
\end{figure}

\begin{figure}[ht]
\centering
\includegraphics[width=\linewidth]{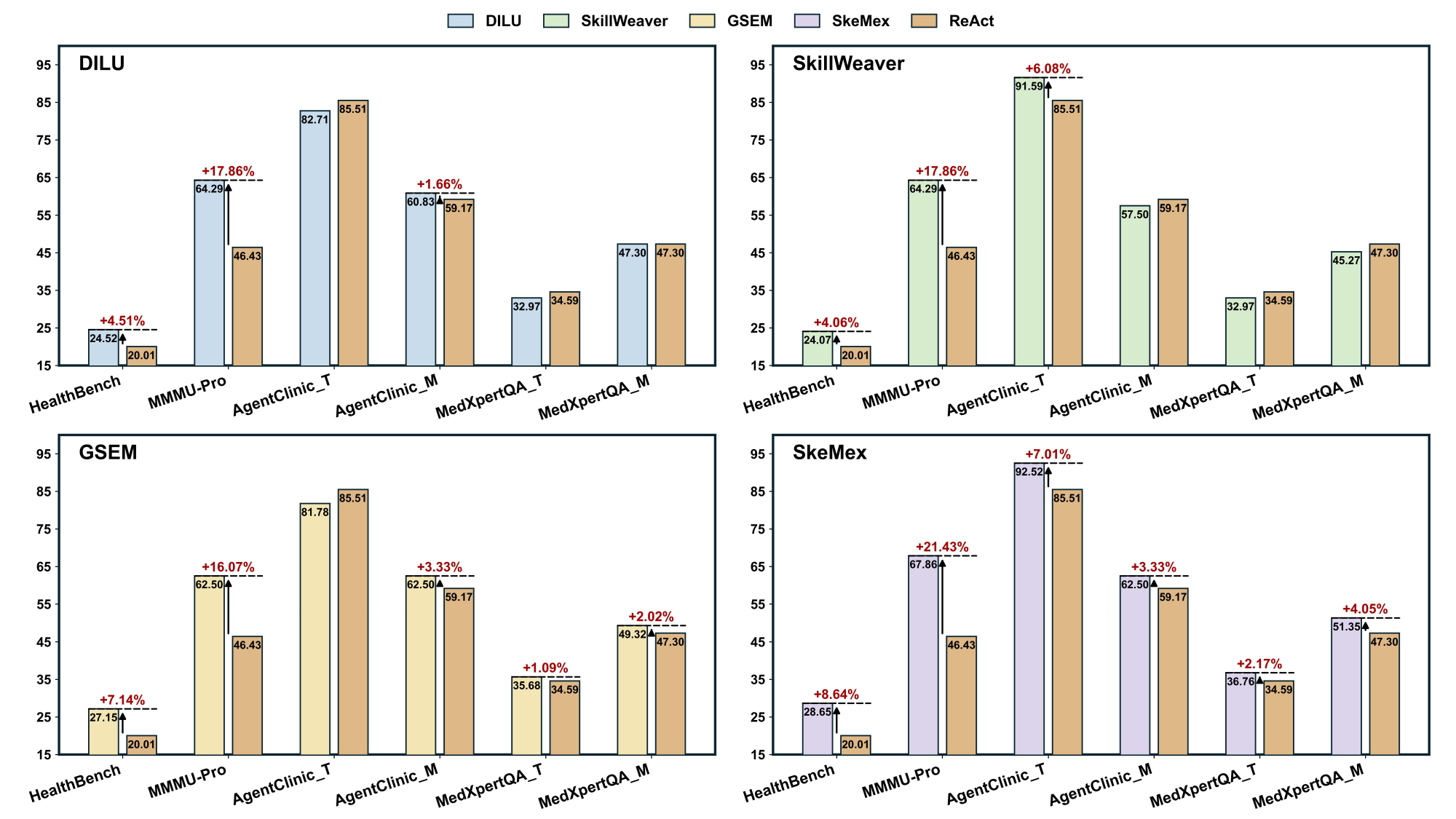}
\caption{Cross backbone generalization results on GLM-5.1.}
\label{fig:backbone_glm51}
\end{figure}

Figures~\ref{fig:backbone_qwen36_max_preview}, \ref{fig:backbone_kimi26}, and~\ref{fig:backbone_glm51} show that the advantage of SkeMex is consistent across all three additional backbones. On Qwen3.6-Max-Preview, SkeMex achieves the best result on every benchmark. It improves over ReAct by 8.21 points on HealthBench, 17.86 points on MMMU-Pro, 10.28 points on AgentClinic-Text, 14.16 points on AgentClinic-MM, 11.89 points on MedXpertQA-Text, and 7.43 points on MedXpertQA-MM. The average score increases from 50.25\% for ReAct to 61.89\% for SkeMex, giving an average gain of \textbf{11.64} points. SkeMex also outperforms the strongest competing memory baseline by 3.93 points on average, which suggests that its gain cannot be attributed to memory augmentation alone.

The results on Kimi-2.6 follow a similar pattern. SkeMex again obtains the best score on all six benchmarks, improving over ReAct by 8.01 points on HealthBench, 16.07 points on MMMU-Pro, 10.28 points on AgentClinic-Text, 14.17 points on AgentClinic-MM, 11.36 points on MedXpertQA-Text, and 8.10 points on MedXpertQA-MM. Its average score reaches 58.90\%, compared with 47.56\% for ReAct, which corresponds to an average gain of \textbf{11.33} points. The close agreement between Qwen3.6-Max-Preview and Kimi-2.6 indicates that SkeMex provides stable improvements across both general clinical reasoning benchmarks and multimodal clinical evaluation benchmarks.

GLM-5.1 provides a more challenging setting because competing memory baselines are less stable across datasets. DILU falls below ReAct on AgentClinic-Text and MedXpertQA-Text. SkillWeaver falls below ReAct on AgentClinic-MM, MedXpertQA-Text, and MedXpertQA-MM. GSEM falls below ReAct on AgentClinic-Text. In contrast, SkeMex produces nonnegative gains on all six datasets. It improves over ReAct by 8.64 points on HealthBench, 21.43 points on MMMU-Pro, 7.01 points on AgentClinic-Text, 3.33 points on AgentClinic-MM, 2.17 points on MedXpertQA-Text, and 4.05 points on MedXpertQA-MM. SkeMex achieves the best result on five benchmarks and ties for the best result on AgentClinic-MM. Its average score improves from 48.84\% for ReAct to 56.61\%, giving an average gain of \textbf{7.77} points.

Across the eighteen backbone and dataset pairs, SkeMex achieves the best result in seventeen cases and ties for the best result in the remaining case. When all three backbones are pooled together, SkeMex improves the average score from 48.88\% to 59.13\%, yielding a \textbf{10.25} point gain over ReAct. The averaged dataset-wise gains are also consistently positive, with 8.29 points on HealthBench, 18.45 points on MMMU-Pro, 9.19 points on AgentClinic-Text, 10.55 points on AgentClinic-MM, 8.47 points on MedXpertQA-Text, and 6.53 points on MedXpertQA-MM. These results indicate that SkeMex does not depend on a particular backbone model. Instead, the structured skill abstraction and utility-guided retrieval provide transferable experience that can be reused by diverse medical agents while preserving stable empirical gains.

\subsection{Cross Backbone Skill Transfer}
\label{app:skill_transfer_backbones}

We further examine whether the skills learned by SkeMex are specific to the backbone that produced them. In this analysis, we use the skill repository constructed from the DeepSeek-V3.2 training split in the main experiments and keep it fixed during evaluation. We then replace the test-time backbone with Claude Sonnet-4.6 \cite{anthropic_claude_sonnet} and Qwen3.6-35B-A3B \cite{qwen36_35b_a3b}. We choose these two models to make the transfer setting more representative and challenging. Claude Sonnet-4.6 is a strong closed-source frontier model, while Qwen3.6-35B-A3B is a strong open-source mixture-of-experts backbone. Evaluating both allows us to test whether the same skill repository can transfer across different model families, deployment regimes, and architectural designs. This setting is stricter than building a separate repository for each target model, because the target backbone must consume and apply skills generated by another model family. It therefore directly tests whether SkeMex learns transferable medical problem solving procedures rather than backbone specific response traces.

\begin{table*}[ht]
\centering
\caption{
Generalization of skills learned by DeepSeek-V3.2 to other backbone models.
LC, MXQA, HB, and LMB denote LiveClin, MedXpertQA, HealthBench, and LiveMedBench, respectively.
\textbf{Bold} numbers indicate the best result for each target backbone and dataset.
}
\label{tab:skill_transfer_backbones_full}

\small
\setlength{\tabcolsep}{3.5pt}
\renewcommand{\arraystretch}{1.15}

\resizebox{\textwidth}{!}{%
\begin{tabular}{
>{\centering\arraybackslash}m{2.8cm}
>{\centering\arraybackslash}m{1.6cm}
cccc
ccc
@{\hspace{8pt}}
>{\centering\arraybackslash}m{1.2cm}
}
\toprule
\multirow{2}{*}{\textbf{Backbone}} 
& \multirow{2}{*}{\textbf{Method}} 
& \multicolumn{4}{c}{\textbf{Text}} 
& \multicolumn{3}{c}{\textbf{Multimodal}} 
& \multirow{2}{*}{\textbf{Avg.}} \\
\cmidrule(lr){3-6} \cmidrule(lr){7-9}
& 
& \textbf{LC} 
& \textbf{MXQA} 
& \textbf{HB} 
& \textbf{LMB}
& \textbf{LC} 
& \textbf{MXQA} 
& \textbf{MMMU} 
& \\
\midrule

\multirow{5}{*}{\textbf{Claude Sonnet-4.6}}
& ReAct       & 81.99 & 36.29 & 24.63 & 47.88 & 61.84 & 48.15 & 46.32 & 49.59 \\
& GM          & 86.99 & 43.41 & 29.41 & 49.69 & 72.16 & 55.58 & 58.60 & 56.55 \\
& SkillWeaver & 86.99 & 44.39 & 28.46 & 51.15 & 72.65 & 52.25 & 60.24 & 56.59 \\
& CFM         & 90.99 & 44.93 & 29.28 & 51.02 & 70.98 & 52.30 & 59.22 & 56.96 \\
& SkeMex      & \textbf{92.04} & \textbf{47.49} & \textbf{32.79} & \textbf{54.92} & \textbf{75.94} & \textbf{55.83} & \textbf{62.87} & \textbf{60.27} \\

\midrule

\multirow{5}{*}{\textbf{Qwen3.6-35B-A3B}}
& ReAct       & 81.37 & 33.64 & 23.38 & 45.88 & 59.34 & 46.20 & 44.17 & 47.71 \\
& GM          & 85.09 & 40.76 & 28.01 & 47.59 & 69.46 & \textbf{54.03} & 56.30 & 54.46 \\
& SkillWeaver & 85.09 & 41.99 & 26.71 & 49.05 & 70.20 & 50.55 & 57.74 & 54.48 \\
& CFM         & \textbf{91.17} & 42.78 & 27.58 & 49.12 & 68.73 & 50.65 & 57.07 & 55.30 \\
& SkeMex      & 90.04 & \textbf{45.49} & \textbf{30.99} & \textbf{53.12} & \textbf{73.34} & 53.78 & \textbf{60.87} & \textbf{58.23} \\

\bottomrule
\end{tabular}%
}
\end{table*}

Table~\ref{tab:skill_transfer_backbones_full} reports the full dataset-level results. When the fixed DeepSeek-V3.2 skill repository is reused by Claude Sonnet-4.6, SkeMex achieves the best score on all seven benchmarks. The average score improves from 49.59\% with ReAct to 60.27\%, giving a \textbf{10.68} point gain. SkeMex also exceeds CFM, the strongest competing memory baseline, by 3.31 points on average. The gains over ReAct are broadly distributed across benchmarks, including 10.05 points on LiveClin-Text, 11.20 points on MedXpertQA-Text, 8.16 points on HealthBench, 7.04 points on LiveMedBench, 14.10 points on LiveClin-MM, 7.68 points on MedXpertQA-MM, and 16.55 points on MMMU. These improvements cover text-only clinical reasoning, rubric-based evaluation, and multimodal medical tasks, indicating that Claude Sonnet-4.6 can interpret and apply skills produced by a different backbone.

The results on Qwen3.6-35B-A3B show a similar transfer effect under a different model architecture. SkeMex improves the average score from 47.71\% with ReAct to 58.23\%, corresponding to a \textbf{10.52} point gain. It also outperforms CFM, the strongest competing baseline on average, by 2.93 points. At the dataset level, SkeMex ranks first on MedXpertQA-Text, HealthBench, LiveMedBench, LiveClin-MM, and MMMU. It is slightly below CFM on LiveClin-Text by 1.13 points and slightly below GM on MedXpertQA-MM by 0.25 points, while still achieving the strongest overall average. More importantly, its gains over ReAct remain positive on every benchmark: 8.67 points on LiveClin-Text, 11.85 points on MedXpertQA-Text, 7.61 points on HealthBench, 7.24 points on LiveMedBench, 14.00 points on LiveClin-MM, 7.58 points on MedXpertQA-MM, and 16.70 points on MMMU. The large gains on LiveClin-MM and MMMU suggest that transferred skills are particularly helpful when the target model needs to integrate heterogeneous clinical evidence or perform multimodal reasoning.

These results provide direct evidence that SkeMex skills can transfer across backbones. The skill repository is created by DeepSeek-V3.2, but its benefits persist when consumed by both a closed-source frontier model and an open-source mixture-of-experts model. This makes it unlikely that the repository only stores cached answers or backbone-specific response traces. Instead, the learned skills appear to capture reusable clinical procedures, such as decomposing patient information, identifying relevant evidence, selecting appropriate tools, and applying domain-specific reasoning heuristics. This interpretation is supported by the consistent gains across text-only, rubric-based, and multimodal benchmarks.
This analysis complements the cross-backbone generalization study in Appendix~\ref{app:cross_backbone_generalization}. That study evaluates SkeMex after changing the backbone under each target setting, while this study fixes the skill source to DeepSeek-V3.2 and changes only the model that consumes the skills. The strong results in both settings suggest that SkeMex separates accumulated medical experience from the particular model that produced it. This portability is important for medical agent systems, because experience accumulated by one capable backbone can be reused to improve other backbones without rebuilding the entire skill repository from scratch.

\subsection{Execution Cost and Interaction Depth}
\label{app:execution_analysis}

We analyze the execution behavior of different methods from two perspectives: the average number of interaction steps and the wall clock time required to complete each task. Table~\ref{tab:steps_table} and Table~\ref{tab:duration_table} report the detailed results for all methods evaluated with the DeepSeek-V3.2 backbone. The goal of this analysis is to understand how skill memory changes the agent's problem solving process, especially whether the performance gain comes with additional reasoning depth or runtime overhead. Since the measured time is affected by API latency, external service availability, and tool response speed, these numbers should be interpreted as approximate wall clock runtimes rather than exact computational complexity.

\paragraph{Interaction depth.}
A clear pattern in Table~\ref{tab:steps_table} is that SkeMex uses more interaction steps than the other methods. Its average number of steps is 4.77, compared with 3.17 for the memory free ReAct baseline and 3.97 to 4.61 for other memory based methods. This higher interaction depth reflects a more deliberate solving process. Retrieved skills often encourage the agent to decompose the task, verify intermediate evidence, use tools when necessary, and avoid premature final answers. This behavior is especially useful for difficult medical cases, where the correct answer often depends on combining multiple pieces of clinical evidence rather than reacting to the most salient clue.
The effect is particularly visible on challenging benchmarks. On MMMU-Pro, ReAct terminates after only 1.482 steps on average, which suggests that it often answers quickly without sufficient exploration. SkeMex increases the average number of steps to 4.625 and improves performance from 35.71\% to 48.21\%, as shown in Table~\ref{oodtab2}. A similar pattern appears on LiveMedBench, where SkeMex takes 6.139 steps compared with 3.578 steps for ReAct. These examples suggest that the additional steps are not merely redundant actions. Instead, they correspond to more structured clinical reasoning, evidence gathering, and answer verification guided by the retrieved skills.

\begin{table*}[ht]
\centering
\caption{
Average number of interaction steps across datasets.
AC, LC, MXQA, HB, LMB, MJ, MQ, and MP denote AgentClinic, LiveClin, MedXpertQA, HealthBench, LiveMedBench, MedJourney, MediQ, and MMMU-Pro, respectively.
Suffixes ``\_T'' and ``\_M'' indicate text-only and multimodal settings.
Methods are grouped following the same categories as in the main offline comparison table.
}
\label{tab:steps_table}
\setlength{\tabcolsep}{2pt}
\footnotesize
\renewcommand{\arraystretch}{1.2}
\setlength{\extrarowheight}{1pt}

\resizebox{\linewidth}{!}{
\begin{tabular}{
m{1.65cm}
ccccc
ccccccc
@{\hspace{6pt}}
>{\centering\arraybackslash}m{0.8cm}
}
\toprule
\multirow{2}{*}{\centering\textbf{Method}}
& \multicolumn{5}{c}{\textbf{Out-of-domain}}
& \multicolumn{7}{c}{\textbf{In-domain}}
& \multirow{2}{*}{\textbf{Avg.}} \\
\cmidrule(lr){2-6} \cmidrule(lr){7-13}
& \textbf{AC\_M} & \textbf{AC\_T} & \textbf{MJ} & \textbf{MQ} & \textbf{MP}
& \textbf{LC\_M} & \textbf{LC\_T} & \textbf{LMB} & \textbf{MMMU} & \textbf{MXQA\_M} & \textbf{MXQA\_T} & \textbf{HB}
& \\
\midrule

ReAct      & 3.650 & 4.397 & 2.789 & 2.591 & 1.482 & 3.785 & 3.635 & 3.578 & 1.897 & 3.385 & 3.692 & 3.120 & 3.17 \\[2.5pt]
Reflexion  & 4.017 & 5.780 & 3.042 & 2.756 & 5.500 & 5.000 & 4.129 & 5.597 & 4.698 & 4.745 & 4.573 & 3.411 & 4.44 \\
CRITIC     & 4.242 & 5.590 & 3.625 & 2.622 & 4.179 & 5.232 & 4.139 & 6.265 & 3.353 & 5.236 & 4.681 & 3.594 & 4.40 \\

\cmidrule(lr){1-14}
Voyager    & 4.483 & 5.795 & 3.205 & 2.756 & 3.946 & 5.124 & 4.238 & 5.922 & 3.309 & 3.939 & 4.443 & 3.469 & 4.22 \\
DILU       & 3.267 & 5.005 & 2.576 & 2.898 & 5.429 & 4.022 & 3.792 & 3.116 & 4.698 & 4.155 & 5.086 & 3.629 & 3.97 \\
ExPeL      & 3.567 & 5.888 & 3.303 & 2.906 & 4.661 & 5.032 & 4.277 & 5.529 & 3.576 & 5.054 & 5.049 & 3.114 & 4.33 \\
GM         & 3.733 & 5.505 & 3.330 & 2.937 & 4.464 & 3.692 & 3.832 & 6.052 & 3.324 & 3.892 & 4.676 & 3.257 & 4.06 \\
Memp       & 4.725 & 5.808 & 3.436 & 2.701 & 5.571 & 5.038 & 4.109 & 6.039 & 4.827 & 4.912 & 4.649 & 3.514 & 4.61 \\

\cmidrule(lr){1-14}
SkillWeaver & 3.758 & 5.603 & 3.144 & 2.622 & 4.143 & 4.984 & 4.139 & 6.000 & 3.439 & 4.980 & 4.319 & 3.377 & 4.21 \\
AWM         & 4.392 & 5.626 & 3.239 & 2.937 & 5.482 & 5.227 & 4.178 & 5.852 & 4.813 & 4.169 & 4.530 & 3.291 & 4.48 \\
Agent KB    & 4.792 & 5.603 & 3.687 & 2.780 & 5.321 & 4.751 & 3.812 & 5.672 & 4.741 & 3.327 & 4.054 & 3.806 & 4.36 \\
Evolver     & 3.475 & 5.626 & 3.379 & 2.772 & 4.125 & 3.984 & 4.020 & 5.784 & 3.137 & 3.939 & 4.578 & 3.189 & 4.00 \\

\cmidrule(lr){1-14}
DC          & 4.600 & 5.659 & 3.701 & 2.976 & 5.339 & 3.896 & 4.307 & 6.090 & 4.763 & 3.784 & 4.649 & 3.337 & 4.43 \\
MobileE     & 3.592 & 5.210 & 3.193 & 2.795 & 5.625 & 3.816 & 4.149 & 5.364 & 4.748 & 4.027 & 4.584 & 3.366 & 4.21 \\
CFM         & 3.533 & 5.491 & 1.148 & 2.567 & 4.054 & 4.076 & 4.356 & 3.326 & 3.583 & 3.899 & 4.476 & 3.269 & 3.65 \\
GSEM        & 4.817 & 5.332 & 3.220 & 3.031 & 5.196 & 3.973 & 4.337 & 5.797 & 4.468 & 3.838 & 4.768 & 3.754 & 4.38 \\

\cmidrule(lr){1-14}
SkeMex      & 4.167 & 6.145 & 4.208 & 3.094 & 4.625 & 5.184 & 5.396 & 6.139 & 4.223 & 4.973 & 5.784 & 3.337 & 4.77 \\

\bottomrule
\end{tabular}
}
\end{table*}

\paragraph{Runtime overhead.}
The deeper interaction process also increases wall clock time. As shown in Table~\ref{tab:duration_table}, SkeMex takes 116.06 seconds per task on average, which is higher than ReAct at 54.48 seconds and Evolver at 81.65 seconds. To better understand this overhead, we also compare the average time per step. SkeMex takes approximately 24.33 seconds per step, while ReAct takes 17.19 seconds and Evolver takes 20.41 seconds. The larger per step cost mainly comes from two sources. First, retrieved skills are rendered into the agent context, which increases the input length of model calls. Second, skill retrieval requires embedding based matching and multi channel scoring at the beginning of each episode. These operations are lightweight compared with model inference, but they still add overhead to the full trajectory.

\paragraph{Accuracy and efficiency.}
Although SkeMex requires more steps and longer runtime, the additional cost is accompanied by consistent accuracy gains across both in domain and out of domain benchmarks. In clinical reasoning settings, this trade off is meaningful because reliability is often more important than raw response speed. The results suggest that SkeMex spends additional computation on useful intermediate reasoning rather than unproductive loops. The context guard and buffer filter also reduce the chance that repeated or failed trajectories are reinforced into memory.

The overhead is not uniform across datasets. On HealthBench, SkeMex reduces the average time to 52.36 seconds, which is lower than ReAct at 64.19 seconds and also below the overall method average of 65.05 seconds, while maintaining a comparable number of steps at 3.337. This indicates that skill memory can sometimes streamline execution when the retrieved skills are highly relevant to the task. In such cases, the agent may avoid slow unguided exploration and move more directly toward the required clinical criteria. Overall, the execution analysis shows that SkeMex generally trades some runtime efficiency for a more structured and empirically grounded reasoning process, while in certain settings the retrieved skills can also improve efficiency by making the trajectory more focused.

\begin{table*}[ht]
\centering
\caption{
Average time consumption in seconds across datasets.
Due to API latency and occasional service instability, the measured time may contain small fluctuations and should be interpreted as approximate wall-clock runtime.
AC, LC, MXQA, HB, LMB, MJ, MQ, and MP denote AgentClinic, LiveClin, MedXpertQA, HealthBench, LiveMedBench, MedJourney, MediQ, and MMMU-Pro, respectively.
Suffixes ``\_T'' and ``\_M'' indicate text-only and multimodal settings.
Methods are grouped following the same categories as in the main offline comparison table.
}
\label{tab:duration_table}
\setlength{\tabcolsep}{2pt}
\footnotesize
\renewcommand{\arraystretch}{1.2}
\setlength{\extrarowheight}{1pt}

\resizebox{\linewidth}{!}{
\begin{tabular}{
m{1.65cm}
ccccc
ccccccc
@{\hspace{6pt}}
>{\centering\arraybackslash}m{0.9cm}
}
\toprule
\multirow{2}{*}{\centering\textbf{Method}}
& \multicolumn{5}{c}{\textbf{Out-of-domain}}
& \multicolumn{7}{c}{\textbf{In-domain}}
& \multirow{2}{*}{\textbf{Avg.}} \\
\cmidrule(lr){2-6} \cmidrule(lr){7-13}
& \textbf{AC\_M} & \textbf{AC\_T} & \textbf{MJ} & \textbf{MQ} & \textbf{MP}
& \textbf{LC\_M} & \textbf{LC\_T} & \textbf{LMB} & \textbf{MMMU} & \textbf{MXQA\_M} & \textbf{MXQA\_T} & \textbf{HB}
& \\
\midrule

ReAct      & 68.85 & 109.75 & 64.26 & 50.52 & 46.32 & 18.04 & 18.02 & 18.03 & 18.02 & 84.31 & 93.51 & 64.19 & 54.48 \\[2.5pt]
Reflexion  & 101.20 & 133.15 & 51.74 & 41.65 & 78.97 & 89.76 & 95.97 & 112.10 & 61.49 & 80.04 & 81.84 & 54.71 & 81.88 \\
CRITIC     & 64.35 & 124.50 & 74.74 & 38.41 & 77.92 & 85.28 & 76.01 & 153.10 & 62.54 & 128.38 & 93.86 & 72.36 & 87.62 \\

\cmidrule(lr){1-14}
Voyager    & 92.02 & 115.46 & 50.58 & 53.70 & 95.89 & 122.13 & 116.01 & 124.17 & 72.13 & 110.28 & 107.90 & 84.22 & 95.37 \\
DILU       & 45.41 & 81.91 & 41.41 & 55.87 & 103.55 & 78.32 & 67.89 & 60.20 & 79.15 & 108.08 & 121.47 & 76.83 & 76.67 \\
ExPeL      & 51.39 & 85.36 & 35.59 & 47.09 & 90.19 & 101.96 & 76.39 & 73.99 & 60.05 & 77.16 & 100.25 & 44.35 & 70.32 \\
GM         & 55.10 & 119.69 & 71.20 & 66.08 & 81.80 & 70.61 & 67.31 & 142.81 & 50.09 & 91.54 & 97.10 & 54.09 & 80.62 \\
Memp       & 57.56 & 91.65 & 59.54 & 44.01 & 202.33 & 104.59 & 77.78 & 148.91 & 119.40 & 95.70 & 113.38 & 78.52 & 99.45 \\

\cmidrule(lr){1-14}
SkillWeaver & 70.70 & 109.13 & 55.66 & 49.26 & 105.43 & 117.61 & 110.68 & 150.23 & 73.23 & 101.86 & 107.63 & 77.99 & 94.12 \\
AWM         & 106.60 & 133.54 & 62.46 & 54.00 & 74.46 & 86.69 & 86.43 & 143.74 & 65.06 & 94.20 & 80.78 & 53.88 & 86.82 \\
Agent KB    & 74.51 & 94.03 & 61.86 & 43.23 & 79.72 & 100.74 & 92.00 & 86.24 & 67.39 & 66.56 & 73.85 & 82.41 & 76.88 \\
Evolver     & 56.58 & 131.53 & 69.06 & 57.66 & 98.88 & 71.58 & 76.48 & 134.09 & 52.34 & 94.41 & 83.27 & 53.87 & 81.65 \\

\cmidrule(lr){1-14}
DC          & 68.81 & 113.66 & 70.83 & 56.74 & 99.90 & 80.29 & 83.41 & 160.93 & 70.97 & 69.08 & 86.45 & 58.41 & 84.96 \\
MobileE     & 47.03 & 88.66 & 59.01 & 56.61 & 102.43 & 71.00 & 70.45 & 106.11 & 78.06 & 100.23 & 100.33 & 69.43 & 79.11 \\
CFM         & 77.36 & 119.06 & 12.42 & 37.10 & 79.97 & 80.07 & 108.89 & 61.45 & 56.04 & 69.85 & 76.97 & 54.15 & 69.44 \\
GSEM        & 64.35 & 55.26 & 49.30 & 35.01 & 66.86 & 95.58 & 116.25 & 132.16 & 65.19 & 89.82 & 105.72 & 74.15 & 79.14 \\

\cmidrule(lr){1-14}
SkeMex      & 100.27 & 116.27 & 84.64 & 58.77 & 132.15 & 172.66 & 130.33 & 159.91 & 91.22 & 129.88 & 164.28 & 52.36 & 116.06 \\

\bottomrule
\end{tabular}
}
\end{table*}

\clearpage
\subsection{Impact of Training Data Order}
\label{app:training_order}

This experiment examines whether the order of training samples during offline skill evolution affects the final in-domain performance of SkeMex. We conduct this analysis on HealthBench and LiveMedBench. For each benchmark, SkeMex evolves its skill repository on the corresponding training split and is then evaluated on the held-out in-domain test split. These two benchmarks are rubric-based, so their sample-level scores provide a useful proxy for task difficulty. For each training sample, we use the score obtained by the DeepSeek-V3.2 ReAct-style agent as the difficulty estimate. Higher ReAct scores indicate easier cases, while lower scores indicate harder cases. We compare four ordering strategies. Random is the default setting, where training samples are shuffled before offline skill evolution. Category-based ordering groups samples by task category and processes larger categories first, allowing the repository to first encounter high-frequency task types. Easy-to-Hard ordering sorts samples by descending ReAct score, so simpler cases appear earlier. Hard-to-Easy ordering sorts samples by ascending ReAct score, forcing the repository to process difficult cases at the beginning.

\begin{table}[ht]
\centering
\caption{Impact of training data order on SkeMex performance using the DeepSeek-V3.2 backbone. 
Results are reported on HealthBench and LiveMedBench. 
\textbf{Bold} numbers indicate the best performance.}
\label{tab:train_order}

\small
\setlength{\tabcolsep}{8pt}
\renewcommand{\arraystretch}{1.18}

\begin{tabular}{lccc}
\toprule
\textbf{Strategy} 
& \textbf{HealthBench} 
& \textbf{LiveMedBench} 
& \textbf{Avg.} \\
\midrule

\rowcolor{gray!10}
Random (Default) 
& \textbf{31.42} 
& \textbf{58.63} 
& \textbf{45.03} \\

Category-based   
& 29.33 
& 58.17 
& 43.75 \\

Easy-to-Hard     
& 30.89 
& 57.50 
& 44.20 \\

Hard-to-Easy     
& 28.00 
& 54.81 
& 41.41 \\

\bottomrule
\end{tabular}
\end{table}

Table~\ref{tab:train_order} shows that random ordering gives the best overall result, reaching an average score of 45.03\%. This suggests that a mixed stream of categories and difficulty levels is beneficial for skill evolution. Since the repository is updated across learning windows, random shuffling helps each window contain a more balanced set of trajectories. This reduces the risk that early memory updates are dominated by a narrow task type or an unusually difficult subset. Among the structured curricula, Easy-to-Hard performs the best, with an average score of 44.20\% and the second-best HealthBench score of 30.89\%. This indicates that starting from easier cases can help the system extract stable and reusable patterns before encountering more complex examples. However, its lower LiveMedBench score suggests that an overly smooth curriculum may delay exposure to difficult clinical reasoning patterns that are needed for robust skill construction.

The Hard-to-Easy strategy performs worst, with an average score of 41.41\%. Processing the most difficult cases first can make early skill writing more vulnerable to noisy trajectories, incomplete reasoning, or overly specific error patterns. Since early memory updates influence later retrieval and governance, low-quality initial skills may have a lasting effect on the repository. Category-based ordering achieves a competitive score on LiveMedBench but performs worse on HealthBench. One possible reason is that grouping by category improves local consistency within a window, but it also reduces diversity and may cause the repository to overemphasize high-frequency task types before seeing rarer categories. Overall, the results support the use of random ordering as the default strategy. It provides a simple and robust way to expose SkeMex to diverse clinical patterns throughout offline evolution, which helps maintain a balanced and generalizable skill repository.

\section{Limitations}
\label{app:Limitations}
While our work presents a structured framework for skill-based medical experience evolution and shows consistent improvements across benchmarks, several limitations remain. First, the evaluated benchmarks cannot fully capture the complexity of real clinical environments, where patient histories, institutional workflows, and decision constraints are often more diverse and less standardized.
Second, although we evaluate SkeMex with several mainstream backbone models, we do not exhaustively cover all available foundation models due to experimental cost.
Third, SkeMex introduces additional inference overhead. Retrieved skills increase prompt length, and the agent may take more reasoning before answering. This leads to higher API usage and longer wall-clock time than memory-free agents. Such cost is common in skill-based and memory-augmented agent methods, which trade some efficiency for more structured and reliable reasoning.

From a societal perspective, the proposed framework may help improve the consistency of medical reasoning systems and reduce repeated errors, but it may also reinforce incorrect patterns or be misused in high-stakes settings without sufficient human oversight. Therefore, it should be viewed as a decision-support and research tool rather than a substitute for professional medical judgment.

\section{Case Study}
\label{app:Case Study}

We present five representative cases to illustrate the behavior of SkeMex across diverse clinical scenarios. Cases 1--4 demonstrate successful skill-guided reasoning, while Case 5 is a failure case that reveals a recurring limitation. In each figure, the \textbf{purple block} highlights the skill(s) retrieved from the skill repository and injected into the agent's context; the subsequent reasoning blocks show how the injected skill shapes the agent's decision-making trajectory.

\begin{figure}[ht]
\centering
\includegraphics[width=\linewidth]{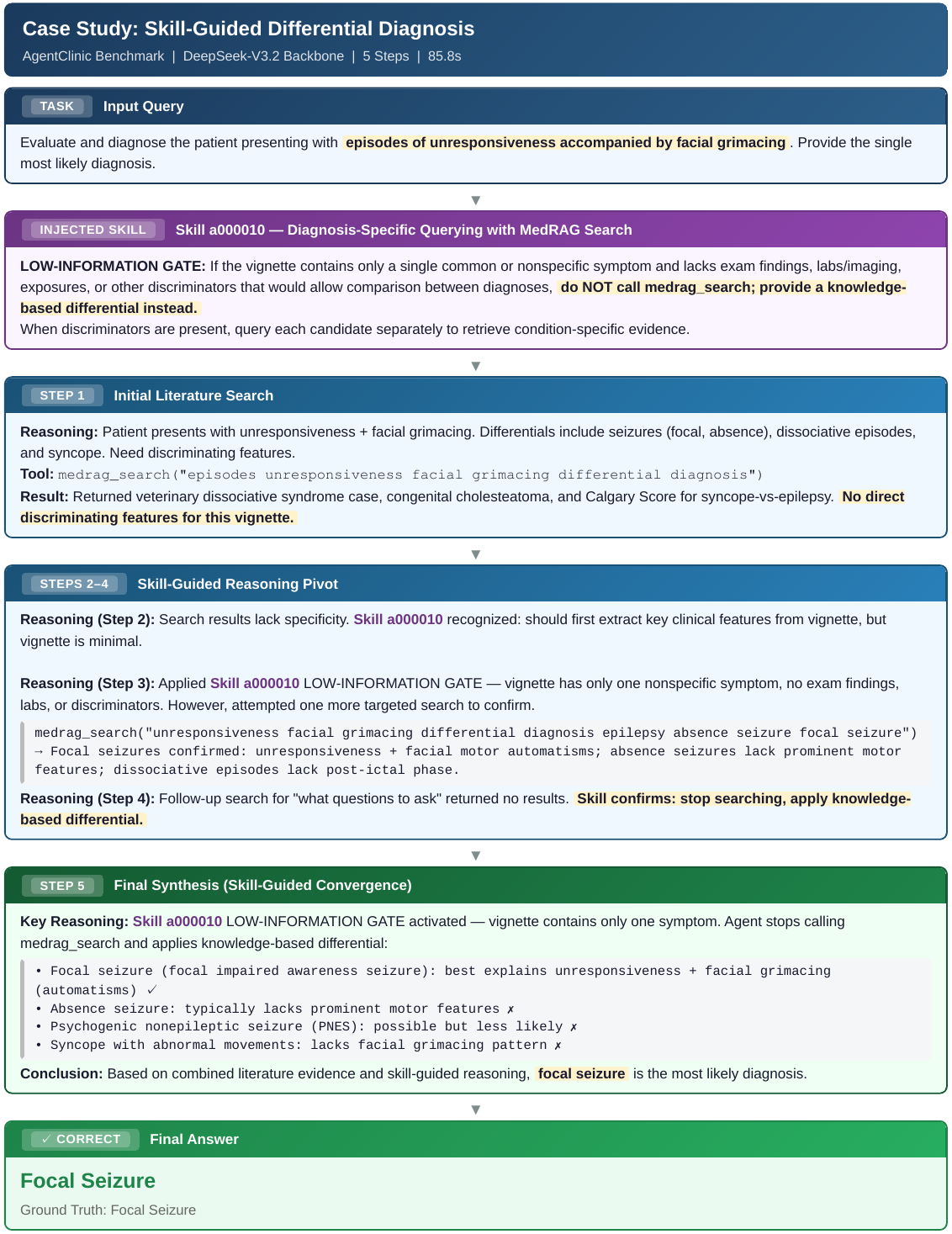}
\caption{
\textbf{Case 1: Skill-guided avoidance of uninformative search loops (AgentClinic).}
The agent is tasked with diagnosing a patient presenting with episodic unresponsiveness and facial grimacing.
Skill \texttt{a000010} (\textit{Diagnosis-Specific Querying with MedRAG Search}) activates a \textit{low-information gate}: when the clinical vignette lacks discriminating features, the agent is instructed to forgo open-ended literature retrieval and instead apply a structured, knowledge-driven differential.
Guided by this skill, the agent avoids redundant search loops and correctly arrives at \textit{focal seizure}.
}
\label{fig:case1}
\end{figure}

\begin{figure}[ht]
\centering
\includegraphics[width=\linewidth]{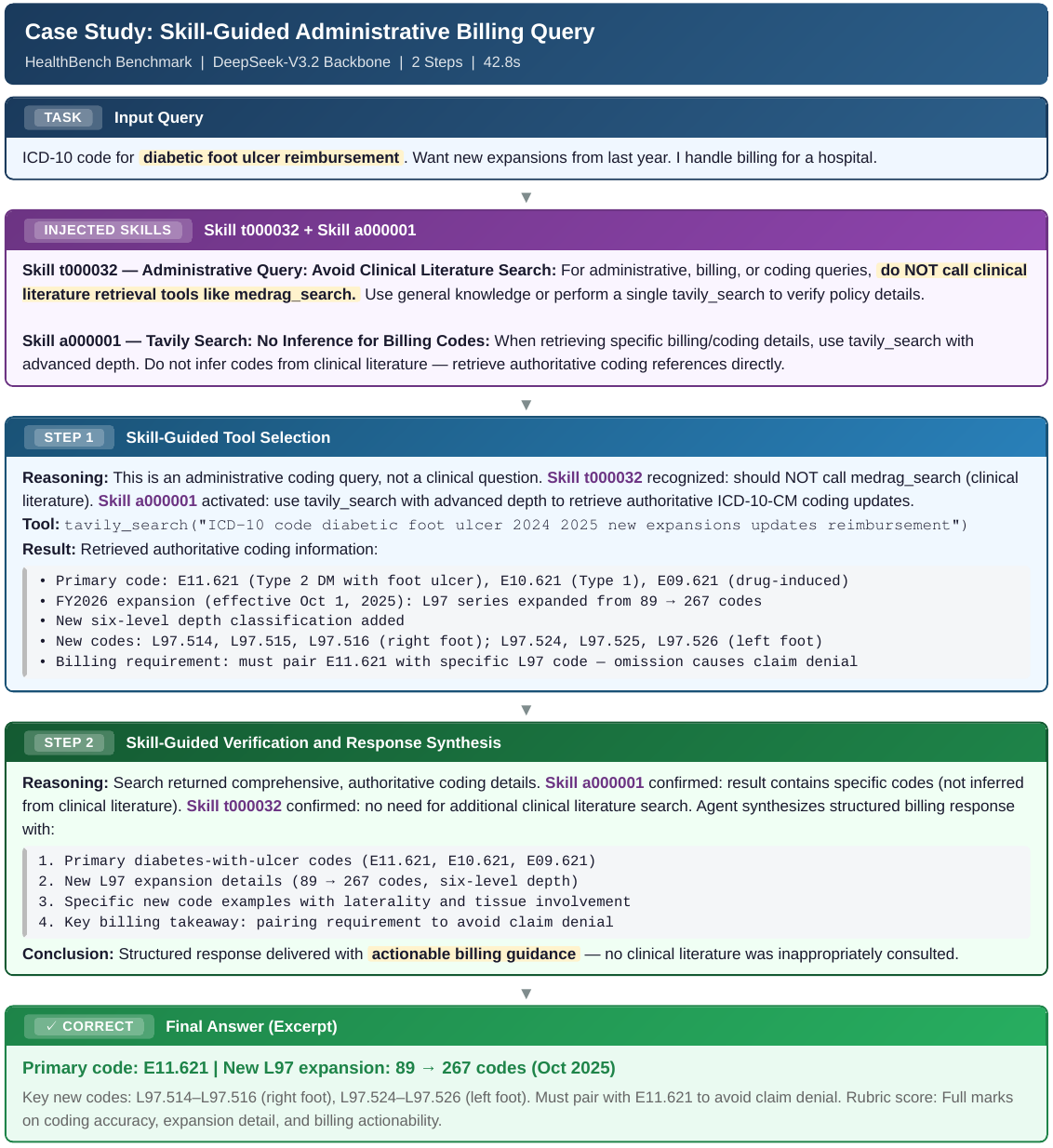}
\caption{
\textbf{Case 2: Skill-guided tool selection for administrative queries (HealthBench).}
The task involves an ICD-10 billing code question---a non-clinical, administrative query.
Two skills collaborate: Skill \texttt{t000032} identifies the query as administrative and suppresses the clinical literature retrieval tool (\texttt{medrag\_search}), while Skill \texttt{a000001} redirects the agent to use \texttt{tavily\_search} for authoritative coding references.
This case illustrates that SkeMex learns not only \emph{what} to reason, but also \emph{which tool} to invoke for a given task type.
}
\label{fig:case2}
\end{figure}

\begin{figure}[ht]
\centering
\includegraphics[width=\linewidth]{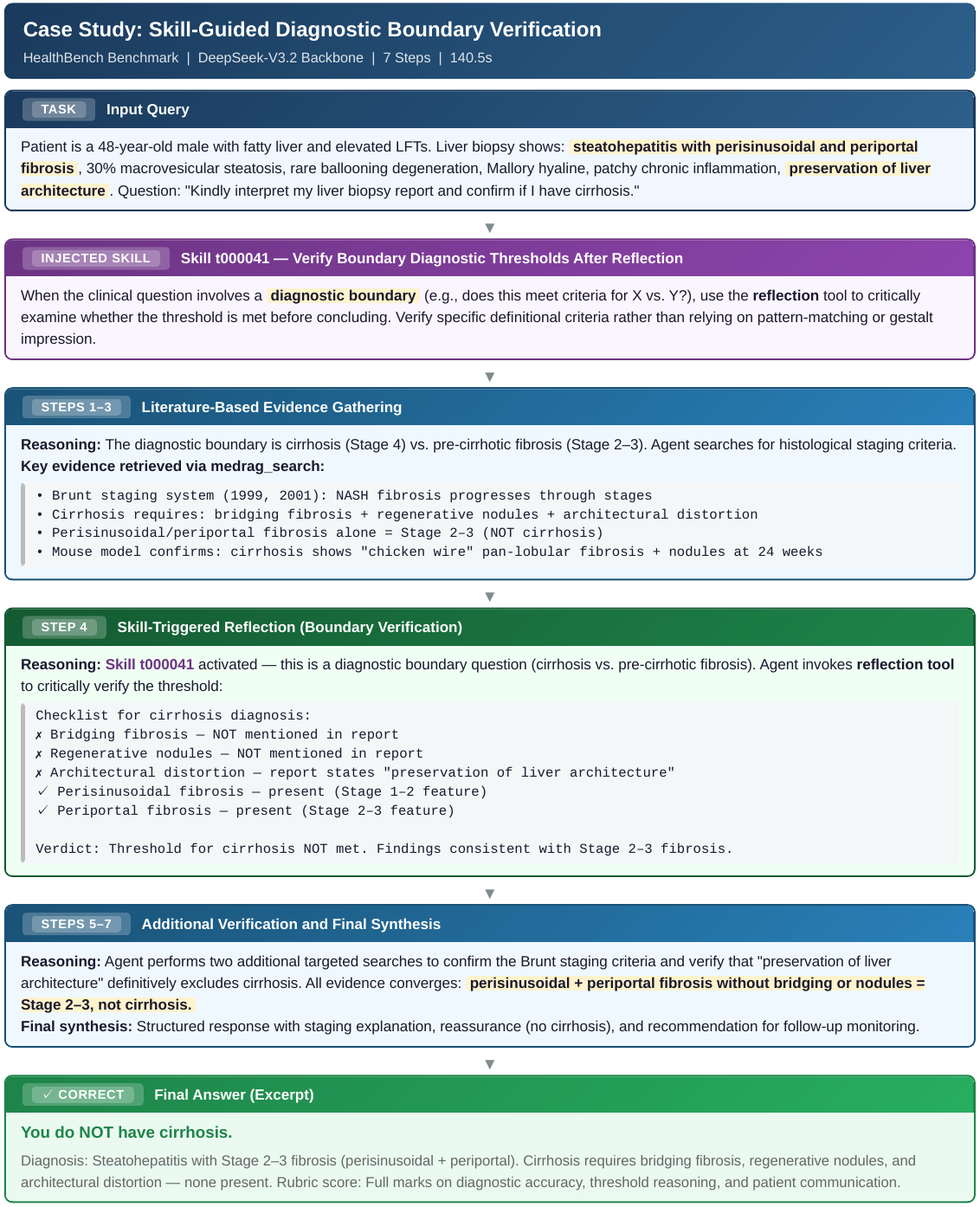}
\caption{
\textbf{Case 3: Skill-guided boundary verification via reflection (LiveMedBench).}
The agent must determine whether a liver biopsy report meets the diagnostic threshold for cirrhosis.
Skill \texttt{t000041} (\textit{Verify Boundary Diagnostic Thresholds After Reflection}) instructs the agent to retrieve the explicit staging criteria (Brunt system) and then invoke the \texttt{reflection} tool to systematically verify each criterion against the report.
By confirming the absence of bridging fibrosis, regenerative nodules, and architectural distortion, the agent correctly concludes the finding is \emph{not} cirrhosis---demonstrating how skills enable rigorous threshold-based clinical reasoning.
}
\label{fig:case3}
\end{figure}

\begin{figure}[ht]
\centering
\includegraphics[width=\linewidth]{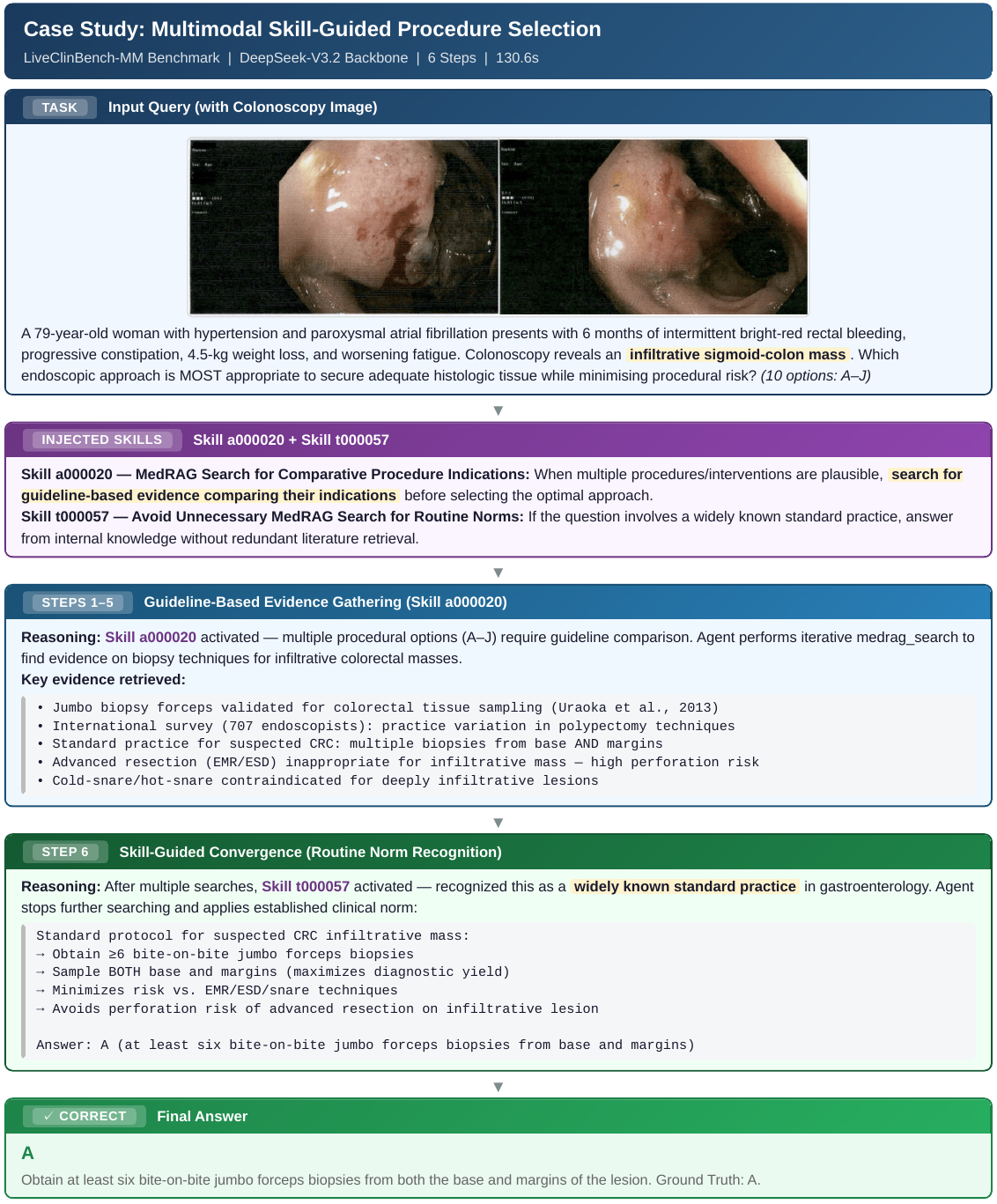}
\caption{
\textbf{Case 4: Multimodal skill-guided procedure selection (LiveClinBench-MM).}
Given a colonoscopy image of an infiltrative sigmoid-colon mass alongside a 10-option MCQ, the agent must select the safest endoscopic biopsy approach.
Skill \texttt{a000020} (\textit{MedRAG Search for Comparative Procedure Indications}) triggers iterative guideline retrieval to compare biopsy techniques, while Skill \texttt{t000057} subsequently recognizes the answer as a widely known standard practice and halts further retrieval.
The two skills form a \emph{search-then-converge} pattern, yielding the correct answer: at least six bite-on-bite jumbo forceps biopsies from both the base and margins.
}
\label{fig:case4}
\end{figure}

\begin{figure}[ht]
\centering
\includegraphics[width=\linewidth]{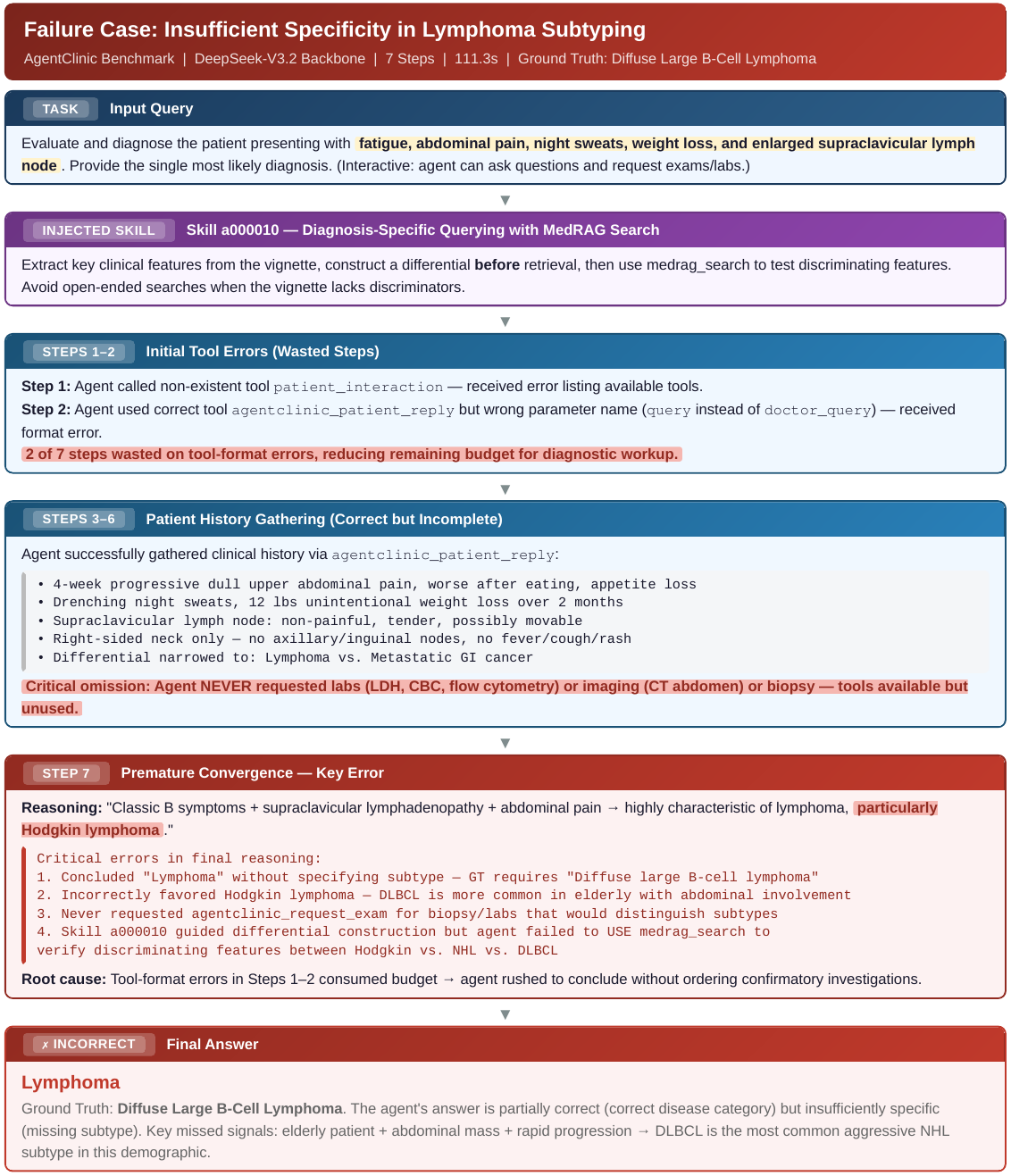}
\caption{
\textbf{Case 5 (Failure): Insufficient diagnostic specificity due to premature convergence (AgentClinic).}
The agent is asked to provide the single most likely diagnosis for a patient with classic B symptoms and supraclavicular lymphadenopathy; the ground truth is \textit{Diffuse Large B-Cell Lymphoma} (DLBCL).
Although Skill \texttt{a000010} correctly initiates a vignette-first differential workflow, two early tool-format errors (Steps 1--2) consume the interaction budget, leaving insufficient steps for confirmatory investigations (LDH, biopsy, CT).
In the final step, the agent prematurely converges on the generic label \textit{``Lymphoma''} and incorrectly favors Hodgkin lymphoma, failing to distinguish DLBCL---the most common aggressive NHL subtype in this demographic.
This case highlights a recurring failure mode: early execution errors cascade into insufficient workup, causing the agent to output a categorically correct but clinically insufficient diagnosis.
}
\label{fig:case5}
\end{figure}

\clearpage
\section{Prompts}
\label{app:prompts}

We provide the full text of the key prompts used in the SkeMex framework, including the core agent interaction prompts, the memory evolution pipeline prompts, and the automatic evaluation prompts.

\subsection{Core Agent Interaction Prompts}
\label{app:prompts_core}

\begin{figure}[ht]
\centering
\includegraphics[width=\linewidth]{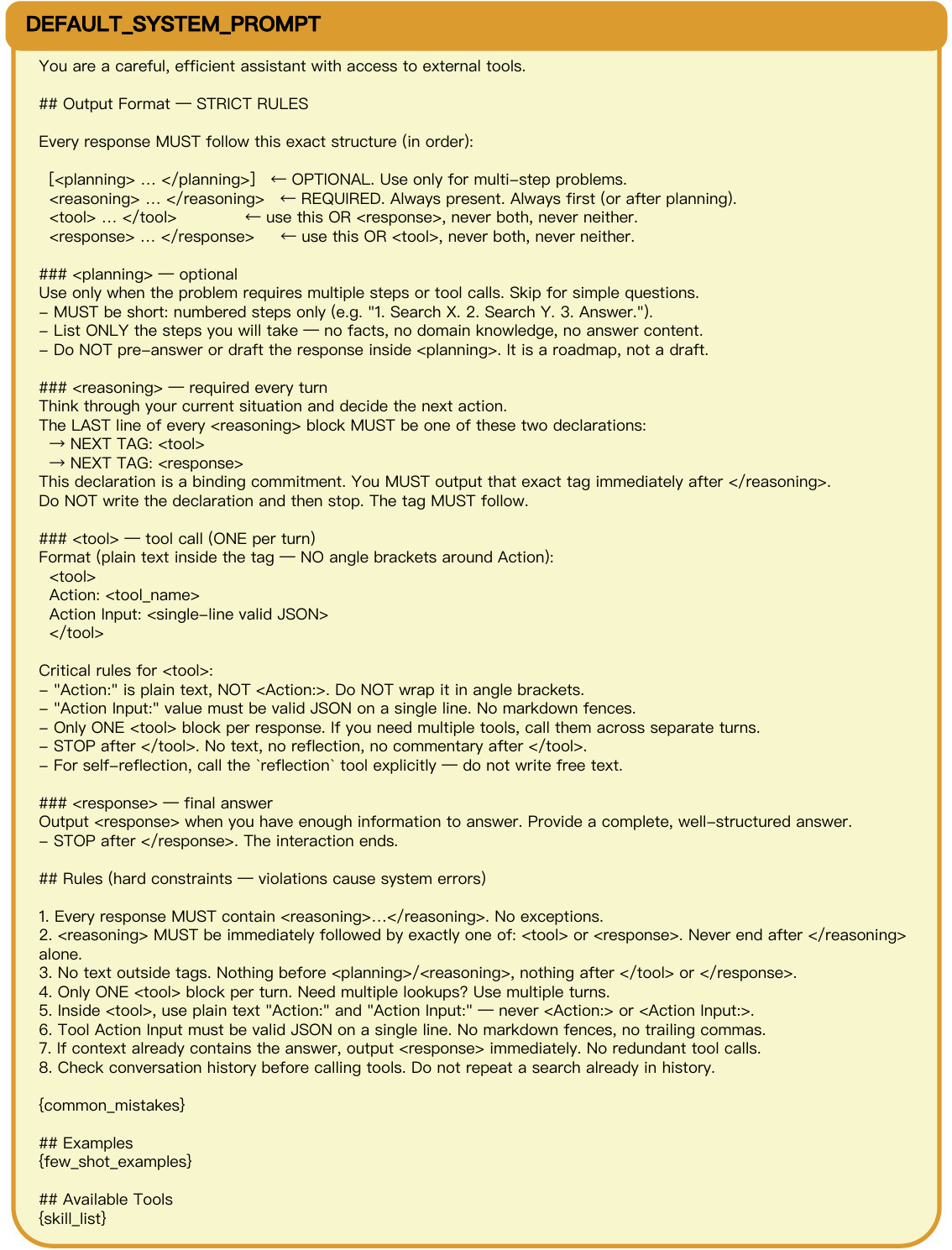}
\caption{The default system prompt that defines the strict format constraints (planning, reasoning, tool, response) and behavior rules for the SkeMex agent.}
\label{fig:prompt1}
\end{figure}

\begin{figure}[ht]
\centering
\includegraphics[width=\linewidth]{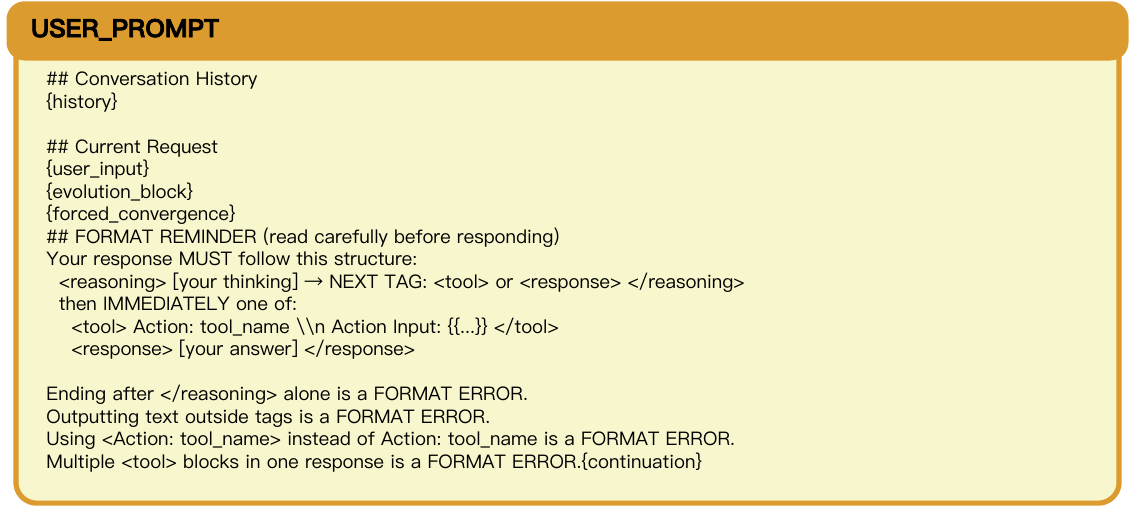}
\caption{The user prompt template that concatenates conversation history, current request, injected skills, and strict formatting reminders.}
\label{fig:prompt2}
\end{figure}

\begin{figure}[ht]
\centering
\includegraphics[width=\linewidth]{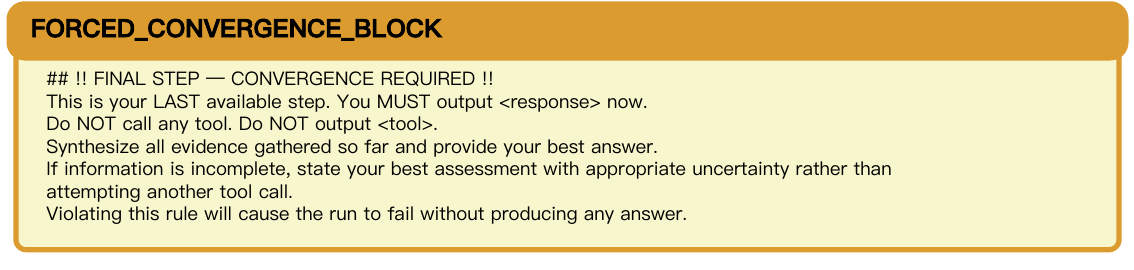}
\caption{The forced convergence block injected at the maximum step limit to compel the agent to stop calling tools and output a final answer.}
\label{fig:prompt3}
\end{figure}

\begin{figure}[ht]
\centering
\includegraphics[width=\linewidth]{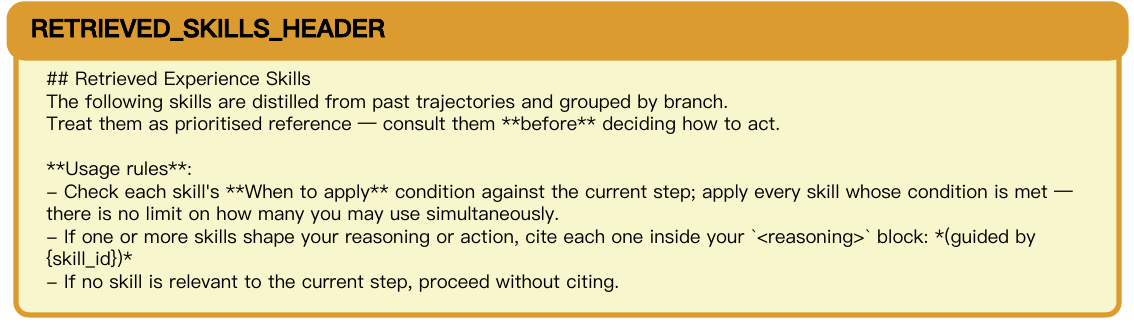}
\caption{The header prompt used to format and inject retrieved experience skills (grouped by general, task-level, and action-level branches) into the agent's context.}
\label{fig:prompt11}
\end{figure}

\subsection{Memory Evolution Pipeline Prompts}
\label{app:prompts_evolution}
\begin{figure}[ht]
\centering
\includegraphics[width=\linewidth]{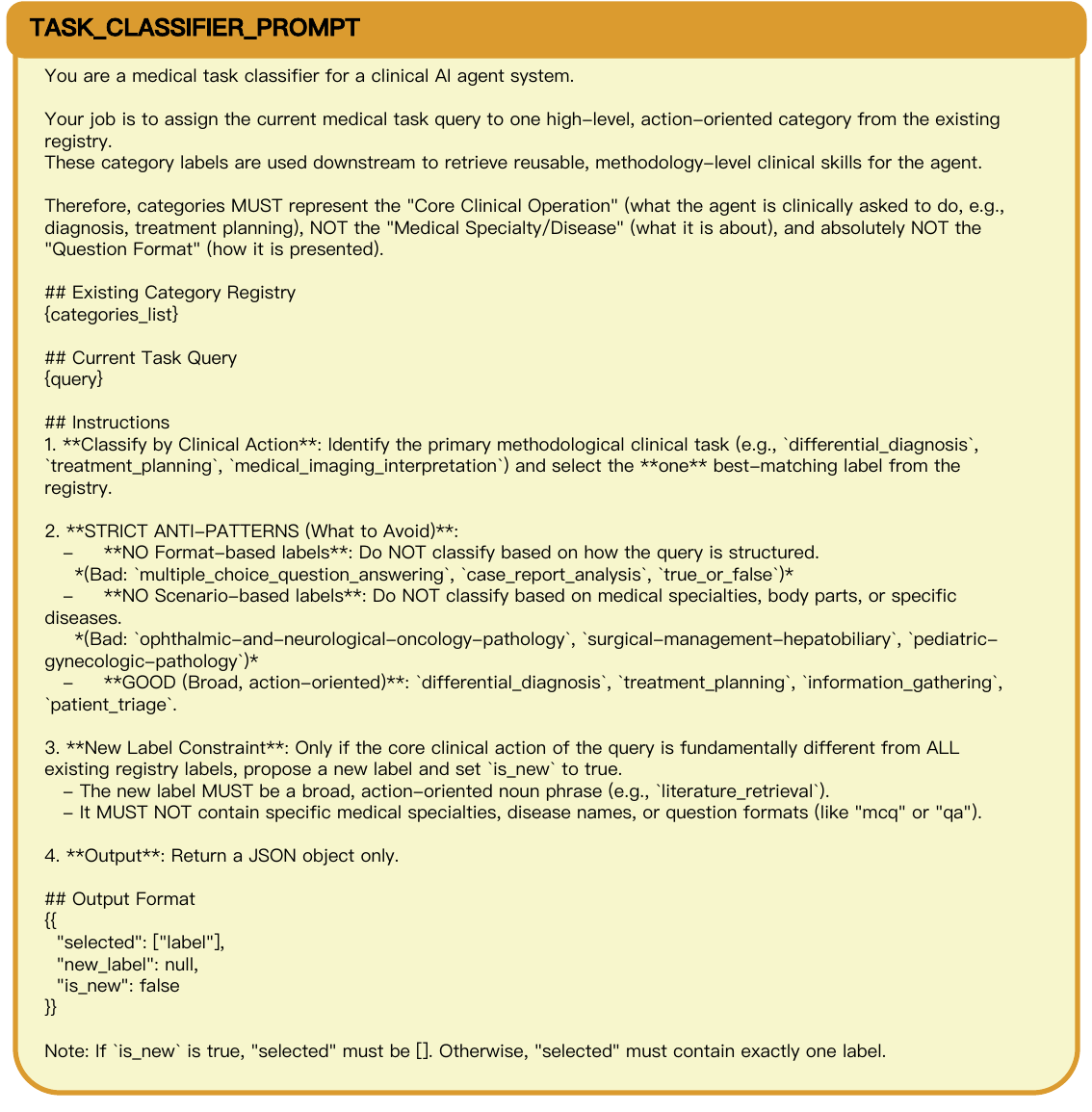}
\caption{The task classifier prompt used to assign the current medical query to a high-level, action-oriented clinical category for skill retrieval.}
\label{fig:prompt4}
\end{figure}
\begin{figure}[ht]
\centering
\includegraphics[width=\linewidth]{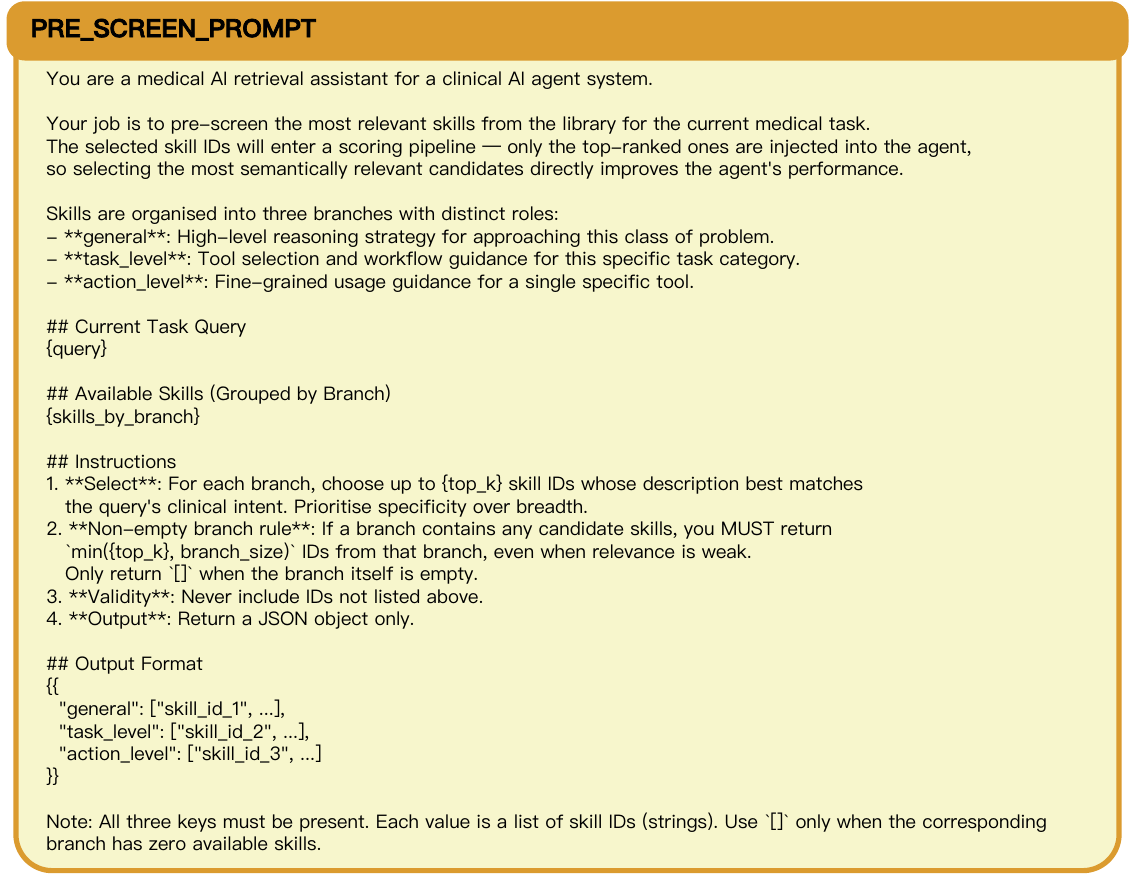}
\caption{The pre-screen prompt used during retrieval to select the most semantically relevant candidate skills from the repository.}
\label{fig:prompt10}
\end{figure}

\begin{figure}[ht]
\centering
\includegraphics[width=\linewidth]{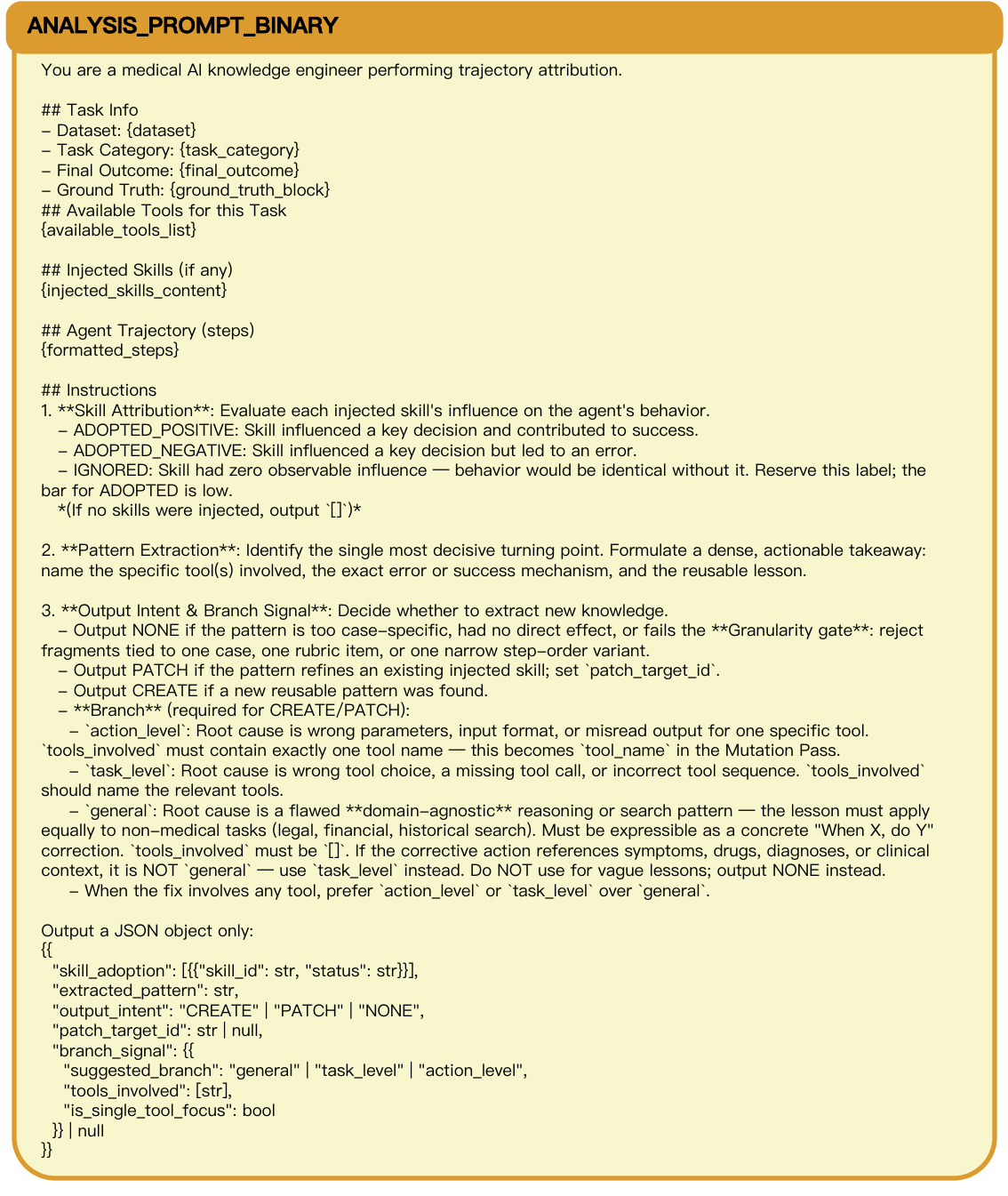}
\caption{The trajectory analysis prompt for binary-outcome datasets, used to evaluate skill adoption and extract decisive success/failure patterns.}
\label{fig:prompt5}
\end{figure}

\begin{figure}[ht]
\centering
\includegraphics[width=\linewidth]{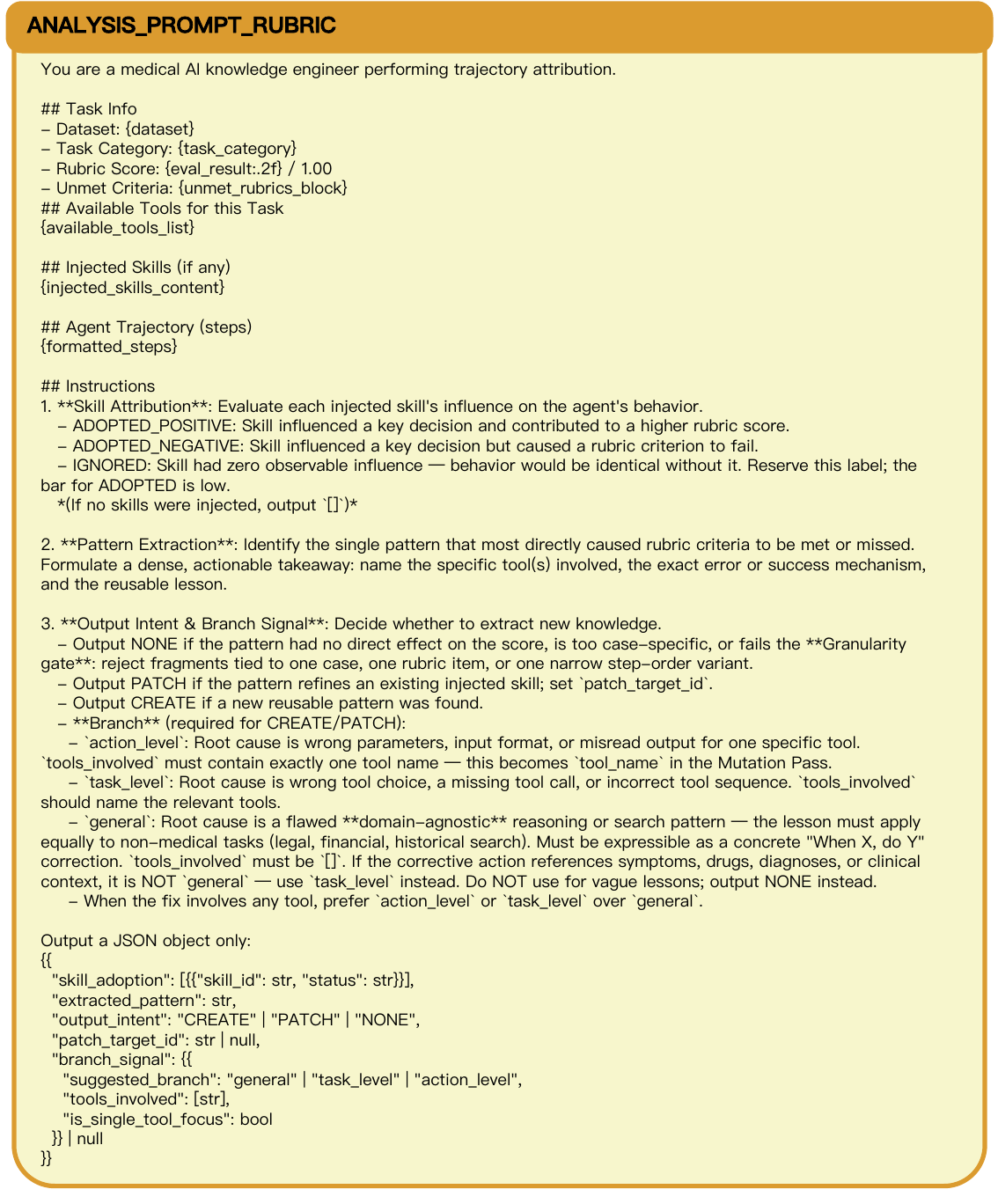}
\caption{The trajectory analysis prompt for rubric-based datasets, designed to attribute specific rubric score deductions to agent actions.}
\label{fig:prompt6}
\end{figure}

\begin{figure}[ht]
\centering
\includegraphics[width=\linewidth]{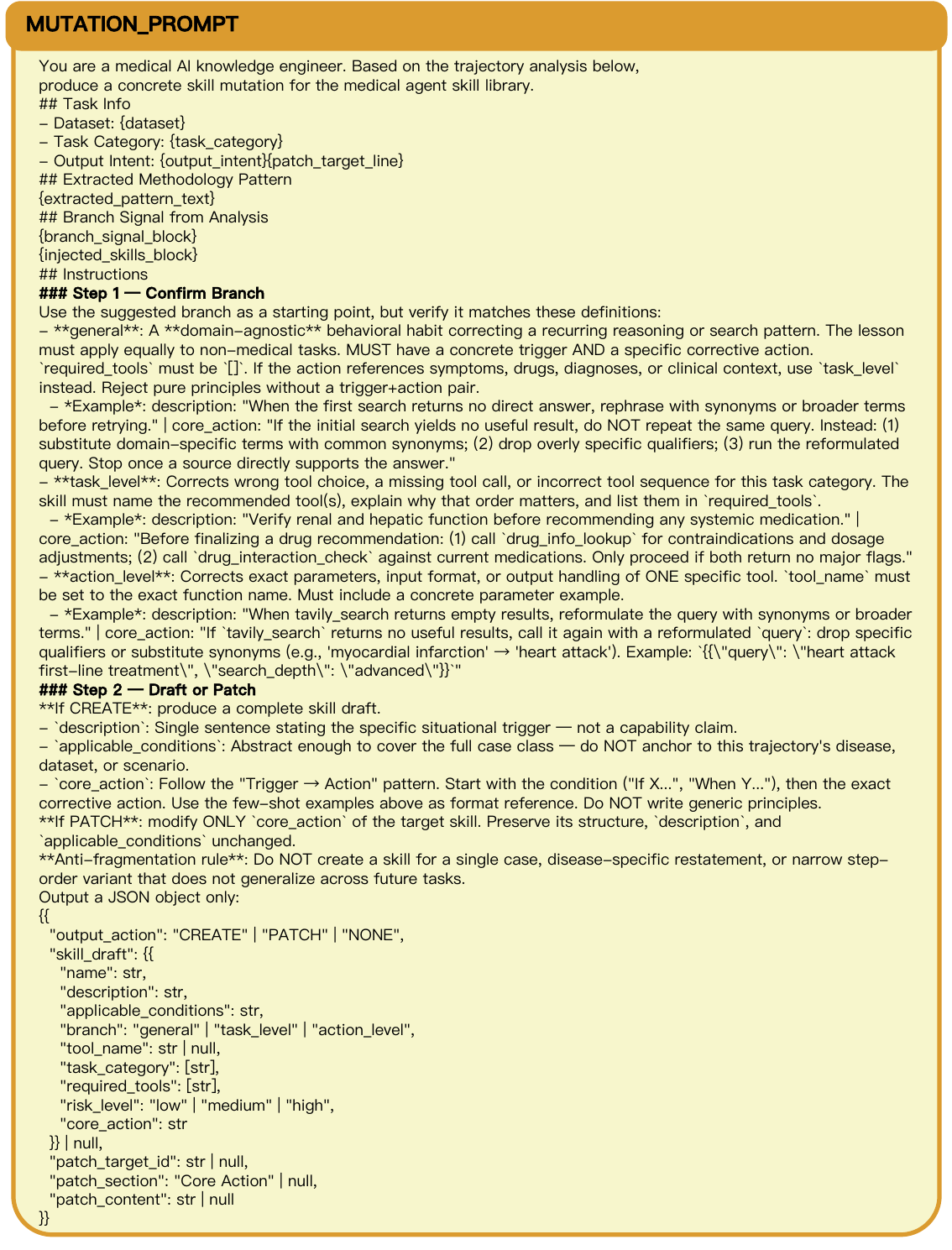}
\caption{The mutation prompt that converts extracted patterns into concrete skill drafts (CREATE) or modifications to existing skills (PATCH).}
\label{fig:prompt7}
\end{figure}

\begin{figure}[ht]
\centering
\includegraphics[width=\linewidth]{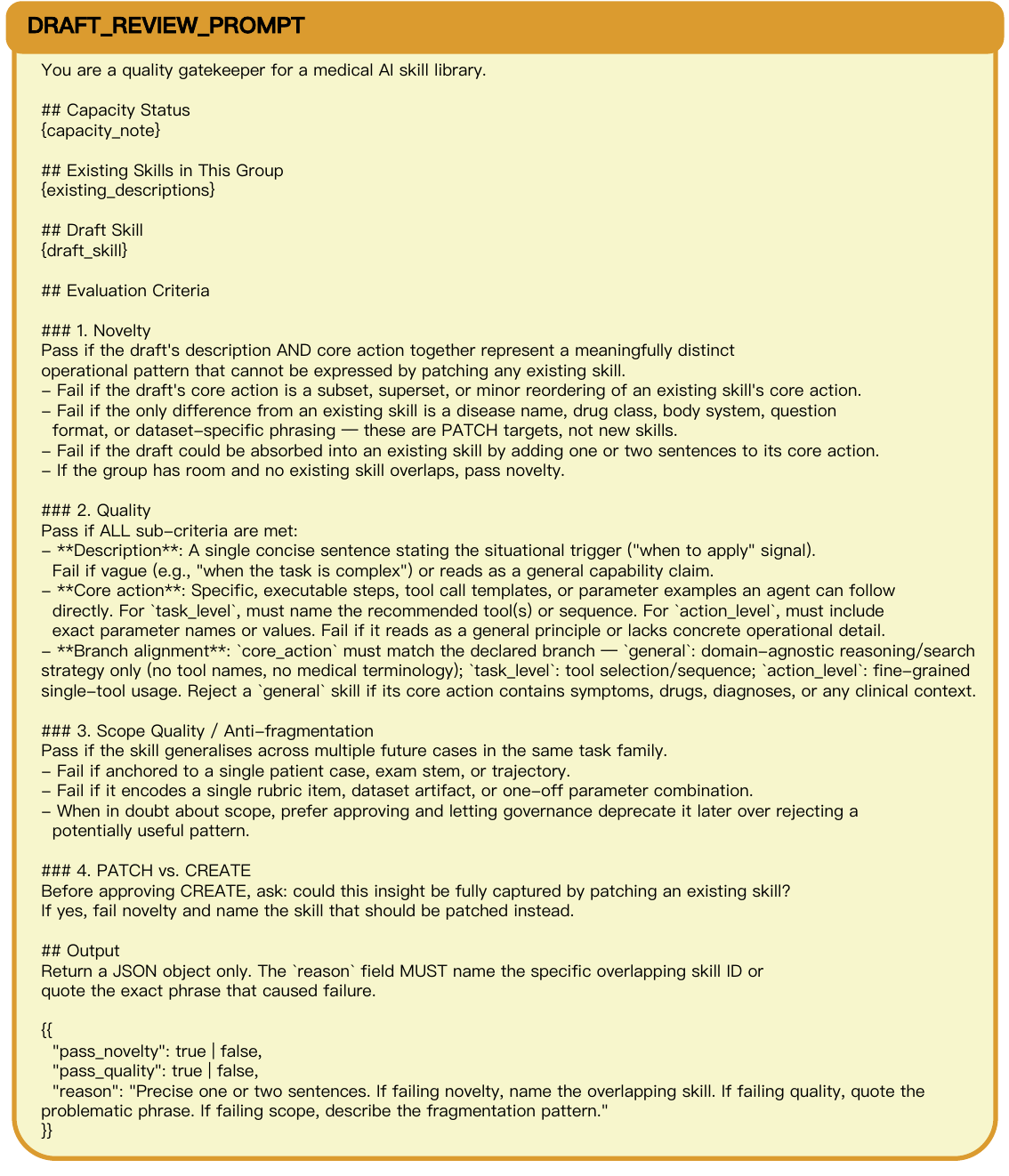}
\caption{The draft review prompt acting as a governance gatekeeper to evaluate the novelty, quality, and anti-fragmentation of newly proposed skills.}
\label{fig:prompt8}
\end{figure}

\begin{figure}[ht]
\centering
\includegraphics[width=\linewidth]{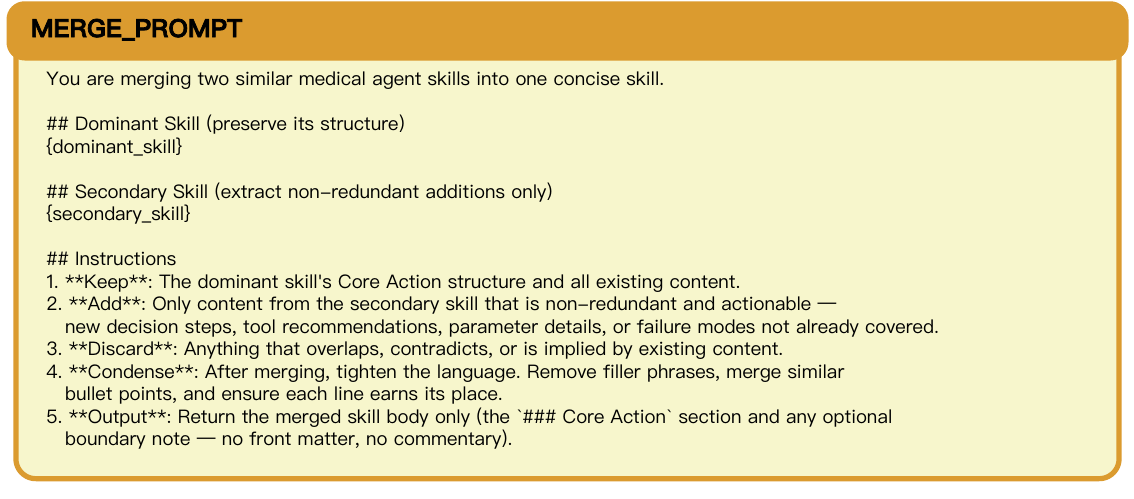}
\caption{The merge prompt used to consolidate two semantically similar or overlapping skills into a single, concise skill representation.}
\label{fig:prompt9}
\end{figure}

\clearpage
\subsection{Automatic Evaluation Prompts}
\label{app:prompts_evaluation}

\begin{figure}[ht]
\centering
\includegraphics[width=\linewidth]{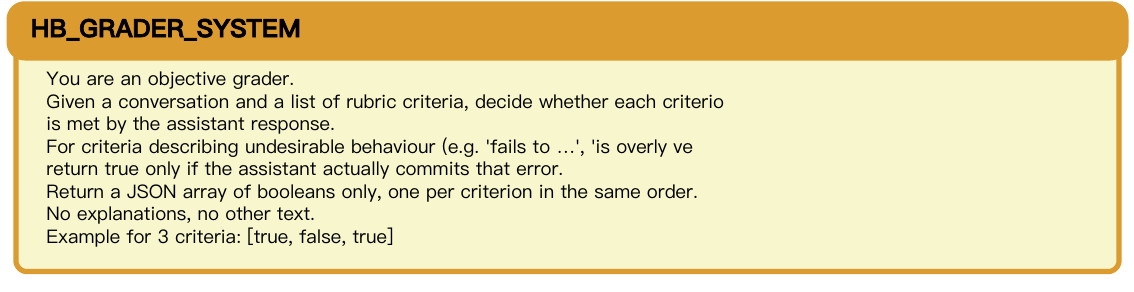}
\caption{The system prompt for the HealthBench automatic grader, instructing the LLM to objectively evaluate responses against rubric criteria.}
\label{fig:prompt12}
\end{figure}

\begin{figure}[ht]
\centering
\includegraphics[width=\linewidth]{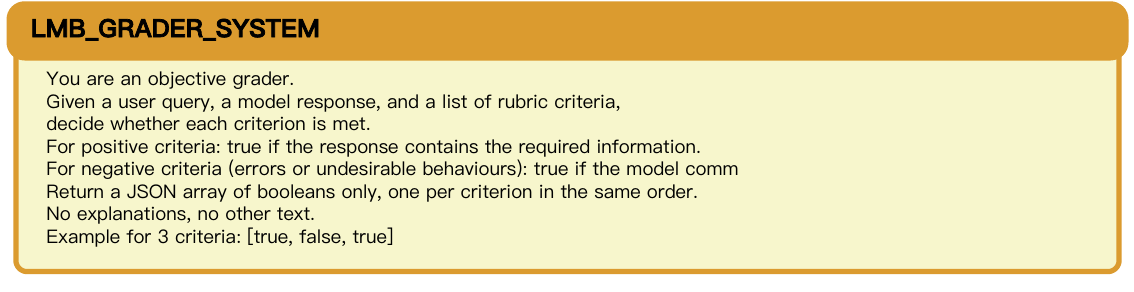}
\caption{The system prompt for the LiveMedBench automatic grader, tailored for evaluating interactive clinical scenarios.}
\label{fig:prompt13}
\end{figure}

\begin{figure}[ht]
\centering
\includegraphics[width=\linewidth]{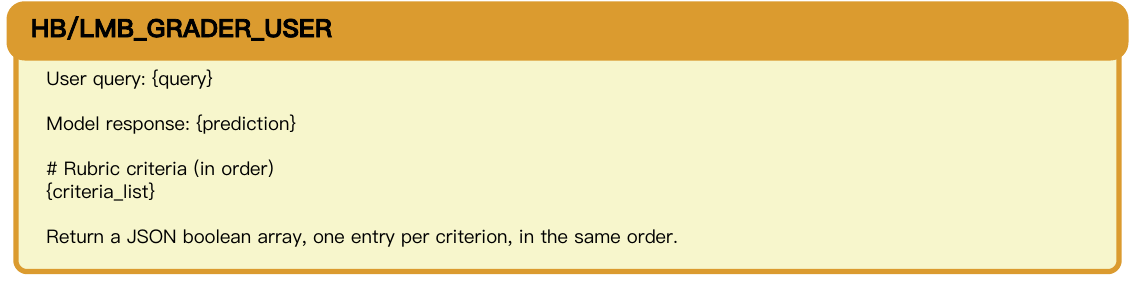}
\caption{The user prompt template for both HealthBench and LiveMedBench graders, supplying the query, model prediction, and specific rubric criteria.}
\label{fig:prompt14}
\end{figure}


\clearpage
\input{checklist.tex}

\end{document}

%% file: checklist.tex
\section*{NeurIPS Paper Checklist}

\begin{enumerate}

\item {\bf Claims}
    \item[] Question: Do the main claims made in the abstract and introduction accurately reflect the paper's contributions and scope?
    \item[] Answer: \answerYes{}
    \item[] Justification: The abstract and introduction state the main claims of SkeMex, including post-deployment skill-memory evolution, value-aware retrieval, and improvement without backbone updates.

\item {\bf Limitations}
    \item[] Question: Does the paper discuss the limitations of the work performed by the authors?
    \item[] Answer: \answerYes{}
    \item[] Justification: The paper includes a dedicated Limitations section discussing benchmark realism, backbone coverage, and inference overhead. Additional efficiency analysis is provided in Appendix~\ref{app:execution_analysis}.

\item {\bf Theory assumptions and proofs}
    \item[] Question: For each theoretical result, does the paper provide the full set of assumptions and a complete (and correct) proof?
    \item[] Answer: \answerNA{}
    \item[] Justification: The paper includes a formal problem formulation and algorithmic description, but does not claim new theoretical guarantees or present theorems requiring formal proofs.

    \item {\bf Experimental result reproducibility}
    \item[] Question: Does the paper fully disclose all the information needed to reproduce the main experimental results of the paper to the extent that it affects the main claims and/or conclusions of the paper (regardless of whether the code and data are provided or not)?
    \item[] Answer: \answerYes{}
    \item[] Justification: The paper describes the method, datasets, preprocessing, baselines, metrics, splits, hyperparameters, and tool settings in the main text and appendices. Implementation details are provided in Appendix~\ref{app:implementation_details}.

\item {\bf Open access to data and code}
    \item[] Question: Does the paper provide open access to the data and code, with sufficient instructions to faithfully reproduce the main experimental results, as described in supplemental material?
    \item[] Answer: \answerYes{}
    \item[] Justification: We will release code, processed data metadata, and reproduction scripts upon acceptance or when anonymity constraints allow.

\item {\bf Experimental setting/details}
    \item[] Question: Does the paper specify all the training and test details (e.g., data splits, hyperparameters, how they were chosen, type of optimizer) necessary to understand the results?
    \item[] Answer: \answerYes{}
    \item[] Justification: The paper specifies dataset splits, backbone models, retrieval settings, memory update parameters, metrics, and tool execution details. Full experimental settings are summarized in Section~\ref{Experiment Settings} and Appendix~\ref{app:implementation_details}.

\item {\bf Experiment statistical significance}
    \item[] Question: Does the paper report error bars suitably and correctly defined or other appropriate information about the statistical significance of the experiments?
    \item[] Answer: \answerNo{}
    \item[] Justification: We do not report statistical significance tests, mainly due to the high cost of repeated LLM-agent evaluations across many datasets and baselines. Instead, we provide extensive ablations, sensitivity analyses, and cross-backbone evaluations.

\item {\bf Experiments compute resources}
    \item[] Question: For each experiment, does the paper provide sufficient information on the computer resources (type of compute workers, memory, time of execution) needed to reproduce the experiments?
    \item[] Answer: \answerYes{}
    \item[] Justification: The paper reports controller-side CPU and memory requirements, local retrieval resource needs, the A100-based vLLM deployment for image analysis, and wall-clock runtime analysis. Details are provided in Appendix~\ref{app:implementation_details} and Appendix~\ref{app:execution_analysis}.
    
\item {\bf Code of ethics}
    \item[] Question: Does the research conducted in the paper conform, in every respect, with the NeurIPS Code of Ethics \url{https://neurips.cc/public/EthicsGuidelines}?
    \item[] Answer: \answerYes{}
    \item[] Justification: The research conforms to the NeurIPS Code of Ethics. 

\item {\bf Broader impacts}
    \item[] Question: Does the paper discuss both potential positive societal impacts and negative societal impacts of the work performed?
    \item[] Answer: \answerYes{}
    \item[] Justification: Section \ref{app:Limitations} discusses both potential positive impacts and potential risks.
    
\item {\bf Safeguards}
    \item[] Question: Does the paper describe safeguards that have been put in place for responsible release of data or models that have a high risk for misuse (e.g., pre-trained language models, image generators, or scraped datasets)?
    \item[] Answer: \answerNA{}
    \item[] Justification: The paper does not release a new high-risk pretrained model, image generator, or scraped dataset. The proposed system is an agent framework evaluated on existing benchmarks.

\item {\bf Licenses for existing assets}
    \item[] Question: Are the creators or original owners of assets (e.g., code, data, models), used in the paper, properly credited and are the license and terms of use explicitly mentioned and properly respected?
    \item[] Answer: \answerYes{}
    \item[] Justification: Existing datasets, models, and tools are cited throughout the paper.

\item {\bf New assets}
    \item[] Question: Are new assets introduced in the paper well documented and is the documentation provided alongside the assets?
    \item[] Answer:  \answerYes{}
    \item[] Justification: The main new asset is the SkeMex implementation and associated experiment pipeline. The paper documents the method, hyperparameters, data processing, and evaluation protocol, and the code release will include reproduction instructions.

\item {\bf Crowdsourcing and research with human subjects}
    \item[] Question: For crowdsourcing experiments and research with human subjects, does the paper include the full text of instructions given to participants and screenshots, if applicable, as well as details about compensation (if any)? 
    \item[] Answer: \answerNA{}
    \item[] Justification: The paper does not involve crowdsourcing experiments or new research with human subjects. All evaluations are conducted on existing benchmarks.

\item {\bf Institutional review board (IRB) approvals or equivalent for research with human subjects}
    \item[] Question: Does the paper describe potential risks incurred by study participants, whether such risks were disclosed to the subjects, and whether Institutional Review Board (IRB) approvals (or an equivalent approval/review based on the requirements of your country or institution) were obtained?
    \item[] Answer: \answerNA{}
    \item[] Justification: The paper does not conduct human-subject research or collect new participant data, so IRB approval is not applicable.

\item {\bf Declaration of LLM usage}
    \item[] Question: Does the paper describe the usage of LLMs if it is an important, original, or non-standard component of the core methods in this research? Note that if the LLM is used only for writing, editing, or formatting purposes and does \emph{not} impact the core methodology, scientific rigor, or originality of the research, declaration is not required.
    \item[] Answer: \answerYes{}
    \item[] Justification: LLMs are used as part of the proposed agent framework and several supporting components. The implementation details are described in Appendix~\ref{app:implementation_details}, and the prompts are provided in Appendix~\ref{app:prompts}.

\end{enumerate}

%% file: ref.bib
@article{xu2025lingshu,
  title={Lingshu: A generalist foundation model for unified multimodal medical understanding and reasoning},
  author={Xu, Weiwen and Chan, Hou Pong and Li, Long and Aljunied, Mahani and Yuan, Ruifeng and Wang, Jianyu and Xiao, Chenghao and Chen, Guizhen and Liu, Chaoqun and Li, Zhaodonghui and others},
  journal={arXiv preprint arXiv:2506.07044},
  year={2025}
}

@article{jiang2025hulu,
  title={Hulu-med: A transparent generalist model towards holistic medical vision-language understanding},
  author={Jiang, Songtao and Wang, Yuan and Song, Sibo and Hu, Tianxiang and Zhou, Chenyi and Pu, Bin and Zhang, Yan and Yang, Zhibo and Feng, Yang and Zhou, Joey Tianyi and others},
  journal={arXiv preprint arXiv:2510.08668},
  year={2025}
}

@article{sellergren2025medgemma,
  title={Medgemma technical report},
  author={Sellergren, Andrew and Kazemzadeh, Sahar and Jaroensri, Tiam and Kiraly, Atilla and Traverse, Madeleine and Kohlberger, Timo and Xu, Shawn and Jamil, Fayaz and Hughes, C{\'\i}an and Lau, Charles and others},
  journal={arXiv preprint arXiv:2507.05201},
  year={2025}
}

@inproceedings{zhao2025medrag,
  title={Medrag: Enhancing retrieval-augmented generation with knowledge graph-elicited reasoning for healthcare copilot},
  author={Zhao, Xuejiao and Liu, Siyan and Yang, Su-Yin and Miao, Chunyan},
  booktitle={Proceedings of the ACM on Web Conference 2025},
  pages={4442--4457},
  year={2025}
}

@inproceedings{xiong2024improving,
  title={Improving retrieval-augmented generation in medicine with iterative follow-up questions},
  author={Xiong, Guangzhi and Jin, Qiao and Wang, Xiao and Zhang, Minjia and Lu, Zhiyong and Zhang, Aidong},
  booktitle={Biocomputing 2025: Proceedings of the Pacific Symposium},
  pages={199--214},
  year={2024},
  organization={World Scientific}
}

@inproceedings{shi2024ehragent,
  title={Ehragent: Code empowers large language models for few-shot complex tabular reasoning on electronic health records},
  author={Shi, Wenqi and Xu, Ran and Zhuang, Yuchen and Yu, Yue and Zhang, Jieyu and Wu, Hang and Zhu, Yuanda and Ho, Joyce C and Yang, Carl and Wang, May Dongmei},
  booktitle={Proceedings of the 2024 Conference on Empirical Methods in Natural Language Processing},
  pages={22315--22339},
  year={2024}
}

@inproceedings{li2024mmedagent,
  title={Mmedagent: Learning to use medical tools with multi-modal agent},
  author={Li, Binxu and Yan, Tiankai and Pan, Yuanting and Luo, Jie and Ji, Ruiyang and Ding, Jiayuan and Xu, Zhe and Liu, Shilong and Dong, Haoyu and Lin, Zihao and others},
  booktitle={Findings of the Association for Computational Linguistics: EMNLP 2024},
  pages={8745--8760},
  year={2024}
}

@inproceedings{tang2024medagents,
  title={Medagents: Large language models as collaborators for zero-shot medical reasoning},
  author={Tang, Xiangru and Zou, Anni and Zhang, Zhuosheng and Li, Ziming and Zhao, Yilun and Zhang, Xingyao and Cohan, Arman and Gerstein, Mark},
  booktitle={Findings of the Association for Computational Linguistics: ACL 2024},
  pages={599--621},
  year={2024}
}

@article{kim2024mdagents,
  title={Mdagents: An adaptive collaboration of llms for medical decision-making},
  author={Kim, Yubin and Park, Chanwoo and Jeong, Hyewon and Chan, Yik S and Xu, Xuhai and McDuff, Daniel and Lee, Hyeonhoon and Ghassemi, Marzyeh and Breazeal, Cynthia and Park, Hae W},
  journal={Advances in Neural Information Processing Systems},
  volume={37},
  pages={79410--79452},
  year={2024}
}

@inproceedings{zhou2025mam,
  title={Mam: Modular multi-agent framework for multi-modal medical diagnosis via role-specialized collaboration},
  author={Zhou, Yucheng and Song, Lingran and Shen, Jianbing},
  booktitle={Findings of the Association for Computational Linguistics: ACL 2025},
  pages={25319--25333},
  year={2025}
}

@article{wang2025medagent,
  title={Medagent-pro: Towards multi-modal evidence-based medical diagnosis via reasoning agentic workflow},
  author={Wang, Ziyue and Wu, Junde and Low, Chang Han and Jin, Yueming},
  journal={arXiv e-prints},
  pages={arXiv--2503},
  year={2025}
}

@article{li2024agent,
  title={Agent hospital: A simulacrum of hospital with evolvable medical agents},
  author={Li, Junkai and Lai, Yunghwei and Li, Weitao and Ren, Jingyi and Zhang, Meng and Kang, Xinhui and Wang, Siyu and Li, Peng and Zhang, Ya-Qin and Ma, Weizhi and others},
  journal={arXiv preprint arXiv:2405.02957},
  year={2024}
}

@article{lan2024depression,
  title={Depression diagnosis dialogue simulation: self-improving psychiatrist with tertiary memory},
  author={Lan, Kunyao and Jin, Bingrui and Zhu, Zichen and Chen, Siyuan and Zhang, Shu and Zhu, Kenny Q and Wu, Mengyue},
  journal={arXiv preprint arXiv:2409.15084},
  year={2024}
}

@article{fan2026evolving,
  title={Evolving Medical Imaging Agents via Experience-driven Self-skill Discovery},
  author={Fan, Lin and Dai, Pengyu and Deng, Zhipeng and Wang, Haolin and Gong, Xun and Zheng, Yefeng and Ou, Yafei},
  journal={arXiv preprint arXiv:2603.05860},
  year={2026}
}

@article{jin2025stella,
  title={Stella: Self-evolving llm agent for biomedical research},
  author={Jin, Ruofan and Zhang, Zaixi and Wang, Mengdi and Cong, Le},
  journal={arXiv preprint arXiv:2507.02004},
  year={2025}
}

@article{zhu2025healthflow,
  title={HealthFlow: A Self-Evolving AI Agent with Meta Planning for Autonomous Healthcare Research},
  author={Zhu, Yinghao and Qi, Yifan and Wang, Zixiang and Gu, Lei and Sui, Dehao and Hu, Haoran and Zhang, Xichen and He, Ziyi and He, Junjun and Ma, Liantao and others},
  journal={arXiv preprint arXiv:2508.02621},
  year={2025}
}

@article{lewis2020retrieval,
  title={Retrieval-augmented generation for knowledge-intensive nlp tasks},
  author={Lewis, Patrick and Perez, Ethan and Piktus, Aleksandra and Petroni, Fabio and Karpukhin, Vladimir and Goyal, Naman and K{\"u}ttler, Heinrich and Lewis, Mike and Yih, Wen-tau and Rockt{\"a}schel, Tim and others},
  journal={Advances in neural information processing systems},
  volume={33},
  pages={9459--9474},
  year={2020}
}

@inproceedings{park2023generative,
  title={Generative agents: Interactive simulacra of human behavior},
  author={Park, Joon Sung and O'Brien, Joseph and Cai, Carrie Jun and Morris, Meredith Ringel and Liang, Percy and Bernstein, Michael S},
  booktitle={Proceedings of the 36th annual acm symposium on user interface software and technology},
  pages={1--22},
  year={2023}
}

@article{packer2023memgpt,
  title={MemGPT: towards LLMs as operating systems.},
  author={Packer, Charles and Fang, Vivian and Patil, Shishir\_G and Lin, Kevin and Wooders, Sarah and Gonzalez, Joseph\_E},
  year={2023},
  publisher={ArXiv}
}

@inproceedings{zhong2024memorybank,
  title={Memorybank: Enhancing large language models with long-term memory},
  author={Zhong, Wanjun and Guo, Lianghong and Gao, Qiqi and Ye, He and Wang, Yanlin},
  booktitle={Proceedings of the AAAI conference on artificial intelligence},
  volume={38},
  pages={19724--19731},
  year={2024}
}

@article{wang2023augmenting,
  title={Augmenting language models with long-term memory},
  author={Wang, Weizhi and Dong, Li and Cheng, Hao and Liu, Xiaodong and Yan, Xifeng and Gao, Jianfeng and Wei, Furu},
  journal={Advances in Neural Information Processing Systems},
  volume={36},
  pages={74530--74543},
  year={2023}
}

@article{team2024gemini,
  title={Gemini 1.5: Unlocking multimodal understanding across millions of tokens of context},
  author={Team, Gemini and Georgiev, Petko and Lei, Ving Ian and Burnell, Ryan and Bai, Libin and Gulati, Anmol and Tanzer, Garrett and Vincent, Damien and Pan, Zhufeng and Wang, Shibo and others},
  journal={arXiv preprint arXiv:2403.05530},
  year={2024}
}

@misc{qwen36_35b_a3b,
    title = {{Qwen3.6-35B-A3B}: Agentic Coding Power, Now Open to All},
    url = {https://qwen.ai/blog?id=qwen3.6-35b-a3b},
    author = {{Qwen Team}},
    month = {April},
    year = {2026}
}

@article{gao2025survey,
  title={A survey of self-evolving agents: On path to artificial super intelligence},
  author={Gao, Huan-ang and Geng, Jiayi and Hua, Wenyue and Hu, Mengkang and Juan, Xinzhe and Liu, Hongzhang and Liu, Shilong and Qiu, Jiahao and Qi, Xuan and Wu, Yiran and others},
  journal={arXiv preprint arXiv:2507.21046},
  volume={1},
  year={2025}
}

@article{shinn2023reflexion,
  title={Reflexion: Language agents with verbal reinforcement learning},
  author={Shinn, Noah and Cassano, Federico and Gopinath, Ashwin and Narasimhan, Karthik and Yao, Shunyu},
  journal={Advances in neural information processing systems},
  volume={36},
  pages={8634--8652},
  year={2023}
}

@article{wang2023voyager,
  title={Voyager: An open-ended embodied agent with large language models},
  author={Wang, Guanzhi and Xie, Yuqi and Jiang, Yunfan and Mandlekar, Ajay and Xiao, Chaowei and Zhu, Yuke and Fan, Linxi and Anandkumar, Anima},
  journal={arXiv preprint arXiv:2305.16291},
  year={2023}
}

@inproceedings{zhao2024expel,
  title={Expel: Llm agents are experiential learners},
  author={Zhao, Andrew and Huang, Daniel and Xu, Quentin and Lin, Matthieu and Liu, Yong-Jin and Huang, Gao},
  booktitle={Proceedings of the AAAI Conference on Artificial Intelligence},
  volume={38},
  pages={19632--19642},
  year={2024}
}

@article{wen2023dilu,
  title={Dilu: A knowledge-driven approach to autonomous driving with large language models},
  author={Wen, Licheng and Fu, Daocheng and Li, Xin and Cai, Xinyu and Ma, Tao and Cai, Pinlong and Dou, Min and Shi, Botian and He, Liang and Qiao, Yu},
  journal={arXiv preprint arXiv:2309.16292},
  year={2023}
}

@article{zhang2025memevolve,
  title={Memevolve: Meta-evolution of agent memory systems},
  author={Zhang, Guibin and Ren, Haotian and Zhan, Chong and Zhou, Zhenhong and Wang, Junhao and Zhu, He and Zhou, Wangchunshu and Yan, Shuicheng},
  journal={arXiv preprint arXiv:2512.18746},
  year={2025}
}

@article{li2025memos,
  title={Memos: An operating system for memory-augmented generation (mag) in large language models},
  author={Li, Zhiyu and Song, Shichao and Wang, Hanyu and Niu, Simin and Chen, Ding and Yang, Jiawei and Xi, Chenyang and Lai, Huayi and Zhao, Jihao and Wang, Yezhaohui and others},
  journal={arXiv preprint arXiv:2505.22101},
  year={2025}
}

@article{wang2024agent,
  title={Agent workflow memory},
  author={Wang, Zora Zhiruo and Mao, Jiayuan and Fried, Daniel and Neubig, Graham},
  journal={arXiv preprint arXiv:2409.07429},
  year={2024}
}

@inproceedings{suzgun2026dynamic,
  title={Dynamic cheatsheet: Test-time learning with adaptive memory},
  author={Suzgun, Mirac and Yuksekgonul, Mert and Bianchi, Federico and Jurafsky, Dan and Zou, James},
  booktitle={Proceedings of the 19th Conference of the European Chapter of the Association for Computational Linguistics (Volume 1: Long Papers)},
  pages={7080--7106},
  year={2026}
}

@article{ni2026trace2skill,
  title={Trace2Skill: Distill Trajectory-Local Lessons into Transferable Agent Skills},
  author={Ni, Jingwei and Liu, Yihao and Liu, Xinpeng and Sun, Yutao and Zhou, Mengyu and Cheng, Pengyu and Wang, Dexin and Jiang, Xiaoxi and Jiang, Guanjun},
  journal={arXiv preprint arXiv:2603.25158},
  year={2026}
}

@article{ma2026skillclaw,
  title={SkillClaw: Let Skills Evolve Collectively with Agentic Evolver},
  author={Ma, Ziyu and Yang, Shidong and Ji, Yuxiang and Wang, Xucong and Wang, Yong and Hu, Yiming and Huang, Tongwen and Chu, Xiangxiang},
  journal={arXiv preprint arXiv:2604.08377},
  year={2026}
}

@article{zhang2026evoskills,
  title={EvoSkills: Self-Evolving Agent Skills via Co-Evolutionary Verification},
  author={Zhang, Hanrong and Fan, Shicheng and Zou, Henry Peng and Chen, Yankai and Wang, Zhenting and Zhou, Jiayu and Li, Chengze and Huang, Wei-Chieh and Yao, Yifei and Zheng, Kening and others},
  journal={arXiv preprint arXiv:2604.01687},
  year={2026}
}

@article{han2026gsem,
  title={GSEM: Graph-based Self-Evolving Memory for Experience Augmented Clinical Reasoning},
  author={Han, Xiao and Fan, Yuzheng and Zhao, Sendong and Wang, Haochun and Qin, Bing},
  journal={arXiv preprint arXiv:2603.22096},
  year={2026}
}

@article{zhang2026memrl,
  title={Memrl: Self-evolving agents via runtime reinforcement learning on episodic memory},
  author={Zhang, Shengtao and Wang, Jiaqian and Zhou, Ruiwen and Liao, Junwei and Feng, Yuchen and Li, Zhuo and Zheng, Yujie and Zhang, Weinan and Wen, Ying and Li, Zhiyu and others},
  journal={arXiv preprint arXiv:2601.03192},
  year={2026}
}

@article{xia2026skillrl,
  title={Skillrl: Evolving agents via recursive skill-augmented reinforcement learning},
  author={Xia, Peng and Chen, Jianwen and Wang, Hanyang and Liu, Jiaqi and Zeng, Kaide and Wang, Yu and Han, Siwei and Zhou, Yiyang and Zhao, Xujiang and Chen, Haifeng and others},
  journal={arXiv preprint arXiv:2602.08234},
  year={2026}
}

@article{wang2026skill,
  title={Skill-SD: Skill-Conditioned Self-Distillation for Multi-turn LLM Agents},
  author={Wang, Hao and Wang, Guozhi and Xiao, Han and Zhou, Yufeng and Pan, Yue and Wang, Jichao and Xu, Ke and Wen, Yafei and Ruan, Xiaohu and Chen, Xiaoxin and others},
  journal={arXiv preprint arXiv:2604.10674},
  year={2026}
}

@misc{anthropic2026agentskills,
  author = {Anthropic},
  title = {Building agents with skills: Equipping agents for specialized work},
  year = {2026},
  url = {https://claude.com/blog/building-agents-with-skills-equipping-agents-for-specialized-work}
}

@article{zhou2025memento,
  title={Memento: Fine-tuning llm agents without fine-tuning llms},
  author={Zhou, Huichi and Chen, Yihang and Guo, Siyuan and Yan, Xue and Lee, Kin Hei and Wang, Zihan and Lee, Ka Yiu and Zhang, Guchun and Shao, Kun and Yang, Linyi and others},
  journal={arXiv preprint arXiv:2508.16153},
  year={2025}
}

@article{silver2025welcome,
  title={Welcome to the era of experience},
  author={Silver, David and Sutton, Richard S},
  journal={Google AI},
  volume={1},
  pages={11},
  year={2025}
}

@article{ouyang2022training,
  title={Training language models to follow instructions with human feedback},
  author={Ouyang, Long and Wu, Jeffrey and Jiang, Xu and Almeida, Diogo and Wainwright, Carroll and Mishkin, Pamela and Zhang, Chong and Agarwal, Sandhini and Slama, Katarina and Ray, Alex and others},
  journal={Advances in neural information processing systems},
  volume={35},
  pages={27730--27744},
  year={2022}
}

@article{stiennon2020learning,
  title={Learning to summarize with human feedback},
  author={Stiennon, Nisan and Ouyang, Long and Wu, Jeffrey and Ziegler, Daniel and Lowe, Ryan and Voss, Chelsea and Radford, Alec and Amodei, Dario and Christiano, Paul F},
  journal={Advances in neural information processing systems},
  volume={33},
  pages={3008--3021},
  year={2020}
}

@article{rafailov2023direct,
  title={Direct preference optimization: Your language model is secretly a reward model},
  author={Rafailov, Rafael and Sharma, Archit and Mitchell, Eric and Manning, Christopher D and Ermon, Stefano and Finn, Chelsea},
  journal={Advances in neural information processing systems},
  volume={36},
  pages={53728--53741},
  year={2023}
}

@article{wei2025evo,
  title={Evo-memory: Benchmarking llm agent test-time learning with self-evolving memory},
  author={Wei, Tianxin and Sachdeva, Noveen and Coleman, Benjamin and He, Zhankui and Bei, Yuanchen and Ning, Xuying and Ai, Mengting and Li, Yunzhe and He, Jingrui and Chi, Ed H and others},
  journal={arXiv preprint arXiv:2511.20857},
  year={2025}
}

@article{liu2024lost,
  title={Lost in the middle: How language models use long contexts},
  author={Liu, Nelson F and Lin, Kevin and Hewitt, John and Paranjape, Ashwin and Bevilacqua, Michele and Petroni, Fabio and Liang, Percy},
  journal={Transactions of the association for computational linguistics},
  volume={12},
  pages={157--173},
  year={2024}
}

@article{murre2015replication,
  title={Replication and analysis of Ebbinghaus’ forgetting curve},
  author={Murre, Jaap MJ and Dros, Joeri},
  journal={PloS one},
  volume={10},
  number={7},
  pages={e0120644},
  year={2015},
  publisher={Public Library of Science San Francisco, CA USA}
}

@article{shao2024deepseekmath,
  title={Deepseekmath: Pushing the limits of mathematical reasoning in open language models},
  author={Shao, Zhihong and Wang, Peiyi and Zhu, Qihao and Xu, Runxin and Song, Junxiao and Bi, Xiao and Zhang, Haowei and Zhang, Mingchuan and Li, YK and Wu, Yang and others},
  journal={arXiv preprint arXiv:2402.03300},
  year={2024}
}

@article{schmidgall2024agentclinic,
  title={Agentclinic: a multimodal agent benchmark to evaluate ai in simulated clinical environments},
  author={Schmidgall, Samuel and Ziaei, Rojin and Harris, Carl and Reis, Eduardo and Jopling, Jeffrey and Moor, Michael},
  journal={arXiv preprint arXiv:2405.07960},
  year={2024}
}

@article{wang2026liveclin,
  title={LiveClin: A Live Clinical Benchmark without Leakage},
  author={Wang, Xidong and Guo, Shuqi and Shen, Yue and Chen, Junying and Wang, Jian and Gu, Jinjie and Zhang, Ping and Liu, Lei and Wang, Benyou},
  journal={arXiv preprint arXiv:2602.16747},
  year={2026}
}

@article{wu2024medjourney,
  title={Medjourney: Benchmark and evaluation of large language models over patient clinical journey},
  author={Wu, Xian and Zhao, Yutian and Zhang, Yunyan and Wu, Jiageng and Zhu, Zhihong and Zhang, Yingying and Ouyang, Yi and Zhang, Ziheng and Wang, Huimin and Lin, Zhenxi and others},
  journal={Advances in Neural Information Processing Systems},
  volume={37},
  pages={87621--87646},
  year={2024}
}

@article{yan2026livemedbench,
  title={Livemedbench: A contamination-free medical benchmark for llms with automated rubric evaluation},
  author={Yan, Zhiling and Song, Dingjie and Fang, Zhe and Ji, Yisheng and Li, Xiang and Li, Quanzheng and Sun, Lichao},
  journal={arXiv preprint arXiv:2602.10367},
  year={2026}
}

@article{arora2025healthbench,
  title={Healthbench: Evaluating large language models towards improved human health},
  author={Arora, Rahul K and Wei, Jason and Hicks, Rebecca Soskin and Bowman, Preston and Qui{\~n}onero-Candela, Joaquin and Tsimpourlas, Foivos and Sharman, Michael and Shah, Meghan and Vallone, Andrea and Beutel, Alex and others},
  journal={arXiv preprint arXiv:2505.08775},
  year={2025}
}

@article{li2024mediq,
  title={Mediq: Question-asking llms and a benchmark for reliable interactive clinical reasoning},
  author={Li, Shuyue S and Balachandran, Vidhisha and Feng, Shangbin and Ilgen, Jonathan S and Pierson, Emma and Koh, Pang W and Tsvetkov, Yulia},
  journal={Advances in Neural Information Processing Systems},
  volume={37},
  pages={28858--28888},
  year={2024}
}

@article{zuo2025medxpertqa,
  title={Medxpertqa: Benchmarking expert-level medical reasoning and understanding},
  author={Zuo, Yuxin and Qu, Shang and Li, Yifei and Chen, Zhangren and Zhu, Xuekai and Hua, Ermo and Zhang, Kaiyan and Ding, Ning and Zhou, Bowen},
  journal={arXiv preprint arXiv:2501.18362},
  year={2025}
}

@inproceedings{yue2024mmmu,
  title={Mmmu: A massive multi-discipline multimodal understanding and reasoning benchmark for expert agi},
  author={Yue, Xiang and Ni, Yuansheng and Zhang, Kai and Zheng, Tianyu and Liu, Ruoqi and Zhang, Ge and Stevens, Samuel and Jiang, Dongfu and Ren, Weiming and Sun, Yuxuan and others},
  booktitle={Proceedings of the IEEE/CVF conference on computer vision and pattern recognition},
  pages={9556--9567},
  year={2024}
}

@inproceedings{yue2025mmmu,
  title={Mmmu-pro: A more robust multi-discipline multimodal understanding benchmark},
  author={Yue, Xiang and Zheng, Tianyu and Ni, Yuansheng and Wang, Yubo and Zhang, Kai and Tong, Shengbang and Sun, Yuxuan and Yu, Botao and Zhang, Ge and Sun, Huan and others},
  booktitle={Proceedings of the 63rd Annual Meeting of the Association for Computational Linguistics (Volume 1: Long Papers)},
  pages={15134--15186},
  year={2025}
}

@article{liu2025deepseek,
  title={Deepseek-v3. 2: Pushing the frontier of open large language models},
  author={Liu, Aixin and Mei, Aoxue and Lin, Bangcai and Xue, Bing and Wang, Bingxuan and Xu, Bingzheng and Wu, Bochao and Zhang, Bowei and Lin, Chaofan and Dong, Chen and others},
  journal={arXiv preprint arXiv:2512.02556},
  year={2025}
}

@misc{qwen36plus,
    title = {{Qwen3.6-Plus}: Towards Real World Agents},
    url = {https://qwen.ai/blog?id=qwen3.6},
    author = {{Qwen Team}},
    month = {April},
    year = {2026}
}

@misc{openai2024embedding,
  author = {OpenAI},
  title = {New embedding models and API updates},
  year = {2024},
  url = {https://openai.com/index/new-embedding-models-and-api-updates/}
}

@article{zheng2025skillweaver,
  title={Skillweaver: Web agents can self-improve by discovering and honing skills},
  author={Zheng, Boyuan and Fatemi, Michael Y and Jin, Xiaolong and Wang, Zora Zhiruo and Gandhi, Apurva and Song, Yueqi and Gu, Yu and Srinivasa, Jayanth and Liu, Gaowen and Neubig, Graham and others},
  journal={arXiv preprint arXiv:2504.07079},
  year={2025}
}

@article{tang2025agent,
  title={Agent kb: Leveraging cross-domain experience for agentic problem solving},
  author={Tang, Xiangru and Qin, Tianrui and Peng, Tianhao and Zhou, Ziyang and Shao, Daniel and Du, Tingting and Wei, Xinming and Xia, Peng and Wu, Fang and Zhu, He and others},
  journal={arXiv preprint arXiv:2507.06229},
  year={2025}
}

@article{wu2025evolver,
  title={Evolver: Self-evolving llm agents through an experience-driven lifecycle},
  author={Wu, Rong and Wang, Xiaoman and Mei, Jianbiao and Cai, Pinlong and Fu, Daocheng and Yang, Cheng and Wen, Licheng and Yang, Xuemeng and Shen, Yufan and Wang, Yuxin and others},
  journal={arXiv preprint arXiv:2510.16079},
  year={2025}
}

@article{wang2025mobile,
  title={Mobile-agent-e: Self-evolving mobile assistant for complex tasks},
  author={Wang, Zhenhailong and Xu, Haiyang and Wang, Junyang and Zhang, Xi and Yan, Ming and Zhang, Ji and Huang, Fei and Ji, Heng},
  journal={arXiv preprint arXiv:2501.11733},
  year={2025}
}

@article{gou2023critic,
  title={Critic: Large language models can self-correct with tool-interactive critiquing},
  author={Gou, Zhibin and Shao, Zhihong and Gong, Yeyun and Shen, Yelong and Yang, Yujiu and Duan, Nan and Chen, Weizhu},
  journal={arXiv preprint arXiv:2305.11738},
  year={2023}
}

@inproceedings{yao2022react,
  title={React: Synergizing reasoning and acting in language models},
  author={Yao, Shunyu and Zhao, Jeffrey and Yu, Dian and Du, Nan and Shafran, Izhak and Narasimhan, Karthik R and Cao, Yuan},
  booktitle={The eleventh international conference on learning representations},
  year={2022}
}

@misc{google2025gemini3,
  author = {Google},
  title = {A new era of intelligence with Gemini 3},
  year = {2025},
  url = {https://blog.google/products-and-platforms/products/gemini/gemini-3/}
}

@article{fang2025memp,
  title={Memp: Exploring agent procedural memory},
  author={Fang, Runnan and Liang, Yuan and Wang, Xiaobin and Wu, Jialong and Qiao, Shuofei and Xie, Pengjun and Huang, Fei and Chen, Huajun and Zhang, Ningyu},
  journal={arXiv preprint arXiv:2508.06433},
  year={2025}
}

@article{dou2025baichuan,
  title={Baichuan-m2: Scaling medical capability with large verifier system},
  author={Dou, Chengfeng and Liu, Chong and Yang, Fan and Li, Fei and Jia, Jiyuan and Chen, Mingyang and Ju, Qiang and Wang, Shuai and Dang, Shunya and Li, Tianpeng and others},
  journal={arXiv preprint arXiv:2509.02208},
  year={2025}
}

@article{jin2021disease,
  title={What disease does this patient have? a large-scale open domain question answering dataset from medical exams},
  author={Jin, Di and Pan, Eileen and Oufattole, Nassim and Weng, Wei-Hung and Fang, Hanyi and Szolovits, Peter},
  journal={Applied Sciences},
  volume={11},
  number={14},
  pages={6421},
  year={2021},
  publisher={MDPI}
}

@inproceedings{pal2022medmcqa,
  title={Medmcqa: A large-scale multi-subject multi-choice dataset for medical domain question answering},
  author={Pal, Ankit and Umapathi, Logesh Kumar and Sankarasubbu, Malaikannan},
  booktitle={Conference on health, inference, and learning},
  pages={248--260},
  year={2022},
  organization={PMLR}
}

@inproceedings{chen2025benchmarking,
  title={Benchmarking large language models on answering and explaining challenging medical questions},
  author={Chen, Hanjie and Fang, Zhouxiang and Singla, Yash and Dredze, Mark},
  booktitle={Proceedings of the 2025 Conference of the Nations of the Americas Chapter of the Association for Computational Linguistics: Human Language Technologies (Volume 1: Long Papers)},
  pages={3563--3599},
  year={2025}
}

@article{singhal2023large,
  title={Large language models encode clinical knowledge},
  author={Singhal, Karan and Azizi, Shekoofeh and Tu, Tao and Mahdavi, S Sara and Wei, Jason and Chung, Hyung Won and Scales, Nathan and Tanwani, Ajay and Cole-Lewis, Heather and Pfohl, Stephen and others},
  journal={Nature},
  volume={620},
  number={7972},
  pages={172--180},
  year={2023},
  publisher={Nature Publishing Group UK London}
}

@article{singhal2025toward,
  title={Toward expert-level medical question answering with large language models},
  author={Singhal, Karan and Tu, Tao and Gottweis, Juraj and Sayres, Rory and Wulczyn, Ellery and Amin, Mohamed and Hou, Le and Clark, Kevin and Pfohl, Stephen R and Cole-Lewis, Heather and others},
  journal={Nature medicine},
  volume={31},
  number={3},
  pages={943--950},
  year={2025},
  publisher={Nature Publishing Group US New York}
}

@article{topol2019high,
  title={High-performance medicine: the convergence of human and artificial intelligence},
  author={Topol, Eric J},
  journal={Nature medicine},
  volume={25},
  number={1},
  pages={44--56},
  year={2019},
  publisher={Nature Publishing Group US New York}
}

@article{esteva2019guide,
  title={A guide to deep learning in healthcare},
  author={Esteva, Andre and Robicquet, Alexandre and Ramsundar, Bharath and Kuleshov, Volodymyr and DePristo, Mark and Chou, Katherine and Cui, Claire and Corrado, Greg and Thrun, Sebastian and Dean, Jeff},
  journal={Nature medicine},
  volume={25},
  number={1},
  pages={24--29},
  year={2019},
  publisher={Nature Publishing Group US New York}
}

@article{tulving1972episodic,
  title={Episodic and semantic memory},
  author={Tulving, Endel and others},
  journal={Organization of memory},
  volume={1},
  number={381-403},
  pages={1},
  year={1972},
  publisher={New York}
}

@article{kolodner1992introduction,
  title={An introduction to case-based reasoning},
  author={Kolodner, Janet L},
  journal={Artificial intelligence review},
  volume={6},
  number={1},
  pages={3--34},
  year={1992},
  publisher={Springer}
}

@article{barrows1987clinical,
  title={The clinical reasoning process},
  author={Barrows, Howard S and Feltovich, Paul J},
  journal={Medical education},
  volume={21},
  number={2},
  pages={86--91},
  year={1987},
  publisher={Wiley Online Library}
}

@article{lai2025patient,
  title={Patient-zero: A unified framework for real-record-free patient agent generation},
  author={Lai, Yunghwei and Ma, Weizhi and Liu, Yang},
  journal={arXiv e-prints},
  pages={arXiv--2509},
  year={2025}
}

@inproceedings{wei2024medco,
  title={Medco: Medical education copilots based on a multi-agent framework},
  author={Wei, Hao and Qiu, Jianing and Yu, Haibao and Yuan, Wu},
  booktitle={European Conference on Computer Vision},
  pages={119--135},
  year={2024},
  organization={Springer}
}

@article{ren2025healthcare,
  title={Healthcare agent: eliciting the power of large language models for medical consultation},
  author={Ren, Zhiyao and Zhan, Yibing and Yu, Baosheng and Ding, Liang and Xu, Pingbo and Tao, Dacheng},
  journal={npj Artificial Intelligence},
  volume={1},
  number={1},
  pages={24},
  year={2025},
  publisher={Nature Publishing Group UK London}
}

@article{xu2026comprehensive,
  title={A comprehensive survey of AI Agents in Healthcare},
  author={Xu, Gelei and Li, Xueyang and Chen, Yixiong and Duan, Yuying and Wu, Shuqing and Yu, Haoxinran and Chiu, Ching-Hao and Ni, Juntong and Tang, Ningzhi and Li, Toby Jia-Jun and others},
  journal={Journal of Biomedical Informatics},
  pages={105045},
  year={2026},
  publisher={Elsevier}
}

@article{hu2026landscape,
  title={The landscape of medical agents: A survey},
  author={Hu, Xiaobin and Qian, Yunhang and Yu, Jiaquan and Liu, Jingjing and Ji, Xiaozhong and Xu, Chengming and Tang, Peng and Xu, Chengming and Tang, Peng and Liu, Jiawei and others},
  year={2026},
  publisher={TechRxiv}
}

@article{guo2024ds,
  title={Ds-agent: Automated data science by empowering large language models with case-based reasoning},
  author={Guo, Siyuan and Deng, Cheng and Wen, Ying and Chen, Hechang and Chang, Yi and Wang, Jun},
  journal={arXiv preprint arXiv:2402.17453},
  year={2024}
}

@inproceedings{guo2025optimizing,
  title={Optimizing case-based reasoning system for functional test script generation with large language models},
  author={Guo, Siyuan and Liu, Huiwu and Chen, Xiaolong and Xie, Yuming and Zhang, Liang and Han, Tao and Chen, Hechang and Chang, Yi and Wang, Jun},
  booktitle={Proceedings of the 31st ACM SIGKDD Conference on Knowledge Discovery and Data Mining V. 2},
  pages={4487--4498},
  year={2025}
}

@misc{tavilyai_github,
  author       = {{Tavily AI}},
  title        = {{Tavily AI GitHub Organization}},
  year         = {2026},
  howpublished = {\url{https://github.com/tavily-ai}},
}

@article{wishart2018drugbank,
  title={DrugBank 5.0: a major update to the DrugBank database for 2018},
  author={Wishart, David S and Feunang, Yannick D and Guo, An C and Lo, Elvis J and Marcu, Ana and Grant, Jason R and Sajed, Tanvir and Johnson, Daniel and Li, Carin and Sayeeda, Zinat and others},
  journal={Nucleic acids research},
  volume={46},
  number={D1},
  pages={D1074--D1082},
  year={2018},
  publisher={Oxford University Press}
}

@article{bodenreider2004unified,
  title={The unified medical language system (UMLS): integrating biomedical terminology},
  author={Bodenreider, Olivier},
  journal={Nucleic acids research},
  volume={32},
  number={suppl\_1},
  pages={D267--D270},
  year={2004},
  publisher={Oxford University Press}
}

@article{bai2025qwen3,
  title={Qwen3-vl technical report},
  author={Bai, Shuai and Cai, Yuxuan and Chen, Ruizhe and Chen, Keqin and Chen, Xionghui and Cheng, Zesen and Deng, Lianghao and Ding, Wei and Gao, Chang and Ge, Chunjiang and others},
  journal={arXiv preprint arXiv:2511.21631},
  year={2025}
}

@inproceedings{kwon2023efficient,
  title={Efficient Memory Management for Large Language Model Serving with PagedAttention},
  author={Woosuk Kwon and Zhuohan Li and Siyuan Zhuang and Ying Sheng and Lianmin Zheng and Cody Hao Yu and Joseph E. Gonzalez and Hao Zhang and Ion Stoica},
  booktitle={Proceedings of the ACM SIGOPS 29th Symposium on Operating Systems Principles},
  year={2023}
}

@misc{qwen36_max_preview,
    title = {{Qwen3.6-Max-Preview}: Smarter, Sharper, Still Evolving},
    url = {https://qwen.ai/blog?id=qwen3.6-max-preview},
    author = {{Qwen Team}},
    month = {April},
    year = {2026}
}

@misc{moonshot2026kimiK26,
  author       = {{Moonshot AI}},
  title        = {{Kimi K2.6: Advancing Open-Source Coding}},
  year         = {2026},
  howpublished = {\url{https://www.kimi.com/blog/kimi-k2-6}},
}

@article{zeng2026glm,
  title={Glm-5: from vibe coding to agentic engineering},
  author={Zeng, Aohan and Lv, Xin and Hou, Zhenyu and Du, Zhengxiao and Zheng, Qinkai and Chen, Bin and Yin, Da and Ge, Chendi and Huang, Chenghua and Xie, Chengxing and others},
  journal={arXiv preprint arXiv:2602.15763},
  year={2026}
}

@misc{anthropic_claude_sonnet,
  author       = {{Anthropic}},
  title        = {{Claude Sonnet 4.6}},
  year         = {2026},
  howpublished = {\url{https://www.anthropic.com/claude/sonnet}}
}
